\pdfoutput=1

\documentclass[11pt]{article}

\usepackage[preprint]{acl}
\usepackage{times}
\usepackage{latexsym}
\usepackage[T1]{fontenc}
\usepackage[utf8]{inputenc}
\usepackage[most]{tcolorbox}
\usepackage{microtype}
\usepackage{inconsolata}
\usepackage{fontawesome}
\usepackage{graphicx}
\usepackage{setspace}
\usepackage{booktabs}
\usepackage{xcolor}
\usepackage{threeparttable}
\usepackage{CJKutf8}
\usepackage{enumerate}
\usepackage{amsfonts,amssymb,amsmath} 
\usepackage{color}
\usepackage{graphicx}
\usepackage{tabularx}
\usepackage{algorithm}
\usepackage{algorithmic}
\usepackage{amsthm}
\usepackage{makecell}
\usepackage{graphicx}
\usepackage{soul} 
\usepackage{subcaption}
\usepackage{multirow}
\usepackage{colortbl}
\usepackage{pgf}
\usepackage{float}
\usepackage{pifont}
\usepackage[normalem]{ulem}
\usepackage{longtable}

\definecolor{myblue2}{HTML}{2E9ADF}
\definecolor{myred}{HTML}{FFC8C8}
\definecolor{myblue}{HTML}{C8E7FF}

\definecolor{color0}{rgb}{1,1,1}
\definecolor{color5}{rgb}{0.5,0.7,0.8}

\newcommand{\ccv}[1]{
  \pgfmathsetmacro{\opacity}{#1/5}
  \pgfmathsetmacro{\mixfactor}{\opacity*200}
  \edef\temp{\noexpand\cellcolor{color5!\mixfactor!color0}}
  \temp #1
}

\usepackage{enumitem}
\usepackage{multirow}    
\usepackage{array}       
\usepackage{caption}
\usepackage{geometry}
\usepackage{listings, tcolorbox,soul}
\tcbuselibrary{listings, most}

\definecolor{myyellow}{RGB}{255,255,204}
\definecolor{myblue}{RGB}{221,233,247}
\definecolor{mygrey}{RGB}{128,128,128}

\usepackage{varwidth}

\lstdefinelanguage{mycase}{
    basicstyle=\scriptsize\ttfamily, 
    moredelim = [s][\color{mygrey}]{\{}{\}},
}
\newtcblisting{showcase}[1][]{
  enhanced,
  arc=0em,
  boxrule=.5pt,
  listing only,
  listing options={
    language=mycase,
    upquote=true,
    basicstyle=\scriptsize\ttfamily \setlength{\baselineskip}{1.1\baselineskip},
    breaklines=true,
    breakindent=0pt,
    xleftmargin=0pt,
    xrightmargin=0pt,
    aboveskip=-4pt,
    belowskip=-4pt,
    columns=fullflexible,
    escapeinside={|}{|},
  },
  colback=white,
  colframe=gray,
  breakable,
  colbacktitle=gray!10,
  coltitle=black,
  attach boxed title to top center={yshift=-3mm},
  #1
}

\newtcolorbox[list inside=prompt,auto counter,number within=section]{prompt}[1][]{
    colbacktitle=black!60,
    coltitle=white,
    fontupper=\footnotesize,
    boxsep=5pt,
    left=0pt,
    right=0pt,
    top=0pt,
    bottom=0pt,
    boxrule=1pt,
    #1,
}

\newtcolorbox[auto counter,number within=chapter]{prompt2}[1][]{
  enhanced,
  breakable,
  fontupper=\footnotesize,
  fonttitle=\scshape,
  title={Definition \thetcbcounter},
  #1,
}
\usepackage[%
    framemethod=tikz,
    skipbelow=\topskip,
    skipabove=\topskip
]{mdframed}
\mdfsetup{%
    leftmargin=0pt,
    rightmargin=0pt,
    backgroundcolor=bggray,
    middlelinecolor=black,
    roundcorner=3
}

\newmdenv[
  backgroundcolor=black!05,
  linecolor=quoteborder,
  skipabove=1em,
  skipbelow=1em,
  leftline=true,
  topline=false,
  bottomline=false,
  rightline=false,
  linecolor=black!40,
  linewidth=4pt,
  font=\small,
  leftmargin=0cm
]{prompt_env}

\title{Aspect-Guided Multi-Level Perturbation Analysis of Large Language Models in Automated Peer Review\\
~\\
{\begin{center}
    \small
    \textcolor{orange}{\bf \faWarning\, WARNING: This paper contains model outputs that may be considered offensive.}
\end{center}
\vspace{-0.5em}
}}

\author{
 \textbf{Jiatao Li\textsuperscript{1,2}},
 \textbf{Yanheng Li\textsuperscript{3}},
 \textbf{Xinyu Hu\textsuperscript{1}},
 \textbf{Mingqi Gao\textsuperscript{1}},
 \textbf{Xiaojun Wan\textsuperscript{1}}
\\
 \textsuperscript{1}Wangxuan Institute of Computer Technology, Peking University
\\
 \textsuperscript{2}Information Management Department, Peking University
 \\
 \textsuperscript{3}Renmin University of China
 \\
 \texttt{leejames@stu.pku.edu.cn}, \texttt{yanheng@ruc.edu.cn} \\
 \texttt{\{huxinyu,gaomingqi,wanxiaojun\}@pku.edu.cn} \\
}

\begin{document}
\maketitle
\begin{abstract}
We propose an \emph{aspect-guided, multi-level perturbation framework} to evaluate the robustness of Large Language Models (LLMs) in automated peer review. Our framework explores perturbations in three key components of the peer review process—papers, reviews, and rebuttals—across several quality aspects, including contribution, soundness, presentation, tone, and completeness. By applying targeted perturbations and examining their effects on both \emph{LLM-as-Reviewer} and \emph{LLM-as-Meta-Reviewer}, we investigate how aspect-based manipulations, such as omitting methodological details from papers or altering reviewer conclusions, can introduce significant biases in the review process. We identify several potential vulnerabilities: review conclusions that recommend a ``strong reject'' may significantly influence meta-reviews, negative or misleading reviews may be wrongly interpreted as thorough, and incomplete or hostile rebuttals can unexpectedly lead to higher acceptance rates. Statistical tests show that these biases persist under various Chain-of-Thought prompting strategies, highlighting the lack of robust critical evaluation in current LLMs. Our framework offers a practical methodology for diagnosing these vulnerabilities, thereby contributing to the development of more reliable and robust automated reviewing systems.
\end{abstract}

\section{Introduction}
\begin{figure*}[t]
    \centering
    \includegraphics[width=0.95\linewidth]{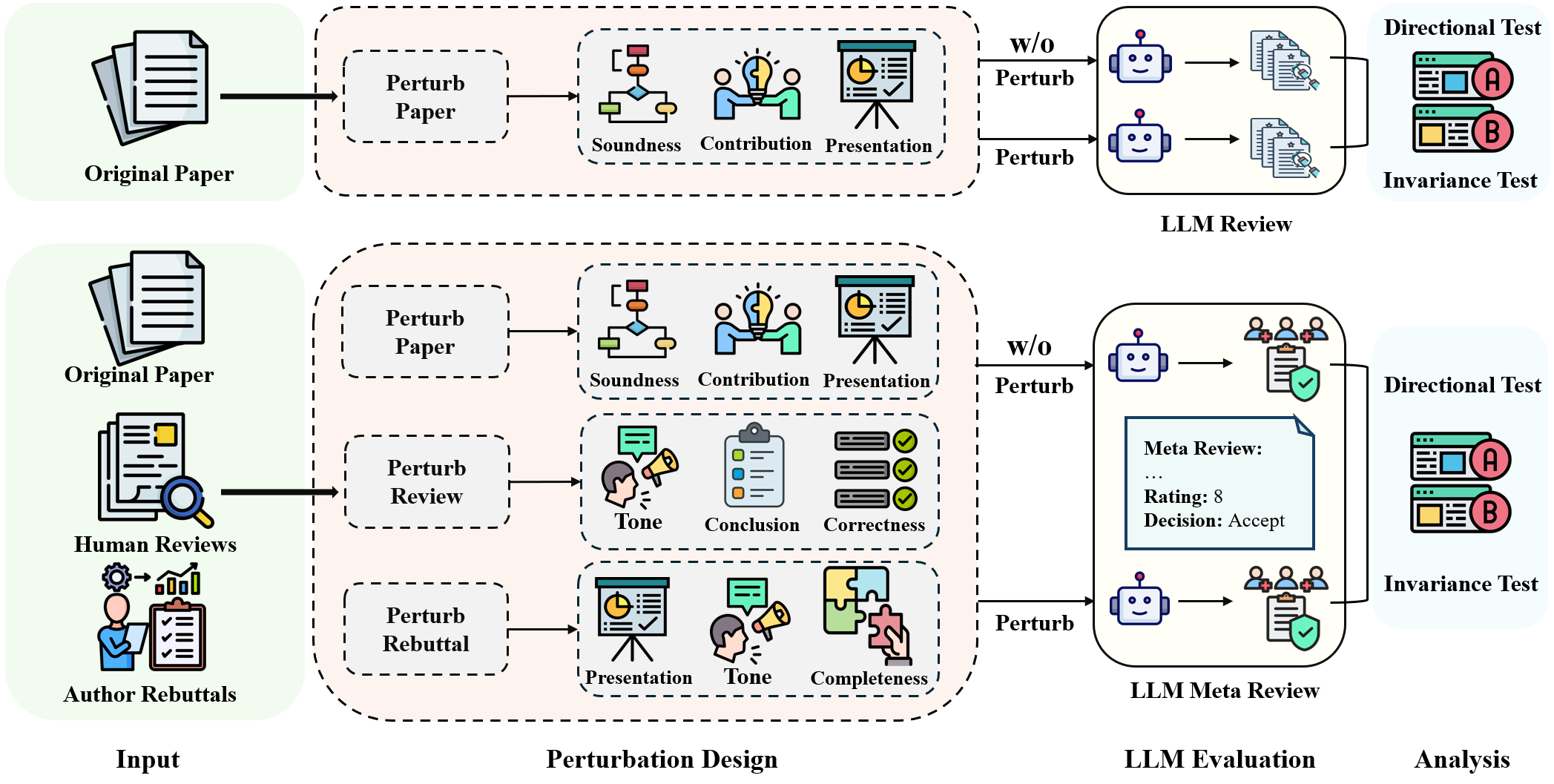}
    \caption{Overview of Our Single-Target Perturbation Approach. We perturb only one component at a time – either the paper, the reviews, or the rebuttal – and then observe the effect of that single change on the LLM acting as either a reviewer or a meta-reviewer.}
    \label{fig:pipeline}
\end{figure*}

Peer review, the cornerstone of scientific validation, is facing increasing strain due to the rapid growth in manuscript submissions \citep{bornmann2015growth}. The traditional review process is facing challenges in securing sufficient qualified reviewers, and it remains both time-consuming and inherently subjective. \citep{aczel2021billion}. This has led to explorations of automated and semi-automated reviewing pipelines, leveraging recent advances in Large Language Models (LLMs) \citep{openai2024gpt4technicalreport}.

LLMs hold the potential to accelerate peer review by drafting critiques, summarizing papers, and consolidating feedback \citep{weissgerber2021automated}. Early studies suggest some overlap between LLM-assisted reviews and human reviews, and AI conferences already report non-trivial use of AI in the review process \citep{liangmonitoring}. This has spurred proposals for fully automated peer review systems \citep{tyser2024ai}.

However, concerns about the reliability, fairness, and transparency of LLM-based reviews remain paramount. While existing research has highlighted LLM vulnerabilities like hallucination, bias, and susceptibility to adversarial edits \citep{zeng2024johnny, gallegos2024bias, xie2023defending,ye2024we}, these studies often focus on isolated aspects of the review process. They do not fully capture the complex interplay among the \emph{paper}, the \emph{reviews}, and the \emph{author rebuttal}—the three core components of peer review. Crucially, prior work lacks a unified, multi-level framework to systematically analyze how targeted manipulations of each component propagate through an LLM-based review pipeline, impacting both individual review scores and the overall meta-review decision.

To address this critical gap, we propose an \emph{aspect-guided, multi-level perturbation analysis} for evaluating LLMs in automated peer review. Our framework (Figure~\ref{fig:pipeline}) modifies each component—paper, review, and rebuttal—along key aspects (e.g., contribution, tone). We focus on single-target perturbations: we alter only one of the paper, the reviews, or the rebuttal at a time. This creates a perturbed input alongside the original, unperturbed input. These inputs are fed into either \textit{LLM-as-Reviewer} (for paper perturbations) or \textit{LLM-as-Meta-Reviewer} (for any of the three perturbation types). Comparing outputs from these scenarios reveals how single-target manipulations, such as removing key methods or flipping a reviewer's recommendation, cascade through the system, impacting scores and decisions.

Analyzing 508 accepted ICLR 2024 papers, we find that LLM-based reviewing pipelines often fail to discount biased or erroneous signals. Even without altering a paper’s underlying quality, manipulations such as adding false criticisms, flipping a positive recommendation, or withholding key rebuttal details can disproportionately shift system verdicts. These distortions even persist under various Chain-of-Thought prompting strategies. Our findings highlight the need for stronger oversight, adversarial defenses, and transparent review processes before fully automating scholarly peer review.

Our contributions are threefold: \footnote{Our code and data will be released to the community to facilitate future research.}:

\begin{itemize}[leftmargin=1em]

\item We curate a dataset comprising 79,302 adversarial, aspect-guided text edits, which incorporate perturbations across nine distinct aspects affecting reviews, papers, and rebuttals, all guided by our perturbation aspect taxonomy.

\item Our \emph{aspect-guided, multi-level} approach effectively integrates both LLM reviewer and meta-reviewer perturbations, employing directional and invariance tests to assess the impact of these perturbations.

\item Our findings illuminate vulnerabilities within the LLM-based review system, revealing that the “strong reject” conclusion can overshadow all other feedback. Furthermore, hostile or erroneous critiques may be misinterpreted as thoroughness, underscoring the urgent need for enhanced safeguards.

\end{itemize}

\section{Perturbation Design}

We present our framework for assessing the robustness of LLMs in automated peer review.  By introducing targeted perturbations at various components—specifically to the paper, the reviews, and the rebuttal—and across multiple quality dimensions, we are able to evaluate the effects of these manipulations on the behavior of LLMs functioning as both reviewers and meta-reviewers. We first describe the LLM roles within our framework, then formalize our single-target perturbation approach, detail the specific perturbation strategies, and finally describe the statistical tests used for analysis.

\subsection{LLM Roles in Peer Review}
We define two distinct roles for the LLMs in our framework, mirroring the human roles in the peer review process:

\paragraph{LLM-as-Reviewer} This LLM role, denoted as \(M_{\mathrm{rev}}\), processes only the paper text, \(p\), and outputs dimension-specific scores (e.g., Contribution, Soundness, Presentation) plus a final recommendation: \(M_{\mathrm{rev}}(p) \to \text{LLM Review}\). \footnote{In practice, a human reviewer may also consult the rebuttal or engage in follow-up discussions, but some do not consistently do so. For simplicity and experimental tractability, our \textit{LLM-as-Reviewer} setting only ingests the paper.}

\paragraph{LLM-as-Meta-Reviewer} This LLM role, denoted as \(M_{\mathrm{meta}}\), takes as input the paper, \(p\), multiple reviews, \(\{r_i\}\), and the rebuttal, \(b\):  \(M_{\mathrm{meta}}(p, r_1, \dots, r_k, b) \to \text{LLM Meta-Review}\).  It produces an overall score and final decision, reflecting the integrated viewpoints of multiple reviewers and the authors' responses.

\subsection{Single-Target Perturbation}
\label{subsec:single_target}

To analyze how each component (paper, review, rebuttal) affects LLM-based evaluations, we introduce a \emph{single-target} perturbation approach. This means that we modify only one of these components at a time. Let \( T = (p, r_1, \dots, r_k, b) \) denote the original, unperturbed inputs (the paper, a set of reviews, and the rebuttal). We define three perturbation functions:

1) \(F_{\mathrm{paper}}(T)\):  Modifies the paper, resulting in \((p', r_1, \dots, r_k, b)\).
2) \(F_{\mathrm{review}}(T)\): Modifies the reviews, resulting in \((p, r_1', \dots, r_k', b)\).
3) \(F_{\mathrm{rebuttal}}(T)\): Modifies the rebuttal, resulting in \((p, r_1, \dots, r_k, b')\).

Applying any of these functions, \(F_{\alpha}(T)\), produces a perturbed input, \(T'\).  We then compare the LLM outputs for the perturbed input, \(o_{\mathrm{pert}}\), with the unperturbed baseline outputs, \(o_{\mathrm{base}}\). For paper perturbations, we evaluate both  \(M_{\mathrm{rev}}(p')\) and \(M_{\mathrm{meta}}(p', r_1, \dots, r_k, b)\). For review or rebuttal perturbations, we keep the paper, \(p\), constant and evaluate only the meta-reviewer: \(M_{\mathrm{meta}}(p, r_1', \dots, r_k', b)\) or \(M_{\mathrm{meta}}(p, r_1, \dots, r_k, b')\).

\subsection{Perturbation Taxonomy}
\label{subsec:main_paper_section_taxonomy}

\begin{figure*}[t]
    \centering
    \includegraphics[width=\linewidth]{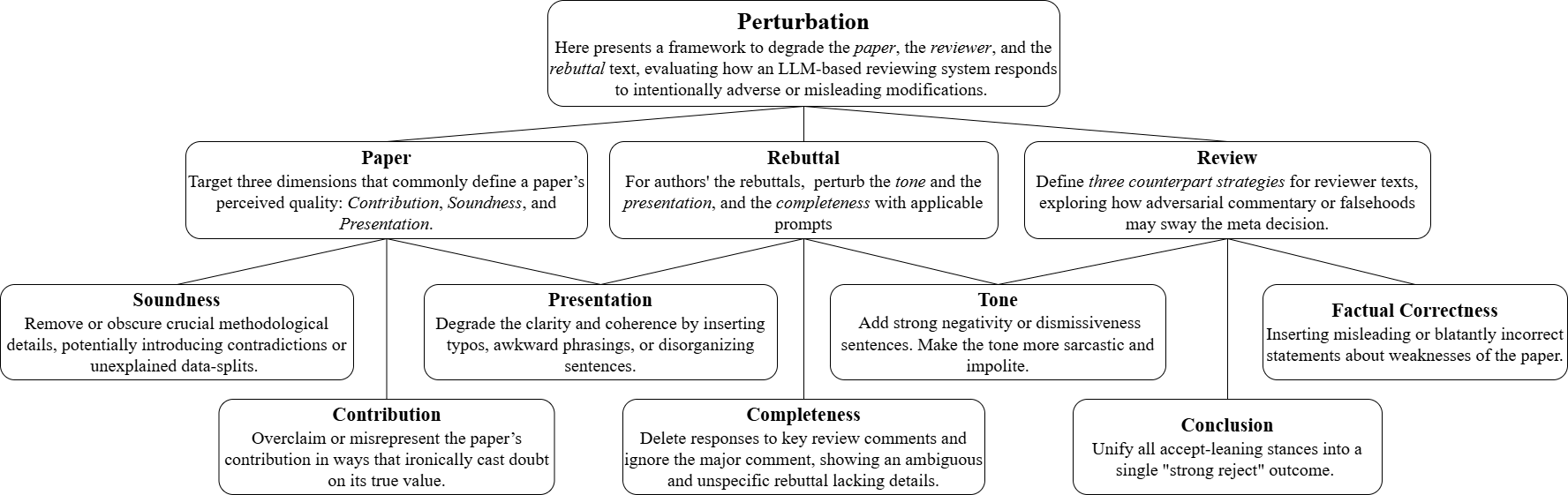}
    \caption{Taxonomy of perturbation strategies for altering the \textbf{paper}, \textbf{review}, or \textbf{rebuttal} text.}
    \label{fig:perturb_design}
\end{figure*}

We introduce a perturbation taxonomy (Figure~\ref{fig:perturb_design}) designed to rigorously assess the robustness of LLMs against targeted manipulations of the core peer review components: paper, review, and rebuttal.

\paragraph{Paper-Level}
1) The perturbation of contribution exaggerates or distorts the novelty and discoveries, 2) the perturbation of soundness removes or conceals essential technical details, 3) the perturbation of presentation introduces typographical errors or disorganized text that undermine the clarity of the paper.

\paragraph{Review-Level}
1) The perturbation of tone injects hostility or negativity into reviewer comments, 2) the perturbation of factual correctness introduces false statements as criticisms, 3) the perturbation of conclusion changes a positive recommendation into a strong rejection.

\paragraph{Rebuttal-Level}
1) The perturbation of tone makes the author’s response hostile, 2) the perturbation of presentation degrades the clarity of the rebuttal through poor formatting or typos, 3) the perturbation of completeness omits substantive replies to major reviewer concerns.

Further implementation details for each aspect appear in Appendix~\ref{appendix:detail_perturbation_design}.

\subsection{Statistical Tests}
\label{subsec:statistical_tests}
Our framework encompasses two core methods for assessing performance shifts:

\paragraph{Directional Test}
\label{subsec:directional_test}
The \emph{Directional Test} determines whether meaningful perturbations produce the expected \emph{direction} of change in model outputs (e.g., lowered scores for more flawed content). Concretely, we employ the \textbf{Wilcoxon Signed-Rank Test}~\cite{wilcoxon} to detect if the model’s scores significantly increase or decrease after perturbation.

\paragraph{Invariance Test}
\label{subsec:invariance_test}
The \emph{Invariance Test} checks whether irrelevant changes fail to produce significant differences, implying the model should yield \emph{invariant} outputs if the core content remains unaffected. We adopt the \textbf{Two One-Sided Tests (TOST)}~\cite{lakens2017equivalence} procedure for equivalence.

\section{Experimental Setup}
\paragraph{Data Collection}
\label{subsec:data_collection}

We focus on \emph{accepted} ICLR 2024 papers to ensure a robust baseline for our perturbation tests.  Accepted papers are more likely to be well-written and methodologically sound, allowing us to isolate the effects of our perturbations. Using a stratified sample that mirrors official acceptance proportions (poster, spotlight, oral), we obtain \textbf{508} papers (406 posters, 83 spotlights, 19 orals), each with multiple reviews and a rebuttal. To preserve structural fidelity (e.g., formulas, tables), we rely on the ICLR~2024 parsed dataset by \citet{yu2024automated}, which encodes math formulas, tables to \LaTeX{} and keeps section hierarchies for reliable LLM processing.

\paragraph{Review Setup}
\label{sec:evaluation_aspects}
To ensure our evaluation aligns with real-world peer review practices, we primarily follow ICLR guidelines \footnote{\url{https://www.iclr.cc/Conferences/2024/ReviewerGuide}} to structure both \emph{LLM-as-Reviewer} and \emph{LLM-as-Meta-Reviewer} outputs. For \emph{LLM-as-Reviewer}, we use the ICLR template covering strengths, weaknesses, numeric ratings (1--4 for each dimension), and a final recommendation (1--10). For \emph{LLM-as-Meta-Reviewer}, we compare three configurations: a minimal version outputting only an overall score (None CoT), a version assigning separate scores for each dimension (Dimension CoT), and an ICLR-based template offering brief justifications (Template COT). This setup allows us to observe how different levels of explanation affect susceptibility to manipulative inputs.

\paragraph{LLMs}
\label{sec:llms}
We adopt \emph{GPT-4o}~\cite{achiam2023gpt} as our primary large language model for both generating text perturbations (paper, review, and rebuttal) and producing the subsequent reviews or meta-reviews. Additionally, we evaluate \emph{gemini-2.0-flash-001}~\cite{geminiteam2024geminifamilyhighlycapable} on the same tasks as additional results in Appendix~\ref{gemini}, allowing for a comparative analysis of their responses under identical perturbation scenarios. All models are prompted in zero-shot setting\footnote{Following prior work \citep{yu2024automated,jin2024agentreview}, we use a zero-shot setting due to the extensive input context length, which makes few-shot prompting impractical.}.

\section{LLM as Reviewer}

In this section, We present the results of applying the perturbation framework to the \emph{LLM-as-Reviewer} role. We analyze how targeted modifications to the input paper affect the LLM's generated reviews and scores. Our analysis is structured around two key tests: a directional test (assessing whether perturbations cause changes in the expected direction), an invariance test (examining whether perturbations lead to changes where they should not), and a summary of key takeaways.

Table~\ref{tab:review_delta_avg_colored} shows the average differences in scores between the perturbed and baseline (unperturbed) conditions. Table~\ref{tab:review-template-dim-test-results-main} presents the results of statistical tests (both Wilcoxon Signed-Rank Test for directional changes and Two One-Sided Tests (TOST) for invariance) to determine the significance of the observed changes.

\begin{table}[!ht]
\centering
\small
\setlength{\tabcolsep}{5pt}  
\begin{tabular}{lcccc}
\toprule
\textbf{Aspect} & \textbf{Contr.} & \textbf{Sound.} & \textbf{Prese.} & \textbf{Overall} \\
\midrule
Contribution  & \cellcolor{myblue2!3}\underline{-0.03} & \cellcolor{myblue2!2}\underline{-0.02} & \cellcolor{myblue2!1}\underline{-0.01} & \cellcolor{myblue2!4}\underline{-0.04} \\
Soundness     & \cellcolor{myblue2!2}\underline{-0.02} & \cellcolor{myblue2!0}{-0.00}            & \cellcolor{myblue2!0}{-0.00}            & \cellcolor{myblue2!7}\underline{-0.07} \\
Presentation  & \cellcolor{myblue2!0}{0.00}            & \cellcolor{myblue2!1}\underline{-0.01} & \cellcolor{myblue2!0}{-0.00}           & \cellcolor{myblue2!4}\underline{-0.04} \\
\bottomrule
\end{tabular}
\caption{Average score differences (perturbed - baseline) for LLM-as-Reviewer, by perturbation aspect (rows) and affected score (columns). Dimension scores: 1-4; Overall Rating: 1-10. Shading indicates magnitude; negative values are \underline{underlined}.}
\label{tab:review_delta_avg_colored}
\end{table}

\begin{table}[t]
\centering
\small
\begin{tabular}{>{\raggedright\arraybackslash}m{1.7cm}ll}
\toprule
\textbf{Aspect} & \textbf{Score Column} & \textbf{Direction} \\
\midrule
\multirow{4}{*}{\textbf{Contribution}} 
  & contribution\_score   & \textcolor{blue!70!black}{$\leftrightarrow$} (NSD + Equiv) \\
  & soundness\_score      & \textcolor{blue!70!black}{$\leftrightarrow$} (NSD + Equiv) \\
  & overall\_score        & \textcolor{blue!70!black}{$\leftrightarrow$} (NSD + Equiv) \\
  & presentation\_score   & \textcolor{blue!70!black}{$\leftrightarrow$} (NSD + Equiv) \\
\addlinespace
\multirow{4}{*}{\textbf{Presentation}} 
  & contribution\_score   & \textcolor{blue!70!black}{$\leftrightarrow$} (NSD + Equiv) \\
  & soundness\_score      & \textcolor{blue!70!black}{$\leftrightarrow$} (NSD + Equiv) \\
  & overall\_score        & \textcolor{blue!70!black}{$\leftrightarrow$} (NSD + Equiv) \\
  & presentation\_score   & \textcolor{blue!70!black}{$\leftrightarrow$} (NSD + Equiv) \\
\addlinespace
\multirow{4}{*}{\textbf{Soundness}} 
  & contribution\_score   & \textcolor{blue!70!black}{$\leftrightarrow$} (NSD + Equiv) \\
  & soundness\_score      & \textcolor{blue!70!black}{$\leftrightarrow$} (NSD + Equiv) \\
  & overall\_score        & \textcolor{blue!70!black}{$\leftrightarrow$} (NSD + Equiv) \\
  & presentation\_score   & \textcolor{red!70!black}{$\downarrow$} (before $>$ after) \\
\bottomrule
\end{tabular}
\caption{Statistical test results for LLM-as-Reviewer, by perturbation aspect and score.  \textcolor{blue!70!black}{$\leftrightarrow$}: No Significant Difference + Equivalence (Wilcoxon and TOST); \textcolor{red!70!black}{$\downarrow$}: Significant decline (Wilcoxon). Equivalence margins: ±0.5 (dimension scores), ±1.0 (overall score).}
\label{tab:review-template-dim-test-results-main}
\end{table}

\subsection{Directional Test}

The directional test evaluates whether modifications aimed at undermining specific aspects of the paper (Contribution, Soundness, Presentation) lead to statistically significant reductions in the corresponding scores and overall rating.

1) Contribution Perturbation: Exaggerating claims of novelty results in small average decreases across all scores (see Table \ref{tab:review_delta_avg_colored}), ranging from -0.01 to -0.04. Although all scores show a decrease on average, Table \ref{tab:review-template-dim-test-results-main} indicates that none of these reductions are statistically significant, as they fall within the equivalence margin.

2) Soundness Perturbation: Omitting key methodological details leads to the largest average decrease in the overall rating (-0.07, as shown in Table \ref{tab:review_delta_avg_colored}), along with a significant decrease in the presentation score (refer to Table \ref{tab:review-template-dim-test-results-main}). However, the soundness score itself does not experience a significant decrease.

3) Presentation Perturbation: Introducing typos and disorganization results in a small average decrease in the overall score (-0.04) and in the soundness score (-0.01), as indicated in Table \ref{tab:review_delta_avg_colored}. Nonetheless, Table \ref{tab:review-template-dim-test-results-main} reveals that these changes are also not statistically significant and remain within the equivalence margin.

\subsection{Invariance Test}
\label{sec:review_invariance}

The invariance test examines whether perturbations to one aspect of the paper (e.g., presentation) inappropriately affect scores for other aspects (e.g., contribution, soundness). We expect a robust system to exhibit invariance: changes to presentation quality, for instance, should not significantly impact judgments of contribution or soundness. We use the Two One-Sided Tests (TOST) procedure to assess equivalence when determining no significant difference with the Wilcoxon test. Table~\ref{tab:review-template-dim-test-results-main} shows the results. 

1) Contribution Perturbation: As anticipated, and confirmed by TOST, perturbations to the contribution aspect do not significantly affect the scores of other dimensions (all \textcolor{blue!70!black}{$\leftrightarrow$}). 

2) Presentation Perturbation: Similarly, superficial changes to presentation do not significantly alter scores of contribution or soundness, demonstrating appropriate invariance (all \textcolor{blue!70!black}{$\leftrightarrow$}). 

3) Soundness Perturbation: Critically, while the Soundness score itself shows no significant difference (and TOST confirms equivalence, \textcolor{blue!70!black}{$\leftrightarrow$}), the Presentation score shows a statistically significant decrease (\textcolor{red!70!black}{$\downarrow$}). This violates the expected invariance: a flaw in soundness is being misattributed to an issue in presentation.

\subsection{Takeaways}

Based on the results above, the \emph{LLM-as-Reviewer} experiments reveal several key vulnerabilities:

\paragraph{1) Limited Sensitivity to Substantive Flaws} While the LLM penalizes some paper-level manipulations, these penalties are often mild, especially for issues related to contribution and presentation. The negative directional changes, though aligned with our expectation, are not evidential enough to show statistical significance.

\paragraph{2) Misinterpretation of Missing Information} The most significant finding is that the LLM sometimes penalizes abstract methodological details without explanation (which is caused by the paper soundness perturbation) as presentation problems, rather than recognizing them as core soundness issues (e.g., see Figure~\ref{fig:case_review-paper_soundness} in Appendix~\ref{Case Study}). This is a clear violation of expected invariance and reveals a fundamental misunderstanding of the different quality dimensions.

\paragraph{3) Partial Robustness to Superficial Errors} The LLM does demonstrate appropriate invariance to superficial presentation errors when assessing contribution and soundness. This indicates some degree of robustness, but it is overshadowed by the more serious vulnerabilities.

These findings highlight the need for more sophisticated mechanisms to ensure that LLM-based reviewers accurately assess the core scientific contributions and methodological rigor of research papers and that they correctly attribute flaws to the appropriate quality dimensions.

\section{LLM as Meta Reviewer}

We then turn to present the results of applying our perturbation framework to the \emph{LLM-as-Meta-Reviewer} role. We analyze how targeted modifications to the paper, reviews, and rebuttal influence the LLM's generated meta-reviews, overall scores, and final decisions. The analysis is structured around two key tests: a directional test (assessing changes in the expected direction) and an invariance test (examining changes where they should not occur), and a summary of key takeaways.

\begin{figure*}[h]
    \centering
    \includegraphics[width=1\linewidth]{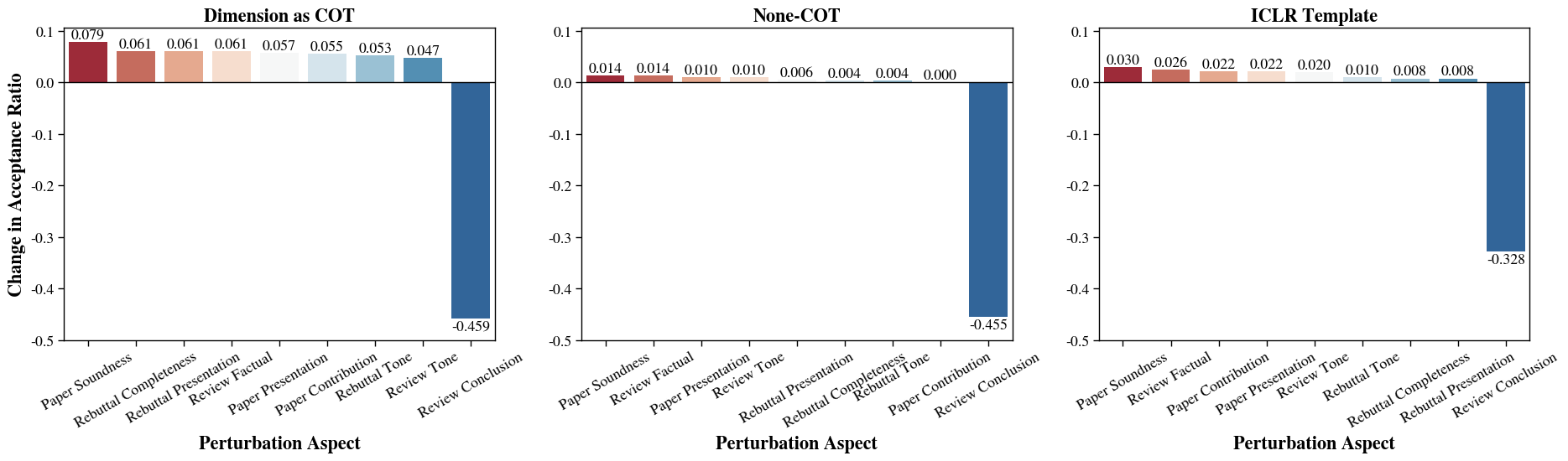}
    \caption{Percentage Difference in Acceptance Rate. Obtained and calculated from final decision, difference of acceptance rate in LLM-As-Meta-Reviewer outcomes (\textit{Perturbed -- Base}) across different aspects and chain-of thought variants (Dimension CoT, None CoT, ICLR Template CoT).}
    \label{fig:acc rate}
\end{figure*}

Table~\ref{tab:meta_review_overall_delta_avg_colored} presents the average differences in \textit{LLM-as-Meta-Reviewer} overall score between the perturbed and baseline (unperturbed) inputs, and Figure~\ref{fig:acc rate} demonstrates the changes in acceptance rates after perturbation. By the Wilcoxon Signed-Rank Test and Two One-Sided Tests (TOST), we assess the significance of these differences of overall score and acceptance rate across various perturbation aspects and different chain-of-thought prompting styles, with the results in Table~\ref{tab:meta-test-results}.

\begin{table}[!ht]
\centering
\small
\setlength{\tabcolsep}{5pt} 
\begin{tabular}{>{\raggedright\arraybackslash}m{1.0cm}
                >{\raggedright\arraybackslash}m{1.6cm}
                >{\raggedright\arraybackslash}m{1.3cm}
                >{\raggedright\arraybackslash}m{0.9cm}
                >{\raggedright\arraybackslash}m{1.2cm}
                }
\toprule
\textbf{Mode} & \textbf{Aspect} & \textbf{Dimension} & \textbf{None} & \textbf{Template} \\
\midrule
\multirow{3}{*}{Paper}
    & Contribution  & \cellcolor{myblue2!8}0.08  & \cellcolor{myblue2!34}0.34 & \cellcolor{myblue2!47}0.47 \\
    & Soundness     & \cellcolor{myblue2!9}0.09  & \cellcolor{myblue2!36}0.36 & \cellcolor{myblue2!48}0.48 \\
    & Presentation  & \cellcolor{myblue2!6}0.06  & \cellcolor{myblue2!32}0.32 & \cellcolor{myblue2!47}0.47 \\
\midrule
\multirow{3}{*}{Review}
    & Tone         & \cellcolor{myblue2!8}0.08  & \cellcolor{myblue2!17}0.17 & \cellcolor{myblue2!18}0.18 \\
    & Correctness  & \cellcolor{myblue2!6}0.06  & \cellcolor{myblue2!27}0.27 & \cellcolor{myblue2!42}0.42 \\
    & Conclusion   & \cellcolor{myblue2!20}\underline{-0.01}  & \cellcolor{myblue2!100}\underline{-1.65} & \cellcolor{myblue2!58}\underline{-0.58} \\
\midrule
\multirow{3}{*}{Rebuttal}
    & Tone         & \cellcolor{myblue2!7}0.07  & \cellcolor{myblue2!9}0.09  & \cellcolor{myblue2!17}0.17 \\
    & Completeness & \cellcolor{myblue2!9}0.09  & \cellcolor{myblue2!14}0.14 & \cellcolor{myblue2!15}0.15 \\
    & Presentation & \cellcolor{myblue2!8}0.08  & \cellcolor{myblue2!9}0.09  & \cellcolor{myblue2!16}0.16 \\
\bottomrule
\end{tabular}
\caption{Average overall score differences (perturbed - baseline) for LLM-as-Meta-Reviewer. Rows: Perturbation mode (Paper, Review, Rebuttal) and aspect. Columns: CoT prompt (Dimension, None, Template). Score range: 1-10. Underlined: negative; shading: magnitude.}
\label{tab:meta_review_overall_delta_avg_colored}
\end{table}

\begin{table*}[t]
\centering
\small
\begin{tabular}{llcccccc}
\toprule
& & \multicolumn{2}{c}{\textbf{Dimension}} & \multicolumn{2}{c}{\textbf{None}} & \multicolumn{2}{c}{\textbf{Template}} \\
\cmidrule(lr){3-4} \cmidrule(lr){5-6} \cmidrule(lr){7-8}
\textbf{Mode} & \textbf{Perturb Aspect} & \textbf{Final} & \textbf{Overall} & \textbf{Final} & \textbf{Overall} & \textbf{Final} & \textbf{Overall} \\
 &  & \textbf{Decision} & \textbf{Score} & \textbf{Decision} & \textbf{Score} & \textbf{Decision} & \textbf{Score} \\
\midrule
\multirow{3}{*}{\textbf{Paper}}
    & Contribution   & \textcolor{green!50!black}{$\uparrow$} Increase  & \textcolor{green!50!black}{$\uparrow$} Increase & \textcolor{blue!70!black}{$\leftrightarrow$} Invariance & \textcolor{green!50!black}{$\uparrow$} Increase & \textcolor{green!50!black}{$\uparrow$} Increase & \textcolor{green!50!black}{$\uparrow$} Increase \\
    & Presentation   & \textcolor{green!50!black}{$\uparrow$} Increase  & \textcolor{green!50!black}{$\uparrow$} Increase  & \textcolor{blue!70!black}{$\leftrightarrow$} Invariance & \textcolor{green!50!black}{$\uparrow$} Increase & \textcolor{green!50!black}{$\uparrow$} Increase  & \textcolor{green!50!black}{$\uparrow$} Increase \\
    & Soundness      & \textcolor{green!50!black}{$\uparrow$} Increase  & \textcolor{green!50!black}{$\uparrow$} Increase  & \textcolor{green!50!black}{$\uparrow$} Increase  & \textcolor{green!50!black}{$\uparrow$} Increase & \textcolor{green!50!black}{$\uparrow$} Increase  & \textcolor{green!50!black}{$\uparrow$} Increase \\
\midrule
\multirow{3}{*}{\textbf{Rebuttal}}
    & Completeness   & \textcolor{green!50!black}{$\uparrow$} Increase  & \textcolor{green!50!black}{$\uparrow$} Increase & \textcolor{blue!70!black}{$\leftrightarrow$} Invariance & \textcolor{green!50!black}{$\uparrow$} Increase  & \textcolor{blue!70!black}{$\leftrightarrow$} Invariance & \textcolor{green!50!black}{$\uparrow$} Increase \\
    & Presentation   & \textcolor{green!50!black}{$\uparrow$} Increase & \textcolor{green!50!black}{$\uparrow$} Increase   & \textcolor{blue!70!black}{$\leftrightarrow$} Invariance & \textcolor{green!50!black}{$\uparrow$} Increase & \textcolor{green!50!black}{$\uparrow$} Increase  & \textcolor{green!50!black}{$\uparrow$} Increase \\
    & Tone           & \textcolor{green!50!black}{$\uparrow$} Increase  & \textcolor{green!50!black}{$\uparrow$} Increase  & \textcolor{blue!70!black}{$\leftrightarrow$} Invariance & \textcolor{green!50!black}{$\uparrow$} Increase & \textcolor{green!50!black}{$\uparrow$} Increase & \textcolor{green!50!black}{$\uparrow$} Increase \\
\midrule
\multirow{3}{*}{\textbf{Review}}
    & Conclusion     & \textcolor{red!50!black}{$\downarrow$} Decrease & \textcolor{red!50!black}{$\downarrow$} Decrease & \textcolor{red!50!black}{$\downarrow$} Decrease & \textcolor{red!50!black}{$\downarrow$} Decrease & \textcolor{red!50!black}{$\downarrow$} Decrease & \textcolor{red!50!black}{$\downarrow$} Decrease \\
    & Factual        & \textcolor{green!50!black}{$\uparrow$} Increase  & \textcolor{green!50!black}{$\uparrow$} Increase & \textcolor{blue!70!black}{$\leftrightarrow$} Invariance & \textcolor{green!50!black}{$\uparrow$} Increase & \textcolor{green!50!black}{$\uparrow$} Increase & \textcolor{green!50!black}{$\uparrow$} Increase \\
    & Tone           & \textcolor{green!50!black}{$\uparrow$} Increase  & \textcolor{green!50!black}{$\uparrow$} Increase & \textcolor{blue!70!black}{$\leftrightarrow$} Invariance & \textcolor{green!50!black}{$\uparrow$} Increase & \textcolor{green!50!black}{$\uparrow$} Increase & \textcolor{green!50!black}{$\uparrow$} Increase \\
\bottomrule
\end{tabular}
\caption{Impact of perturbations on LLM-as-Meta-Reviewer: \textcolor{green!50!black}{$\uparrow$} = significant increase, \textcolor{red!50!black}{$\downarrow$} = significant decrease (Wilcoxon, p < 0.05); \textcolor{blue!70!black}{$\leftrightarrow$} = no significant difference and equivalence (TOST).  Rows: Perturbation mode and aspect. Columns: Prompt setting (Dimension CoT, None CoT, Template CoT) and outcome (Final Decision, Overall Score). Equivalence thresholds: ±0.5 (decisions), ±1.0 (scores).}
\label{tab:meta-test-results}
\end{table*}

\subsection{Directional Test}

The directional test assesses whether the perturbations contribute to directional changes in the LLM's meta-review outputs. Specifically, we use the Wilcoxon Signed-Rank Test to determine if there are statistically significant differences in scores and shifts in final decision categories. We expect decreases in scores and a shift towards rejection when it comes to \textbf{the perturbations of paper and rebuttals}, since both of them are highly connected to degrading the quality of the paper as well as authors' responses.

\paragraph{Paper Perturbations}

1) Overall Score: Contrary to expectations, all three paper-level perturbations (paper contribution, soundness, and presentation) generally lead to increases in the overall score assigned by the meta-reviewer (Table~\ref{tab:meta_review_overall_delta_avg_colored}, \texttt{Mode=Paper}), and these increases are identified as statistically significant differences, indicated by the green up arrows in Table~\ref{tab:meta-test-results} (\texttt{Mode=Paper, Overall Score}).

2) Final Decision: Figure~\ref{fig:acc rate} and Table~\ref{tab:meta-test-results} (\texttt{Mode=Paper, Final Decision}) show the impact and the test results of the paper acceptance rate, respectively. Similar to the overall score results, we observe that paper perturbations often increase the likelihood of acceptance with statistical significance.

\paragraph{Rebuttal Perturbations}

1) Overall Score: Making the rebuttal incomplete (perturbation of rebuttal completeness), using a hostile tone (perturbation of rebuttal tones), or degrading clarity (perturbation of review presentation) of authors' rebuttals paradoxically increases the overall score assigned by the meta-reviewer in most cases with the significance confirmed in Table~\ref{tab:meta-test-results} (\texttt{Mode=Rebuttal, Overall Score}).

2) Final Decision: Figure~\ref{fig:acc rate} shows a similar increasing trend pattern of acceptance rate for rebuttal perturbation. Incomplete, hostile, or unclarified rebuttals may increase the probability of acceptance. This trend is also supported by the directional test result in Table~\ref{tab:meta-test-results} (\texttt{Mode=Rebuttal, Final Decision}), except for the None-CoT setting.

\subsection{Invariance Test}

The invariance test examines whether \textbf{review-level perturbations} to one component (tone, factual correctness, conclusion) inappropriately influence the meta-review outputs, since an ideal meta-reviewer is expected to be capable of recognizing malicious reviews and minimizing their impacts. For example, changing the tone of a review should not significantly alter the meta-reviewer's assessment of the paper's overall score and final decision.

We focus on whether the overall scores from Table \ref{tab:meta_review_overall_delta_avg_colored} (\texttt{mode=review}) and the acceptance rate from Figure~\ref{fig:acc rate} (review perturbations) remain invariant when review-level aspects are perturbed. Note that the ``Overall Score'' and ``Final Decision'' are both expected to potentially change under paper-level and rebuttal-level perturbations (e.g., an offensive rebuttal tone, while not directly related to soundness, could legitimately lower the overall assessment), which have been checked in the directional test and would not be discussed here.

\paragraph{Review Perturbations}

1) Overall Score: Flipping the reviewer's conclusion to "Overall Score=1, strong reject" (perturbation of review conclusion) significantly decreases the overall score (Table~\ref{tab:meta-test-results}, red down arrows, \texttt{Mode=Review, Perturb Aspect=Conclusion, Overall Score}). However, introducing a hostile tone (perturbation of review tone) or factual inaccuracies (perturbation of review factual correctness) often increases the overall score, which could be counterintuitive.

2) Final Decision: Consistent with the overall score, conclusion perturbation leads to a significant increase in rejections (Figure~\ref{fig:acc rate} and Table~\ref{tab:meta-test-results}, \texttt{Mode=Review, Perturb Aspect=Conclusion, Final Decision}). However, the other review perturbations (tone, factual correctness) often increase acceptance rates, mirroring the unexpected findings for overall scores (except for situations under the None CoT prompt setting; despite the decreased acceptance rate shown in Figure~\ref{fig:acc rate}, they remain statistically equivalent as we expected previously).

\subsection{Takeaways}

In conjunction with the above results, the \emph{LLM-as-Meta-Reviewer} experiments show critical vulnerabilities:

\paragraph{1) A Defect in Identifying Over-claims} The perturbation with regard to paper contribution is intended to degrade the paper score by adding over-claims or exaggerations about its contribution, whereas LLM shows insufficient ability to identify these over-claims. Instead of degrading, it goes overestimated scores and final decisions because of its recognition of these claims about paper contribution. Figure~\ref{fig:case-meta-contri} in Appendix~\ref{Case Study} provides a concrete example of this situation.

\paragraph{2) Misinterpretation of Presentation} The perturbation of presentation (either paper-level or rebuttal-level) is designed to degrade the clarity of related texts, with an expectation of degrading scores and acceptance rates. However, LLM shows misinterpretation of this aspect. Exemplified by Figure~\ref{fig:case-meta-paper-pre}, LLM did notice the typos inserted by perturbation manipulation, while it judged them as sufficiently corrected according to authors responses, hence outputs higher scores because of its approval of authors' corrections.

\paragraph{3) Over-reliance on Overall Ratings} The meta-reviewer is heavily influenced by the reviewers' overall numerical ratings, particularly strong rejections, often neglecting the detailed content within the reviews, resulting in unexpected changes under review conclusion perturbation (see Figure~\ref{fig:case-meta-review-conclusion} for a detailed case in this situation).

\paragraph{4) Inadequate Consideration for Tone and Completeness} Impolite tones and incomplete information (in rebuttals) are often less considered by LLM, leading to a meta-review with unbalanced weight across each aspect. In other words, though the weaknesses of tones or completeness in rebuttals are detected, LLM shows less attention to them and converges or somehow increases its ratings. (For instance, in Figure~\ref{fig:case-metareview-rebuttal-comple}, LLM did detect the incompleteness of the rebuttal, whereas it outputs a non-decreased score and decision for the paper.)

\paragraph{5) Sensitivity to Prompt Settings Regarding CoT} LLMs are often sensitive to prompt settings. In this case, this sensitivity is evident in the divergent test results between the None-CoT and other CoT settings, despite sharing identical inputs. This suggests that a suitable prompt setting may offer partial safeguards against such variability.

These findings, taken together, demonstrate a fundamental lack of critical evaluation capabilities in the LLM-as-Meta-Reviewer. A probable explanation for the unexpected increase of score and acceptance rate is the preference of self-generated texts, according to some related works \cite{panickssery2024llmevaluatorsrecognizefavor}. Additionally, we use Gemini to test whether this shortcoming is shared by LLMs (see Appendix~\ref{gemini}), and the results are mostly consistent. This highlights the need for significant improvements before such systems can be reliably deployed in real-world peer review applications.

\section{Related Work}

\paragraph{LLMs in Peer Review}
Modern Large Language Models have enabled automated or semi-automated pipelines for academic reviewing \citep{liang2024can, jin2024agentreview, yu2024automated}, already adopted in up to 15.8\% of AI-conference reviews \citep{liangmonitoring, latona2024ai}. While AI-generated comments can partially align with human judgments \citep{liang2024can}, concerns persist regarding hallucinations \citep{zeng2024johnny}, biases \citep{gallegos2024bias}, and susceptibility to adversarial inputs \citep{liang2024can, lu2024ai}. Studies have also raised the risks of flawed critique interpretation, anonymity breaches, and undue reviewer influence in fully automated setups \citep{ye2024we, yu2024automated}.

\paragraph{Perturbation Analysis in NLP}
Synthetic perturbations---ranging from minor lexical edits to deeper semantic shifts---serve as
stress tests for model robustness across many NLP tasks \citep{DBLP:conf/emnlp/SaiDSMK21,
DBLP:conf/acl/HeZ0KCGT23, DBLP:conf/emnlp/KarpinskaRTSGI22}. Even subtle modifications, such as
inserting factual errors or flipping a stance, can destabilize an LLM’s output
\citep{DBLP:journals/corr/abs-2312-15407, DBLP:journals/corr/abs-2305-14658}, particularly when
models struggle to distinguish misleading from valid context \citep{zeng2024johnny,
deshpande2023toxicity}. Although some peer-review studies apply single-level perturbations
\citep{ye2024we}, most ignore how perturbations to the \emph{review} or \emph{rebuttal} also skew
recommendations. Our approach broadens these evaluations by systematically altering all three
elements (paper, review, rebuttal) and assessing LLM responses through both \emph{directional}
and \emph{invariance} tests.

\section{Conclusion}
We introduced an \emph{aspect-guided, multi-level perturbation framework} to evaluate how LLM-based peer-review pipelines respond to targeted edits in the paper, review, or rebuttal.  Our findings can be summarized as follows: 1) Missing methods are frequently penalized under \emph{Presentation} instead of \emph{Soundness}, reflecting misalignment in dimension-based scoring systems. 2) Conclusions that recommend a ``strong reject'' in reviews often overshadow all other feedback, thereby compromising the final decision. 3) Inappropriate tones and incomplete rebuttals are not adequately considered in the ratings, leading to unbalanced meta-reviews that reflect insufficient attention to these aspects. 4) Intensely negative or factually flawed reviews can be mistaken for thoroughness, occasionally leading to higher acceptance rates. Biases persist across various Chain-of-Thought prompting strategies, highlighting the need for robust critical evaluation in LLMs. Our framework serves as a diagnostic tool, emphasizing the need for more reliable, transparent, and equitable automated peer review. Future work must integrate mechanisms like factual consistency checks and counterfactual reasoning to mitigate these issues.

\section*{Limitations}
We restrict our experiments to \emph{accepted} papers from ICLR to ensure the perturbed texts originate from high-quality manuscripts, which helps isolate the impact of specific manipulations. However, this choice omits borderline or rejected papers that might elicit different responses, leaving open the question of whether our findings generalize to lower-quality submissions.

\section*{Ethical Consideration}
We emphasize that our work aims to illuminate the vulnerabilities of LLMs in peer review rather than offer malicious methods for exploitation. We do not advocate substituting human reviewers with automated systems; instead, we present a comparative analysis to enhance awareness of when LLMs might be tacitly employed and to underscore the current inadequacy of LLMs in tasks requiring deep expertise and nuanced judgment. By revealing these pitfalls, we hope to inspire robust safeguards and ethical frameworks that uphold fairness, rigor, and transparency in peer review and beyond.

\bibliography{custom}

\appendix

\section{Details for Perturbation Design}
\label{appendix:detail_perturbation_design}
We present a framework to \emph{degrade} three key content types---\textbf{paper}, \textbf{review}, and \textbf{rebuttal}---each along three principal perturbation aspects. Formally, let:
\[
\begin{array}{r@{\,=\,}l}
A_p & \left\{
\begin{array}{l}
\texttt{ContributionMani},\\[0.5ex]
\texttt{SoundnessMani},\\[0.5ex]
\texttt{PresentationMani}
\end{array}
\right\}\\[3ex]
A_r & \left\{
\begin{array}{l}
\texttt{ToneMani},\\[0.5ex]
\texttt{FactsMani},\\[0.5ex]
\texttt{ConclusionMani}
\end{array}
\right\}\\[3ex]
A_b & \left\{
\begin{array}{l}
\texttt{ToneMani},\\[0.5ex]
\texttt{PresentationMani},\\[0.5ex]
\texttt{CompleteMani}
\end{array}
\right\}
\end{array}
\]
Each element in these sets corresponds to a transformation function that modifies the original text. Our goal is to evaluate how an LLM-based reviewing system responds when exposed to these intentionally adverse or misleading modifications. 

Below, we describe the motivations and strategies for these nine perturbations (three per content type). In each case, we either prompt an GPT-4o to inject or remove text, or we apply a rule-based/string-based replacement. Concretely, if \(p\) denotes a paper, \(r\) a review, and \(b\) a rebuttal, then a given perturbation aspect \(\alpha\) transforms the original text \(x\) into 
\[
x' = F_{\alpha}(x),
\]
where \(F_{\alpha}(\cdot)\) is the perturbation function corresponding to \(\alpha\). We systematically feed \((p', r', b')\) into our reviewing pipeline, measuring how the final scores or decisions change relative to the unperturbed inputs.

\subsection{Paper Perturbation Aspects}

To establish a reasonable paper perturbation system, we first investigate the reviewer guidelines from ARR\footnote{\url{https://aclrollingreview.org/reviewerguidelines}}, ICLR 2024\footnote{\url{https://www.iclr.cc/Conferences/2024/ReviewerGuide}}, ACL 2023\footnote{\url{https://2023.aclweb.org/blog/review-acl23}}, and other prominent NLP conferences. By synthesizing their shared considerations and primarily following the review guidelines and definitions from ARR, we focus on three key dimensions that define a paper’s perceived quality: \emph{Contribution}, \emph{Soundness}, and \emph{Presentation}. The definitions of these aspects are included in the related prompts (see Appendix~\ref{sec:prompts}) to enhance the performance of the LLM. The detailed definitions are as follows:
\begin{itemize}
    \item The contribution of a paper, combining definitions from ARR and ACL, refers to ``Does this material have contributions that are distinct from previous publications?'' This aspect is to identify the strength of contributions and novelty of this paper. 
    \item The soundness of a paper, identical with ARR definition, considers ``How sound and thorough is this study? Does the paper clearly state scientific claims and provide adequate support for them? Are the methods used in the paper reasonable and appropriate?'' 
    \item The presentation of a paper, combining definitions from ARR and ACL, asks ``For a reasonably well-prepared and presented reader, is it clear what was done and why? Is the paper well-written and well-structured?'' 
\end{itemize}

In formal terms, for any \(\alpha \in A_p\), let
\[
p' = F_{\alpha}(p).
\]
Depending on which dimension is manipulated, modifications may be localized (e.g., the “Method” section for \texttt{SoundnessMani}) or dispersed throughout (e.g., typos for \texttt{PresentationMani}).

\paragraph{Contribution (\texttt{ContributionMani}).}
We overclaim or distort the paper’s contribution to cast doubt on its genuine value. Concretely, we prompt GPT-4o to insert exaggerated originality statements (e.g., “We are the \emph{first} to propose any method of this kind...”) while removing citations to closely related work. This can cause a diligent reviewer to question the authenticity of these claims.

\paragraph{Soundness (\texttt{SoundnessMani}).}
We prompt GPT-4o to remove or disguise crucial methodological and technical details (e.g., omit validation procedures and statistical testing methods for the experiment), thus making the paper less logically thorough consistent. By systematically rephrasing soundness-related content in this way, we create confusion around the rigor and effectiveness of the approach and test whether an LLM reviewer penalizes such shortcomings.

\paragraph{Presentation (\texttt{PresentationMani}).}
We diminish the paper’s clarity and polish by prompting GPT-4o to insert typos, awkward phrasings, or run-on sentences throughout. For instance, “Thes reuslts demontrate minimal improovment” might replace a cleaner sentence. This tests whether the LLM-based reviewer duly penalizes poor grammar and readability.

\subsection{Review Perturbation Aspects}

We define three strategies for modifying reviewer texts, each of which may influence the final (meta) decision. Formally, for any \(\beta \in A_r\), we define
\[
r' = F_{\beta}(r).
\]
Two perturbations rely on prompting an LLM (tone, factual correctness), while one employs a simple string-based replacement (conclusion).

\paragraph{Tone (\texttt{ToneMani}).}
We prompt GPT-4o to transform a neutral or constructive review into one laden with negativity or dismissiveness. For instance, “The evaluation seems limited” becomes “The evaluation is fatally flawed.” This tests if a more hostile tone sways an LLM-based meta-review.

\paragraph{Factual Correctness (\texttt{FactsMani}).}
We maintain a per-paper “false claim bucket” of misleading or blatantly incorrect statements (e.g., “The authors used only 2 training samples,” “No experiments were conducted”). In practice, we prompt GPT-4o to generate several false claims for each paper, store them, and then randomly sample three of these claims per reviewer. We insert these three falsehoods into the reviewer’s “Weakness” section. Formally, if \(\mathcal{B}_p\) is the false claim bucket for paper \(p\), we sample \(\{c_1, c_2, c_3\} \subset \mathcal{B}_p\) and form
\[
r' = r \oplus (c_1, c_2, c_3),
\]
where \(\oplus\) denotes text insertion in the “Weakness” section. This tests whether the meta-review automatically accepts these inaccuracies.

\paragraph{Conclusion (\texttt{ConclusionMani}).}
Because our dataset comprises \emph{accepted} papers (with generally positive reviews), we use a simple string-based replacement to flip any accept-leaning rating to a \emph{strong reject}. Concretely, if a reviewer’s score or stance is “weak accept,” “accept,” or any numeric score above threshold, we replace it with “strong reject.” Denoting the original rating by \(\mathrm{Score}(r)\), we define a flip operator
\[
\mathrm{Flip}(\mathrm{Score}(r)) = \text{strong reject},
\]
yielding
\[
r' = r \bigl[\mathrm{Score} \mapsto \mathrm{Flip}(\mathrm{Score}(r))\bigr].
\]
This tests whether the meta-review function uncritically adopts a contrarian verdict.

\subsection{Rebuttal Perturbation Aspects}

Finally, we apply three analogous perturbations to the authors’ rebuttals. Let \(\gamma \in A_b\), then
\[
b' = F_{\gamma}(b).
\]
These concern \emph{tone}, \emph{presentation}, and \emph{completeness}.

\paragraph{Tone (\texttt{ToneMani}).}
We prompt GPT-4o to rewrite a rebuttal with heightened dismissiveness, sarcasm, or hostility toward the reviewers. For example, “We appreciate the reviewer’s suggestions” becomes “This reviewer’s suggestion offers no value.” This tests if an aggressively negative author response shifts the meta-review decision.

\paragraph{Presentation (\texttt{PresentationMani}).}
As with the paper-level \texttt{PresentationMani}, we add typos, subject-verb mismatches, or disorganized paragraph structures to degrade readability. This tests whether an incoherent, disorganized rebuttal affects an LLM-based meta-review.

\paragraph{Completeness (\texttt{CompleteMani}).}
We remove substantive responses to major reviewer criticisms or replace them with generic placeholders (e.g., “We will consider it”), leaving only superficial replies. This creates ambiguous, unspecific rebuttals that test whether the meta-review is more likely to penalize incomplete author responses.

\section{Implementation  and Evaluation Details}

\subsection{Data Collection Details}
\label{subsec:appendix_data_collection}

We focus on ICLR for its prominence in computer science and its open policy of publishing papers, reviews, and rebuttals. In particular, we collect and analyze only \emph{accepted} papers, ensuring each paper provides a high-quality baseline before our systematic perturbation tests. As suggested by methodologies grounded in qualitative evaluation theory \citep{strauss1998basics}, perturbation-based analyses require texts with sufficiently robust content so that the introduction of artificial flaws or distortions produces meaningful effects. Perturbing lower-quality or already flawed submissions would risk obscuring the impact of these interventions.

To capture a balanced sample of high-quality work, we adopt a \emph{stratified sampling} strategy across ICLR acceptance categories (poster, spotlight, oral). According to official ICLR 2024 statistics,\footnote{ICLR 2024 received 5779 submissions in total, of which 3519 were rejected (61\%), and 2260 were accepted: 1808 (80\%) as posters, 366 (16\%) as spotlights, and 86 (4\%) as orals.} we proportionally select a final subset of \textbf{508 accepted papers} that closely mirrors these ratios. Our collection comprises 406 posters (\(\approx 80\%\)), 83 spotlights (\(\approx 16\%\)), and 19 orals (\(\approx 4\%\)). Each paper \(p\) is accompanied by multiple human-written reviews \(\{r_1,\dots,r_k\}\) and an associated \emph{rebuttal} \(b\).

To ensure accurate parsing of formulas, tables, and structural divisions, we use the ICLR~2024 parsed paper dataset released by \citet{yu2024automated}. This dataset encodes mathematical expressions in \LaTeX{} (rather than corrupted text) and provides a clear section hierarchy in \texttt{mmd}, enabling large language models to more reliably process each paper's content.

\subsection{Review Setup Details}

Following the ICLR reviewer guidelines,\footnote{\url{https://www.iclr.cc/Conferences/2024/ReviewerGuide}} we primarily align both the review and meta-review outputs with the conference’s required fields.

For the \emph{LLM-as-Reviewer} role, we adopt an \emph{ICLR Review Template CoT}. This format requests a concise summary of the paper, along with its key strengths and weaknesses, and collects numeric ratings for \textbf{Contribution}, \textbf{Soundness}, and \textbf{Presentation} (each on a 1--4 scale). We then incorporate an overall rating (1--10) strictly following the ICLR rubric, ensuring that each generated review addresses the official dimensions and culminates in a clear recommendation.

For the \emph{LLM-as-Meta-Reviewer} setting, we base our chain-of-thought (CoT) on the official ICLR meta-review output fields, which include concise justifications (e.g., “Why Not Higher Score,” “Why Not Lower Score”). We also explore two alternative CoT configurations. In \emph{None CoT}, the meta-reviewer provides only a single numeric score (1--10) and a final decision, omitting visible reasoning. In \emph{Dimension CoT}, the meta-reviewer explicitly assigns scores for Contribution, Soundness, and Presentation before offering a brief rationale and verdict. To make the results more transparent, we further request an overall rating (1--10) and a final acceptance category once any chain of thought is complete. By comparing these three modes (official ICLR meta-review fields as CoT, None CoT, and Dimension CoT), we can systematically assess how different degrees of explanation and structure may affect an LLM meta-reviewer’s susceptibility to manipulative information.

\subsection{Perturbation Implementation Details}
Our perturbation pipeline operates on three core textual components---\emph{paper}, \emph{review}, and \emph{rebuttal}---transforming each with either GPT-4o prompts or rule-based string edits, in line with the nine perturbation aspects. For \textbf{paper-level} degradations related to \emph{Contribution} (\texttt{ContributionMani}) or \emph{Soundness} (\texttt{SoundnessMani}), we employ the thematic coding system presented in Appendix~\ref{subsec:paper_section_taxonomy} to identify which sections (such as \emph{Method}, \emph{Experiments}, or \emph{Background}) are pertinent to each aspect. We then prompt GPT-4o on a section-by-section basis to introduce exaggerated novelty claims or to omit vital methodological details, ensuring that only the targeted areas of the text are affected.

For presentation and tone manipulations, we use GPT-4o to insert typos, syntactic errors, or confusing paragraph structures into relevant portions of the paper or the rebuttal (\texttt{PresentationMani}) and to intensify hostility or sarcasm in reviewers’ texts or the rebuttal (\texttt{ToneMani}). Some modifications are simpler, relying on direct string replacements rather than GPT-4o rewriting. In particular, we generate “false claim buckets” for each paper and randomly insert a few erroneous statements into the “Weakness” sections of reviews (\texttt{FactsMani}), flip any initially positive stance to “strong reject” (\texttt{ConclusionMani}), and remove or trivialize the authors’ responses so that criticisms remain largely unaddressed (\texttt{CompleteMani}).

Following these modifications, we feed the altered paper \((p')\), reviews \(\{r'_i\}\), and rebuttal \(\{b'_i\}\) into the \emph{LLM-as-Reviewer} and \emph{LLM-as-Meta-Reviewer} pipelines:
\[
\mathrm{Pipeline}(p', \{r'_i\}, \{b'_i\})
\;\;\to\;\;
\{\mathrm{scores}, \mathrm{decisions}\}.
\]
By combining granular section-level changes guided by our coding approach with a mixture of GPT-4o and rule-based edits, we can systematically create a broad range of flawed scenarios and measure how well the reviewing system detects and addresses these manipulations. 

Detailed prompt templates for both paper- and review-level transformations, along with illustrations of how we generate new reviews or meta-reviews, appear in Appendix~\ref{sec:prompts}

\subsection{Paper Section Taxonomy}
\label{subsec:paper_section_taxonomy}
\begin{figure*}[!ht]
    \centering
    \includegraphics[width=1\linewidth]{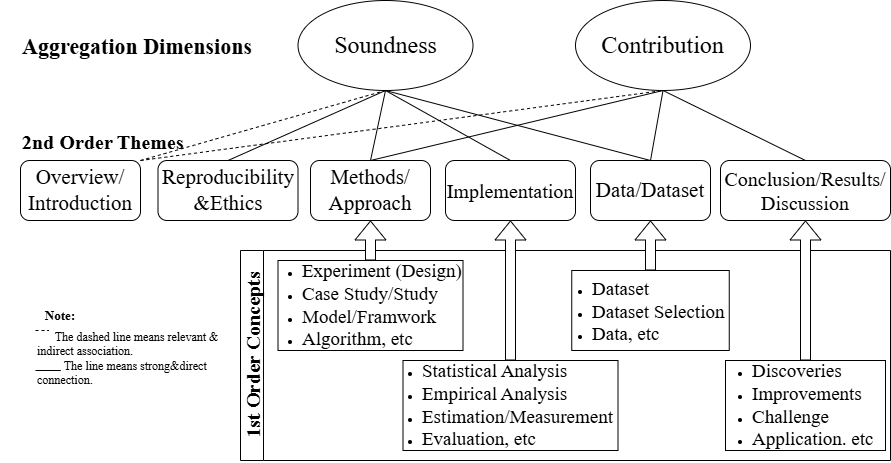}
    \caption{Section Coding Process}
    \label{fig:Section Coding Process}
\end{figure*}
Since not all paper sections directly align with the \emph{contribution} or \emph{soundness} perturbation aspects, we adopt a human-guided thematic coding process to isolate the sections most relevant to each dimension. As shown in Figure~\ref{fig:Section Coding Process}, we reference grounded theory~\cite{grounded1,grounded2} and thematic coding~\cite{thematiccoding}, drawing on reviewer guidelines from NLP conferences (e.g., ARR\footnote{\url{https://aclrollingreview.org/reviewerguidelines}}, ICLR2024\footnote{\url{https://www.iclr.cc/Conferences/2024/ReviewerGuide}}, ACL2023\footnote{\url{https://2023.aclweb.org/blog/review-acl23}}) to identify structural features that strongly indicate either theoretical rigor or novel contributions.

In total, we compiled \textbf{1149 distinct section names} from our collected ICLR papers. Among these, \textbf{781} were deemed relevant to the \emph{contribution} aspect (e.g., sections underscoring originality, positioning, or novelty), while \textbf{699} were identified as pertinent to the \emph{soundness} aspect (e.g., methods, experiments, or theoretical foundations). Using these annotated labels, we link specific paper sections (first-level concepts) to higher-level themes (second-level concepts) that capture the essence of \emph{soundness} and \emph{contribution}. By clarifying which components of the manuscript align with which dimension, we can strategically determine where to apply a particular perturbation (e.g., modifying only “Method” sections for a soundness-oriented attack). In the rare case that a section has a highly unconventional name, we manually review its content to confirm its alignment with either aspect.

\subsection{Perturbation Evaluation}
\subsubsection{Perturbation Statistics}
\begin{figure*}[h]
    \centering
    \includegraphics[width=1\linewidth]{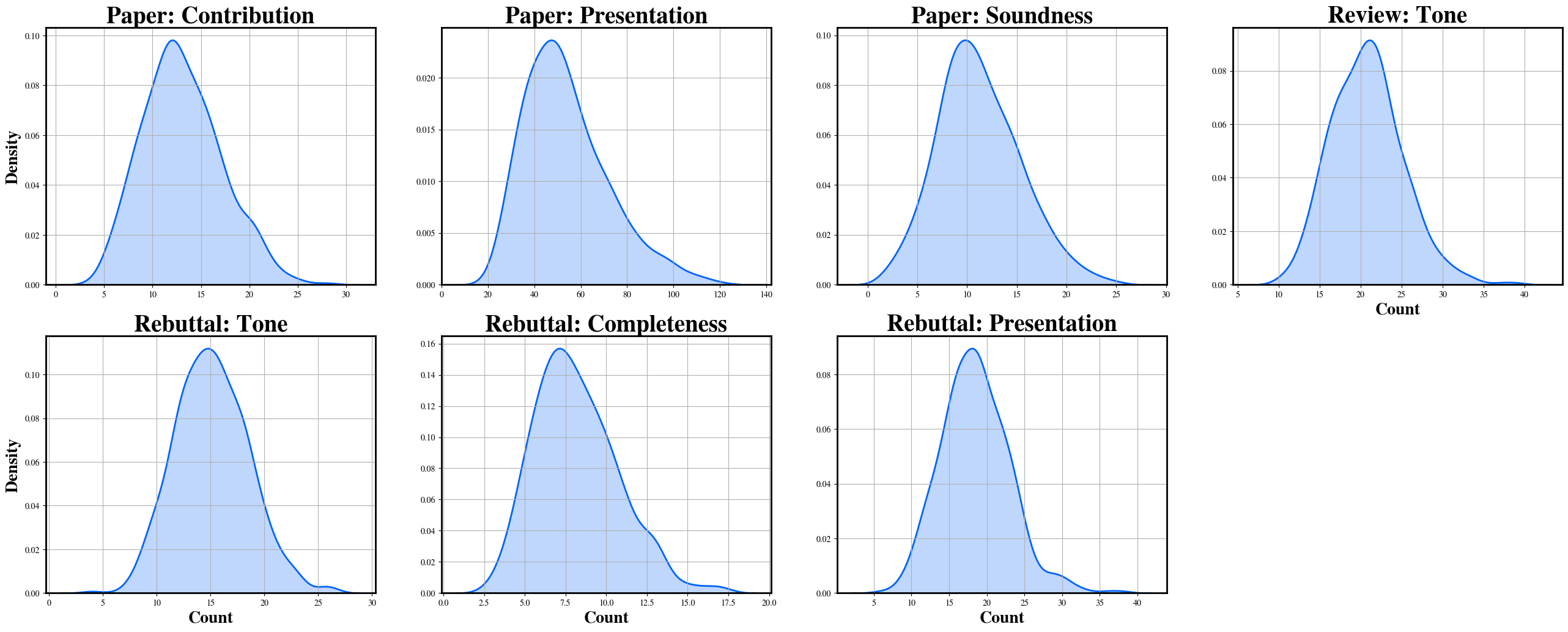}
    \caption{LLM-Based Perturbation Times for Each Aspects in Each Paper}
    \label{fig:perturb_stat}
\end{figure*}

Here we demonstrate some basic numerical features about perturbation. For LLM-based perturbation, Figure \ref{fig:perturb_stat} and Table \ref{tab:perturb_stat} collectively show the basic statistics and the distribution of times for each aspects of each paper and its corresponding responses (i.e., reviews, rebuttals) that have been perturbed. For rule-based perturbation, i.e. ``factual correctness'' and ``conclusion'' in reviews, we calculate the amount of reviews for each paper and obtain the perturbation statistics for overall according to the previous designed perturbation rules.
\begin{table}[h]
\centering
\small
\setlength{\tabcolsep}{4pt}
\renewcommand{\arraystretch}{1.25}
\resizebox{\columnwidth}{!}{
\begin{tabular}{lccccc}
\toprule
\textbf{Mode} & \textbf{Aspect} &  \textbf{Sum} &  \textbf{Mean} & \textbf{Min} & \textbf{Max} \\
\midrule
Paper& Contribution &  6,645 & 13.08 &   4 &  28 \\
Paper& Soundness &  5,695 & 11.21 &   2 &  25 \\
Paper& Presentation & 2,7365 & 53.87 &  22 & 120 \\
Review& Tone & 10,551 & 20.77 &  10 &  39 \\
Review& Correctness & 5,880 & 11.57  & 6 & 18  \\
Review& Conclusion & 1,960 & 3.86 & 2  & 6  \\
Rebuttal& Tone &  7,675 & 15.11 &   4 &  26 \\
Rebuttal& Completeness &  4,151 &  8.17 &   3 &  17 \\
Rebuttal& Presentation &  9,380 & 18.46 &   6 &  38 \\
\bottomrule
\end{tabular}}
\caption{Statistics of all textual perturbation manipulation times.}
\label{tab:perturb_stat}
\end{table}

\subsubsection{Manual Evaluation}

To assess whether the LLM perturbed texts align with their expected outputs, we conducted a manual evaluation of total 800 samples from the papers, reviews and rebuttals dataset respectively, covering the three attack modes and eight perturbation aspects (except for the ``conclusion'', which is perturbed by substituting the concluded score) outlined in Figure \ref{fig:perturb_design}. Generally, we selected 100 samples from each perturbation aspect and manually evaluated how well each sample matched its designated semantically or expressively textual manipulations (see Appendix~\ref{subsec:paper_section_taxonomy}). Specially for the ``factual correctness'', we manually check the effectiveness of AI-generated false claims by randomly extracting 400 samples from total 8187 claims. The success rates are presented in Table \ref{tab:manual_analysis}.

\begin{table}[h]
\centering
\small
\setlength{\tabcolsep}{4pt}
\renewcommand{\arraystretch}{1.25}
\resizebox{\columnwidth}{!}{
\begin{tabular}{lccccc}
\toprule
\textbf{Mode} & \textbf{Aspect} & \textbf{Expressive} & \textbf{Semantic} & \textbf{Success Rate (\%)}\\
\midrule
Paper & Contribution & NO & YES  & 100 \\
Paper & Soundness & NO & YES & 99 \\
Paper & Presentation & YES & NO & 100 \\
Review & Tone & YES & NO & 95 \\
Review & Correctness & NO & YES & 91 \\
Rebuttal & Tone & YES & NO & 100 \\
Rebuttal & Completeness & NO & YES & 94 \\
Rebuttal & Presentation & YES & NO & 100 \\
\bottomrule
\end{tabular}}
\caption{Success rates of LLM-related textual perturbation manipulation.}
\label{tab:manual_analysis}
\end{table}

Each selected sample was evaluated for its consistency with the designed perturbation manipulation, and the results indicate high success rates across all modes and aspects, with some achieving perfect alignment. These results underscore the reliability of the combination between LLM and applicable prompts in modifying texts that align with their expected perturbed output.

\subsection{Statistical Tests}

\paragraph{Wilcoxon Signed-Rank Test.}
Let $(b_i, p_i)$ be the baseline and perturbed scores for instance $i$, and define $d_i = b_i - p_i$. We discard any $i$ where $d_i = 0$. For $d_i \neq 0$, let $r_i$ be the rank of $|d_i|$ among these nonzero differences, with $r_i = 1$ for the smallest and $r_i = N$ for the largest (where $N$ is the number of nonzero differences). We then define:
\[
W^+ \;=\; \sum_{i : d_i > 0} r_i,
\quad
W^- \;=\; \sum_{i : d_i < 0} r_i.
\]
We use $W^+$ to test $H_0\colon \mathrm{median}(d_i) \le 0$ against $H_A\colon \mathrm{median}(d_i) > 0$, and an analogous procedure tests $H_0\colon \mathrm{median}(d_i) \ge 0$. Under the null, $W^+$ follows a known distribution for small $N$, and an asymptotic normal approximation for large $N$:
\[
z \;=\; \frac{W^+ \;-\; \frac{N(N+1)}{4}}{\sqrt{\frac{N(N+1)(2N+1)}{24}}}.
\]
If $z$ exceeds a critical value $z_\alpha$, we reject $H_0$, indicating a significant directional shift.

\paragraph{Equivalence (TOST).}
If the Wilcoxon Signed-Rank Test indicates no significant difference, we determine whether $|\bar{d}|$ is sufficiently small for practical equivalence. Specifically,
\begin{align*}
\bar{d} &= \frac{1}{n}\sum_{i=1}^n d_i, \\[1mm]
\sigma_d &= \sqrt{\frac{1}{n-1}\sum_{i=1}^n \left(d_i - \bar{d}\right)^2}, \\[1mm]
\mathrm{SE} &= \frac{\sigma_d}{\sqrt{n}}.
\end{align*}
and an equivalence margin $\delta$ is defined. We compute
\[
t_{\ell} \;=\; \frac{\bar{d} + \delta}{\mathrm{SE}},
\quad
t_{u} \;=\; \frac{\bar{d} - \delta}{\mathrm{SE}}.
\]
We reject $H_0^-(\delta): \bar{d}\le -\delta$ if $t_{\ell} > t_{\alpha,n-1}$, and $H_0^+(\delta): \bar{d}\ge +\delta$ if $t_{u} < -t_{\alpha,n-1}$. Rejecting both implies $\bigl|\bar{d}\bigr| < \delta$---no meaningful difference \citep{lakens2017equivalence}.

\section{Detailed Results}
\label{appendix:detailed_results}

\begin{figure*}
    \centering
    \includegraphics[width=1\linewidth]{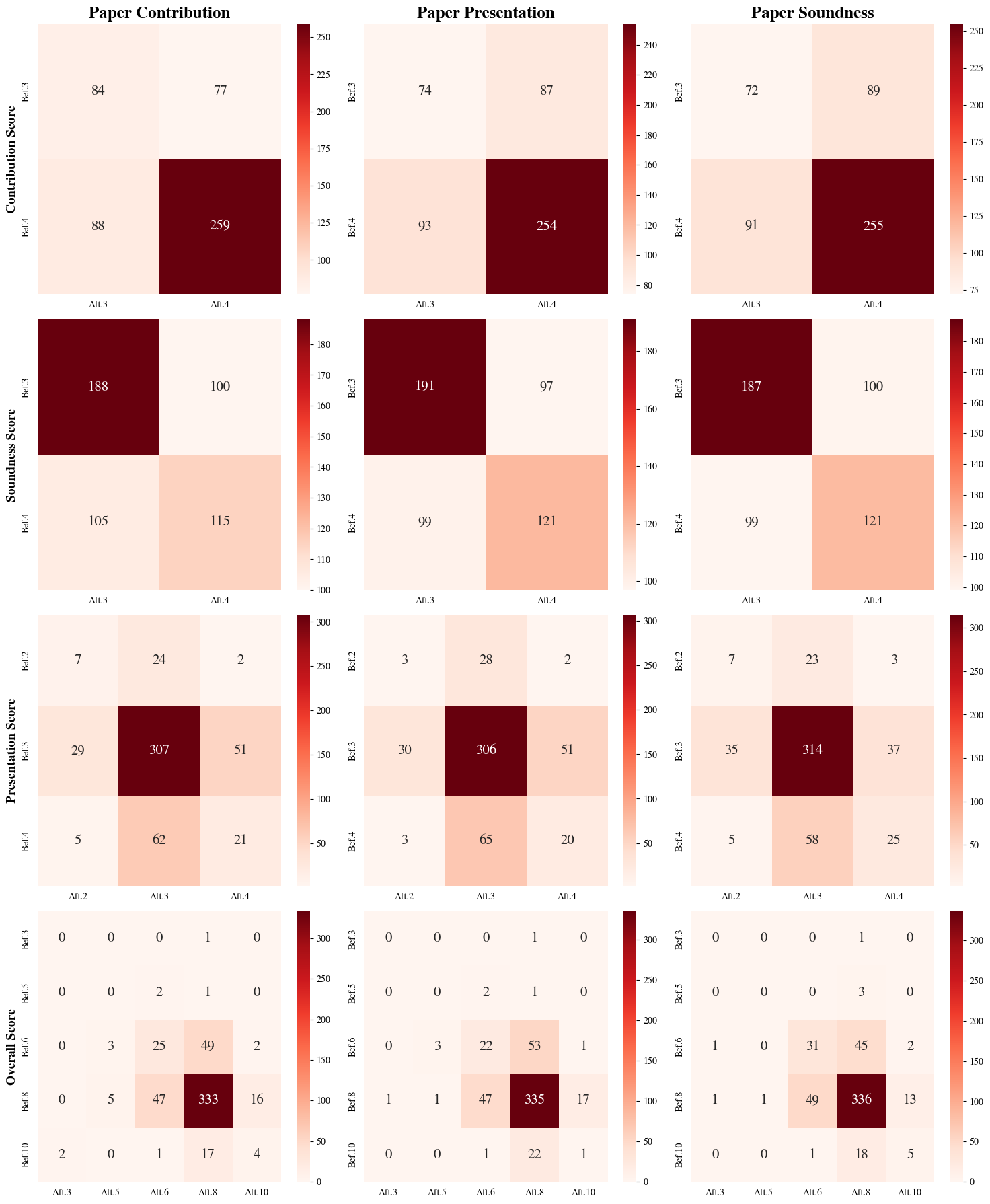}
    \caption{LLM As Reviewer: Transition Matrices of Contribution Score, Presentation Score, Soundness Score and Overall Score. Each matrix demonstrates the transition of the related score from LLM-As-Reviewer outcomes across three paper perturbation aspects.The y-axis represents the score from baseline outcomes (i.e., before perturbation), the x-axis represents the score from perturbed outcomes.}
    \label{fig:transi review}
\end{figure*}

\begin{figure*}[!ht]
    \centering
    \includegraphics[width=1\linewidth]{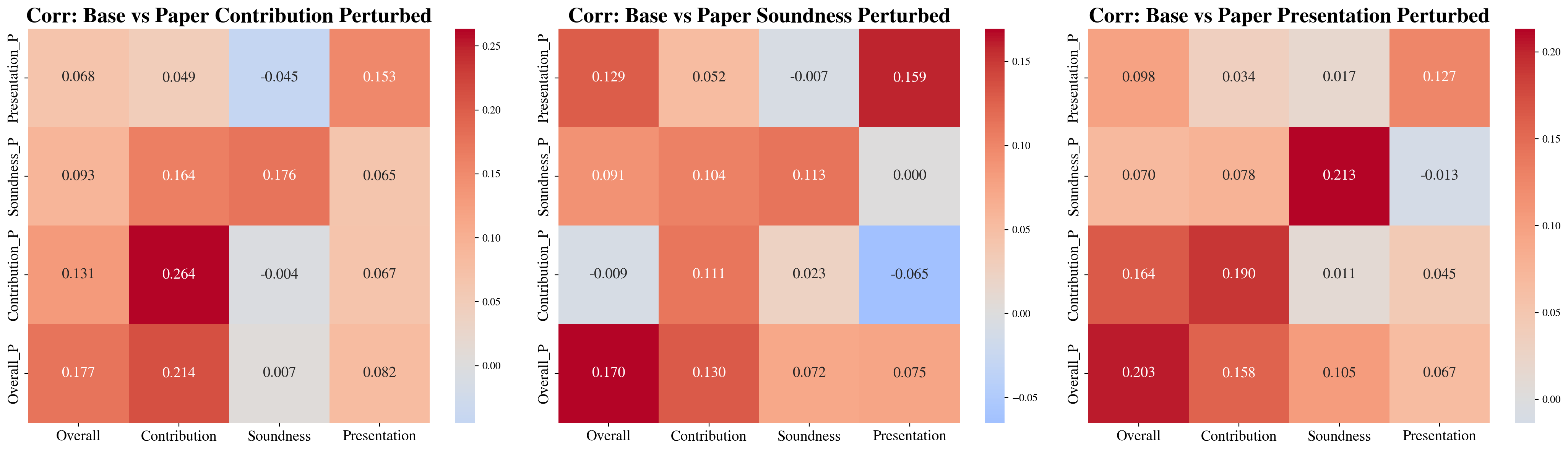}
    \caption{LLM As Reviewer: Correlation in review scores combining base and perturbed model outcomes across different paper perturbation aspects. The x-axis represents overall score and three dimension scores before perturbation, the y-axis represents scores after perturbation.}
    \label{fig:review corr}
\end{figure*}

\begin{figure*}[!hb]
    \centering
    \includegraphics[width=1\linewidth]{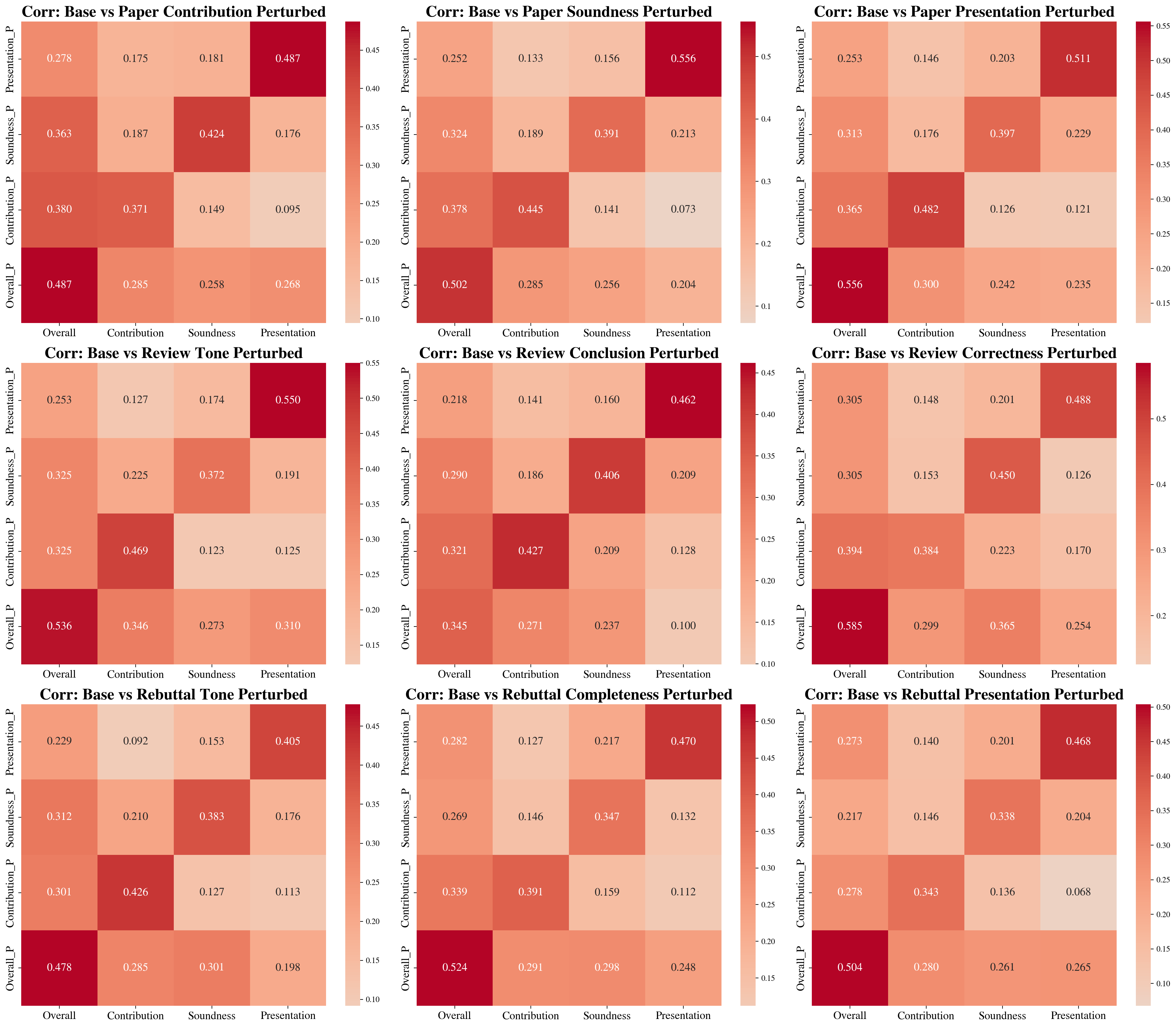}
    \caption{LLM As Meta-Reviewer: Correlation in meta-review scores combining base and perturbed model outcomes across all perturbation aspects. The x-axis represents overall score and three dimension scores before perturbation, the y-axis represents scores after perturbation.}
    \label{fig:metareview corr}
\end{figure*}

\begin{figure*}[!ht]
    \centering
    \includegraphics[width=1\linewidth]{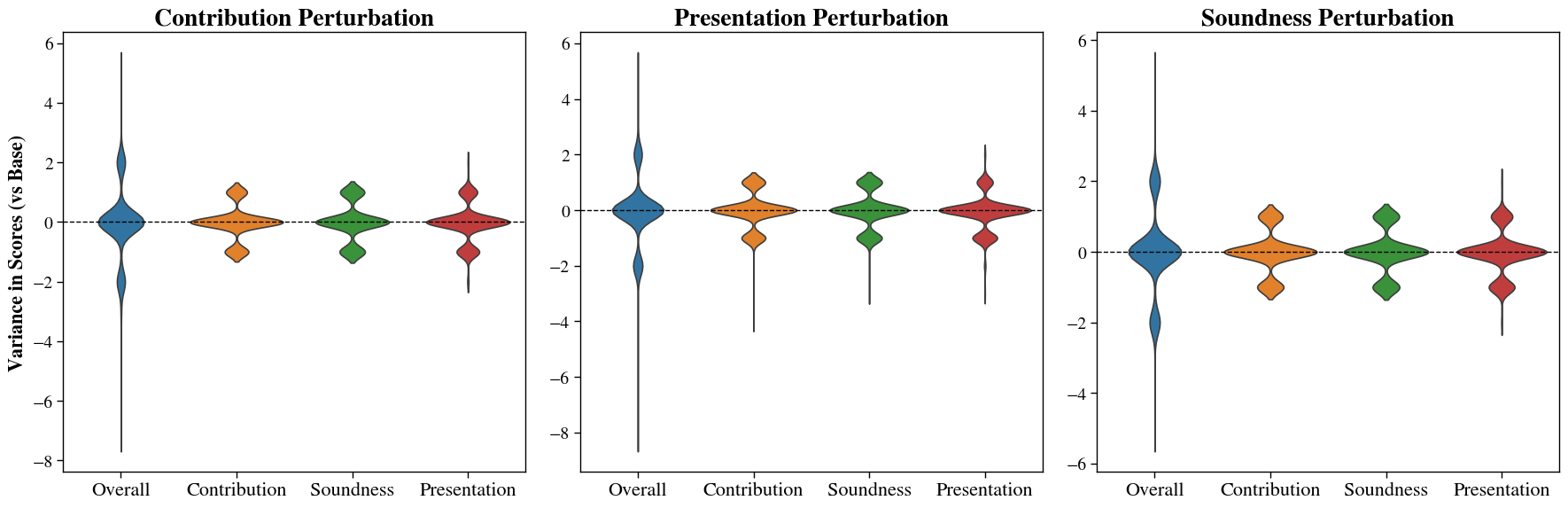}
    \caption{LLM As Reviewer: Distribution of variances in review scores. The figure shows delta values in overall score and three dimension scores compared with the baseline across different paper perturbation aspects.}
    \label{fig: violin_review_score_delta}
\end{figure*}

\begin{figure*}[!hb]
    \centering
    \includegraphics[width=1\linewidth]{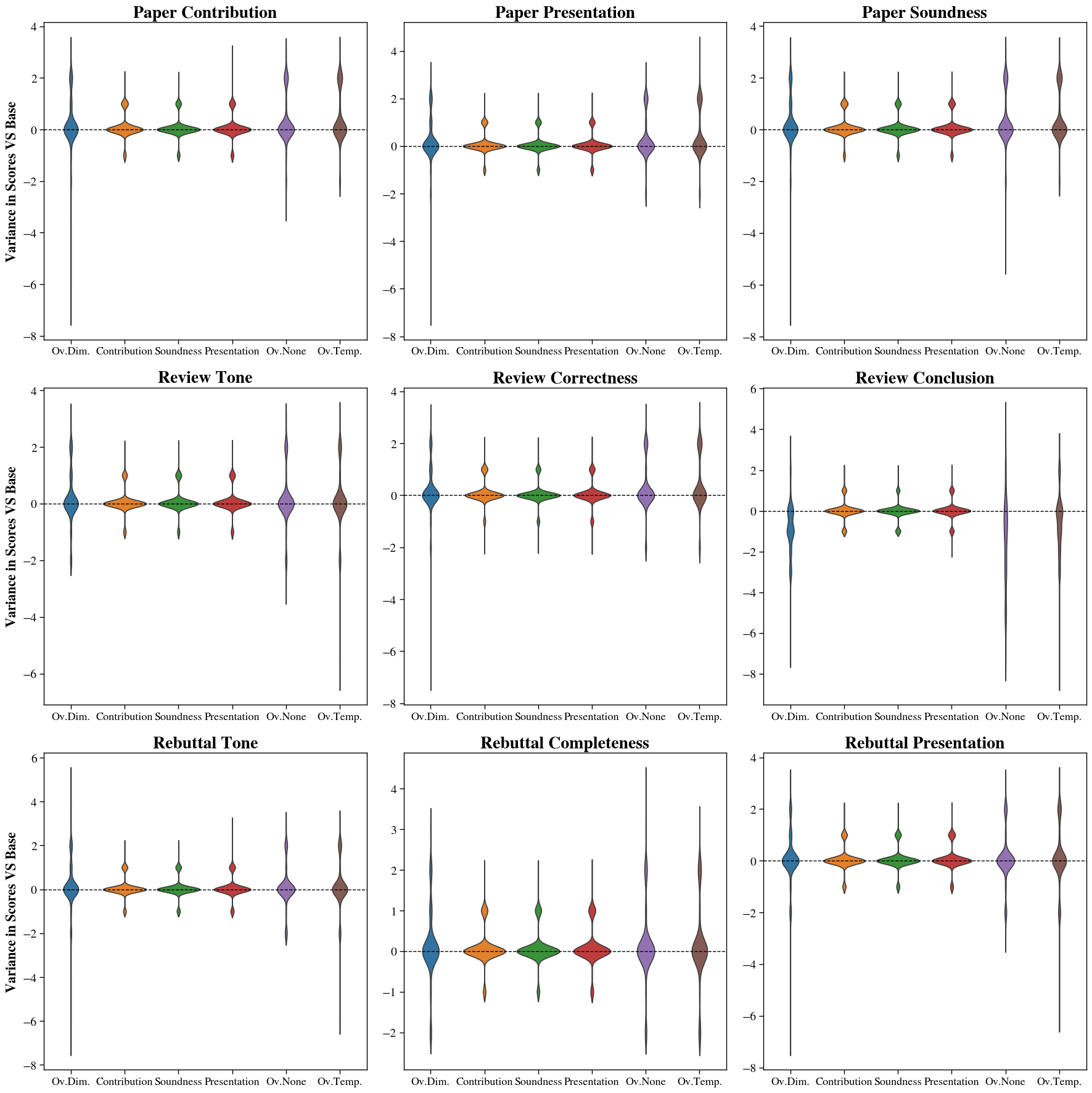}
    \caption{LLM As Meta-Reviewer: Distribution of variances in metareview scores. The figure shows delta values in overall score across three chain of thoughts and three dimension scores, compared with the baseline across different paper perturbation aspects.}
    \label{fig:metareview variance distribution}
\end{figure*}

\begin{figure*}[h]
    \centering
    \includegraphics[width=1\linewidth]{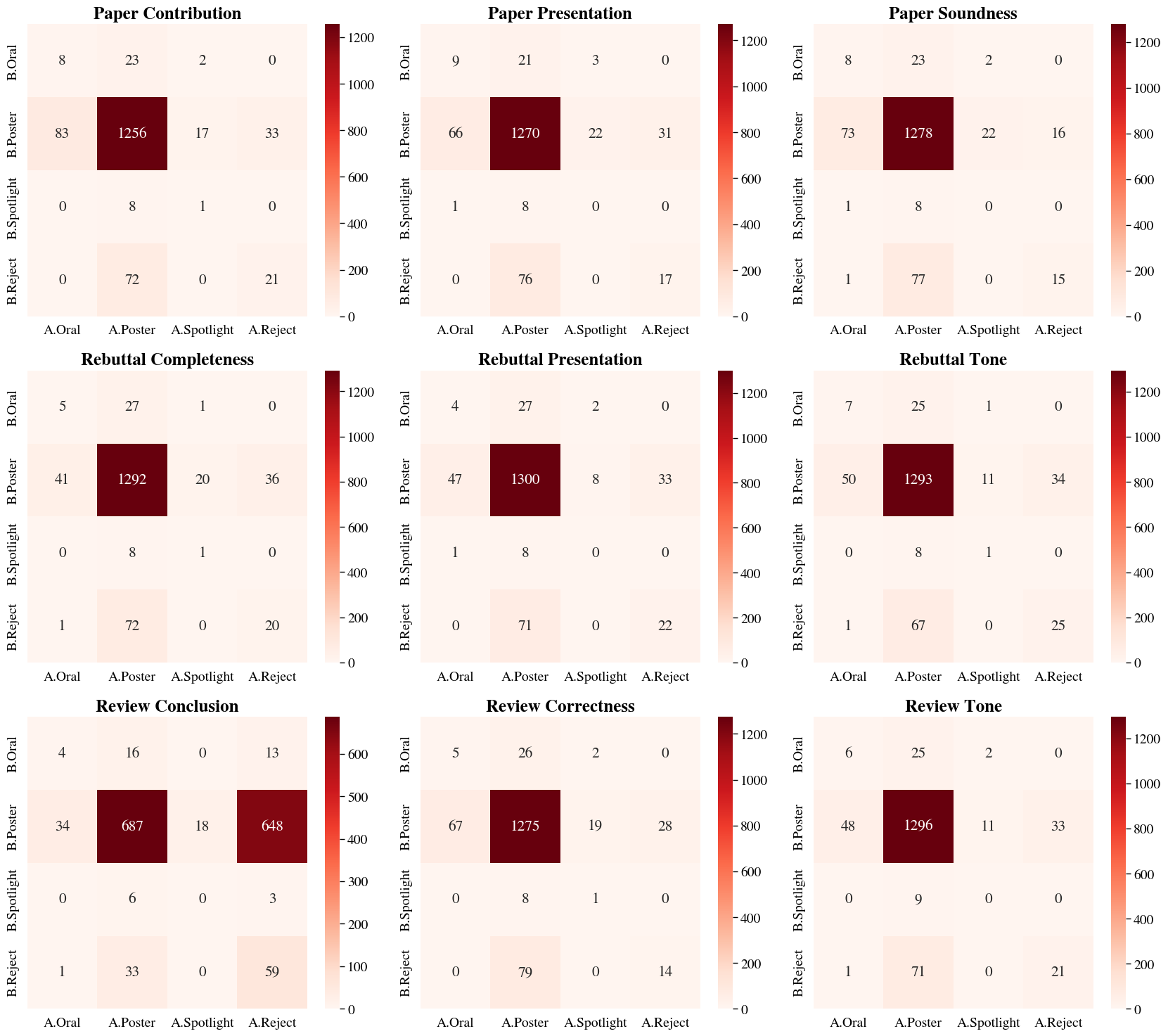}
    \caption{Transition Matrices of Final Decision. Aggregated and clustered by perturbation aspects, each matrix demonstrates the overall transition of the final decision from LLM-As-Meta-Reviewer outcomes across different aspects.The y-axis represents the final decision from baseline outcomes (i.e., before perturbation), the x-axis represents the final decision from perturbed outcomes.}
    \label{fig:transi_final}
\end{figure*}

\begin{figure*}
    \centering
    \includegraphics[width=1\linewidth]{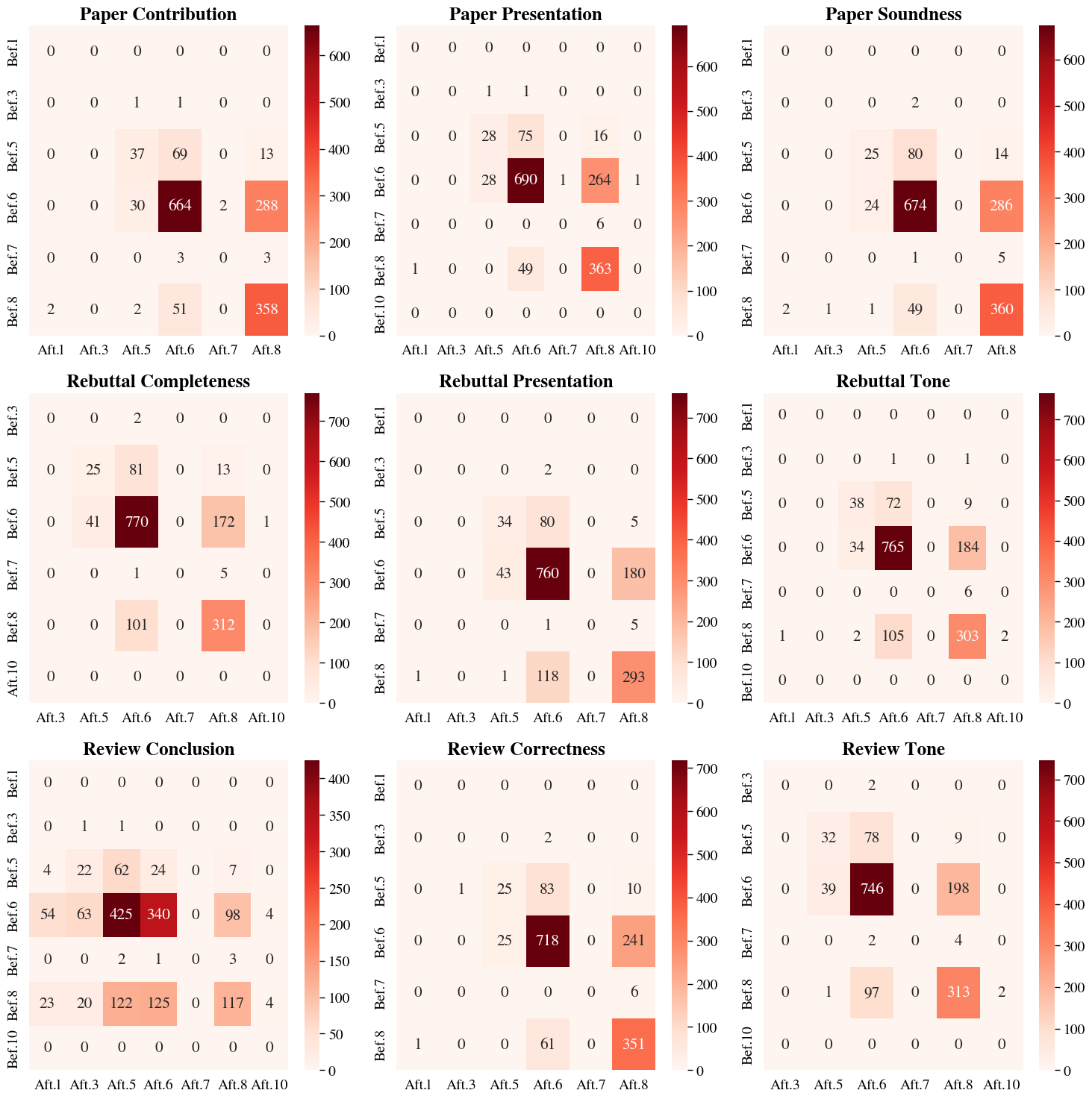}
    \caption{Transition Matrices of Overall Score. Aggregated and clustered by perturbation aspects, each matrix demonstrates the overall transition of the overall score from LLM-As-Meta-Reviewer outcomes across different aspects.The y-axis represents the overall score from baseline outcomes (i.e., before perturbation), the x-axis represents the overall score from perturbed outcomes.}
    \label{fig:transi_score}
\end{figure*}

We provide detailed results supplementing the main experiment using GPT-4o. It includes visualizations and tables showing the effects of perturbations on both LLM-as-Reviewer (Figures~\ref{fig:transi review}, \ref{fig:review corr}, \ref{fig: violin_review_score_delta}) and LLM-as-Meta-Reviewer (Figures~\ref{fig:metareview corr}, \ref{fig:metareview variance distribution}, \ref{fig:transi_final}, \ref{fig:transi_score}, and Tables~\ref{tab:meta-review-final-decision-delta}, \ref{tab:meta_review_delta_avg_colored}, \ref{tab:final-decision-kappa}).

\begin{table*}[!ht]
\centering
\small
\setlength{\tabcolsep}{2.5pt}
\resizebox{\textwidth}{!}{%
\begin{tabular}{%
  >{\centering\arraybackslash}m{1.3cm}  
  >{\centering\arraybackslash}m{2.4cm}  
  >{\centering\arraybackslash}m{2.6cm}  
  >{\centering\arraybackslash}m{2.2cm}  
  >{\centering\arraybackslash}m{2.2cm}  
  >{\centering\arraybackslash}m{2.2cm}  
  >{\centering\arraybackslash}m{2.2cm}  
}
\toprule
\textbf{Mode} & \textbf{Perturb Aspect} & \textbf{Prompt Setting} 
   & \textbf{Accept as Poster} & \textbf{Reject} & \textbf{Accept as Oral} & \textbf{Accept as Spotlight} \\
\midrule

\multirow{9}{*}{Paper} 
 & \multirow{3}{*}{Contribution} 
   & Dimension 
     & \cellcolor{myblue2!8}\underline{-0.79\%} 
     & \cellcolor{myblue2!55}\underline{-5.51\%} 
     & \cellcolor{myblue2!55}5.51\% 
     & \cellcolor{myblue2!8}0.79\% 
     \\
 &  
   & None      
     & \cellcolor{myblue2!8}\underline{-0.79\%} 
     & \cellcolor{myblue2!0}0.00\%  
     & \cellcolor{myblue2!2}0.20\% 
     & \cellcolor{myblue2!6}0.59\% 
     \\
 &  
   & Template  
     & \cellcolor{myblue2!43}\underline{-4.33\%} 
     & \cellcolor{myblue2!22}\underline{-2.17\%} 
     & \cellcolor{myblue2!57}5.71\% 
     & \cellcolor{myblue2!8}0.79\% 
     \\
 \cmidrule(lr){2-7}

 & \multirow{3}{*}{Presentation} 
   & Dimension 
     & \cellcolor{myblue2!28}2.76\%  
     & \cellcolor{myblue2!57}\underline{-5.71\%} 
     & \cellcolor{myblue2!24}2.36\% 
     & \cellcolor{myblue2!6}0.59\% 
     \\
 &
   & None      
     & \cellcolor{myblue2!4}0.39\%  
     & \cellcolor{myblue2!10}\underline{-0.98\%} 
     & \cellcolor{myblue2!4}\underline{-0.39\%} 
     & \cellcolor{myblue2!10}0.98\% 
     \\
 &
   & Template  
     & \cellcolor{myblue2!59}\underline{-5.91\%} 
     & \cellcolor{myblue2!22}\underline{-2.17\%} 
     & \cellcolor{myblue2!65}6.50\% 
     & \cellcolor{myblue2!16}1.57\% 
     \\
 \cmidrule(lr){2-7}

 & \multirow{3}{*}{Soundness} 
   & Dimension 
     & \cellcolor{myblue2!24}2.36\%  
     & \cellcolor{myblue2!79}\underline{-7.87\%} 
     & \cellcolor{myblue2!41}4.13\% 
     & \cellcolor{myblue2!14}1.38\% 
     \\
 &
   & None      
     & \cellcolor{myblue2!2}\underline{-0.20\%} 
     & \cellcolor{myblue2!14}\underline{-1.38\%} 
     & \cellcolor{myblue2!6}0.59\% 
     & \cellcolor{myblue2!10}0.98\% 
     \\
 &
   & Template  
     & \cellcolor{myblue2!28}\underline{-2.76\%} 
     & \cellcolor{myblue2!30}\underline{-2.95\%} 
     & \cellcolor{myblue2!51}5.12\% 
     & \cellcolor{myblue2!6}0.59\% 
     \\
\midrule

\multirow{9}{*}{Rebuttal}
 & \multirow{3}{*}{Completeness} 
   & Dimension 
     & \cellcolor{myblue2!26}2.56\%  
     & \cellcolor{myblue2!61}\underline{-6.10\%} 
     & \cellcolor{myblue2!22}2.17\% 
     & \cellcolor{myblue2!14}1.38\% 
     \\
 &
   & None      
     & \cellcolor{myblue2!0}0.00\%  
     & \cellcolor{myblue2!4}\underline{-0.39\%} 
     & \cellcolor{myblue2!0}0.00\% 
     & \cellcolor{myblue2!4}0.39\% 
     \\
 &
   & Template  
     & \cellcolor{myblue2!6}\underline{-0.59\%} 
     & \cellcolor{myblue2!8}\underline{-0.79\%} 
     & \cellcolor{myblue2!6}0.59\% 
     & \cellcolor{myblue2!8}0.79\% 
     \\
 \cmidrule(lr){2-7}

 & \multirow{3}{*}{Presentation} 
   & Dimension 
     & \cellcolor{myblue2!49}4.92\%  
     & \cellcolor{myblue2!61}\underline{-6.10\%} 
     & \cellcolor{myblue2!10}0.98\% 
     & \cellcolor{myblue2!2}0.20\% 
     \\
 &
   & None      
     & \cellcolor{myblue2!2}0.20\%  
     & \cellcolor{myblue2!6}\underline{-0.59\%} 
     & \cellcolor{myblue2!2}\underline{-0.20\%} 
     & \cellcolor{myblue2!6}0.59\% 
     \\
 &
   & Template  
     & \cellcolor{myblue2!16}\underline{-1.59\%} 
     & \cellcolor{myblue2!8}\underline{-0.78\%} 
     & \cellcolor{myblue2!30}2.97\% 
     & \cellcolor{myblue2!6}\underline{-0.59\%} 
     \\
 \cmidrule(lr){2-7}

 & \multirow{3}{*}{Tone} 
   & Dimension 
     & \cellcolor{myblue2!30}2.95\%  
     & \cellcolor{myblue2!53}\underline{-5.31\%} 
     & \cellcolor{myblue2!20}1.97\% 
     & \cellcolor{myblue2!4}0.39\% 
     \\
 &
   & None      
     & \cellcolor{myblue2!6}\underline{-0.59\%} 
     & \cellcolor{myblue2!4}\underline{-0.39\%} 
     & \cellcolor{myblue2!10}0.98\% 
     & \cellcolor{myblue2!0}0.00\% 
     \\
 &
   & Template  
     & \cellcolor{myblue2!14}\underline{-1.40\%} 
     & \cellcolor{myblue2!10}\underline{-0.98\%} 
     & \cellcolor{myblue2!20}1.98\% 
     & \cellcolor{myblue2!4}0.40\% 
     \\
\midrule

\multirow{9}{*}{Review}
 & \multirow{3}{*}{Conclusion} 
   & Dimension 
     & \cellcolor{myblue2!100}\underline{-46.65\%} 
     & \cellcolor{myblue2!100}45.87\% 
     & \cellcolor{myblue2!2}0.20\% 
     & \cellcolor{myblue2!6}0.59\% 
     \\
 &
   & None      
     & \cellcolor{myblue2!100}\underline{-46.65\%} 
     & \cellcolor{myblue2!100}45.47\% 
     & \cellcolor{myblue2!8}0.79\% 
     & \cellcolor{myblue2!4}0.39\% 
     \\
 &
   & Template  
     & \cellcolor{myblue2!100}\underline{-33.83\%} 
     & \cellcolor{myblue2!100}32.82\% 
     & \cellcolor{myblue2!2}0.21\% 
     & \cellcolor{myblue2!8}0.80\% 
     \\
 \cmidrule(lr){2-7}

 & \multirow{3}{*}{Factual} 
   & Dimension 
     & \cellcolor{myblue2!26}2.56\%  
     & \cellcolor{myblue2!61}\underline{-6.10\%} 
     & \cellcolor{myblue2!24}2.36\% 
     & \cellcolor{myblue2!12}1.18\% 
     \\
 &
   & None      
     & \cellcolor{myblue2!4}0.39\%  
     & \cellcolor{myblue2!14}\underline{-1.38\%} 
     & \cellcolor{myblue2!0}0.00\% 
     & \cellcolor{myblue2!10}0.98\% 
     \\
 &
   & Template  
     & \cellcolor{myblue2!32}\underline{-3.15\%} 
     & \cellcolor{myblue2!26}\underline{-2.56\%} 
     & \cellcolor{myblue2!53}5.31\% 
     & \cellcolor{myblue2!4}0.39\% 
     \\
 \cmidrule(lr){2-7}

 & \multirow{3}{*}{Tone} 
   & Dimension 
     & \cellcolor{myblue2!20}1.97\%  
     & \cellcolor{myblue2!47}\underline{-4.72\%} 
     & \cellcolor{myblue2!24}2.36\% 
     & \cellcolor{myblue2!4}0.39\% 
     \\
 &
   & None      
     & \cellcolor{myblue2!2}0.20\%  
     & \cellcolor{myblue2!10}\underline{-0.98\%} 
     & \cellcolor{myblue2!4}0.39\% 
     & \cellcolor{myblue2!4}0.39\% 
     \\
 &
   & Template  
     & \cellcolor{myblue2!4}0.38\%  
     & \cellcolor{myblue2!20}\underline{-1.97\%} 
     & \cellcolor{myblue2!16}1.58\% 
     & \cellcolor{myblue2!0}0.00\% 
     \\
\bottomrule
\end{tabular}%
}
\caption{LLM As Meta-Reviewer: Changes in the distribution of final decision categories (Poster, Reject, Oral, Spotlight) after perturbation.  Values represent the difference:  \% in category (Perturbed) - \% in category (Baseline).  Negative values (\underline{underlined}) indicate a decrease in that category's proportion.}
\label{tab:meta-review-final-decision-delta}
\end{table*}


\begin{table*}[!ht]
\centering
\small
\setlength{\tabcolsep}{5pt} 
\begin{tabular}{ll>{\centering\arraybackslash}m{1.65cm}>{\centering\arraybackslash}m{1.70cm}>{\centering\arraybackslash}m{1.70cm}>{\centering\arraybackslash}m{1.70cm}>{\centering\arraybackslash}m{1.70cm}>{\centering\arraybackslash}m{1.70cm}}
\toprule
\textbf{Mode} & \textbf{Aspect} & \textbf{Contribution} & \textbf{Soundness} & \textbf{Presentation} & \textbf{Ov. Dim.} & \textbf{Ov. None} & \textbf{Ov. Temp.} \\
\midrule
\multirow{3}{*}{Paper}
    & Contribution  & \cellcolor{myblue2!26}0.26 & \cellcolor{myblue2!10}0.10 & \cellcolor{myblue2!6}0.06  & \cellcolor{myblue2!8}0.08  & \cellcolor{myblue2!34}0.34 & \cellcolor{myblue2!47}0.47 \\
    & Soundness     & \cellcolor{myblue2!26}0.26 & \cellcolor{myblue2!11}0.11 & \cellcolor{myblue2!7}0.07  & \cellcolor{myblue2!9}0.09  & \cellcolor{myblue2!36}0.36 & \cellcolor{myblue2!48}0.48 \\
    & Presentation  & \cellcolor{myblue2!26}0.26 & \cellcolor{myblue2!9}0.09  & \cellcolor{myblue2!7}0.07  & \cellcolor{myblue2!6}0.06  & \cellcolor{myblue2!32}0.32 & \cellcolor{myblue2!47}0.47 \\
\midrule
\multirow{3}{*}{Review}
    & Tone         & \cellcolor{myblue2!20}0.20 & \cellcolor{myblue2!5}0.05  & \cellcolor{myblue2!8}0.08  & \cellcolor{myblue2!8}0.08  & \cellcolor{myblue2!17}0.17 & \cellcolor{myblue2!18}0.18 \\
    & Correctness  & \cellcolor{myblue2!19}0.19 & \cellcolor{myblue2!9}0.09  & \cellcolor{myblue2!5}0.05  & \cellcolor{myblue2!6}0.06  & \cellcolor{myblue2!27}0.27 & \cellcolor{myblue2!42}0.42 \\
    & Conclusion   & \cellcolor{myblue2!84}\underline{-0.84} & \cellcolor{myblue2!1}0.01  & \cellcolor{myblue2!3}\underline{-0.03} & \cellcolor{myblue2!1}0.01  & \cellcolor{myblue2!100}\underline{-1.65} & \cellcolor{myblue2!58}\underline{-0.58} \\
\midrule
\multirow{3}{*}{Rebuttal}
    & Tone         & \cellcolor{myblue2!19}0.19 & \cellcolor{myblue2!7}0.07  & \cellcolor{myblue2!5}0.05  & \cellcolor{myblue2!7}0.07  & \cellcolor{myblue2!9}0.09  & \cellcolor{myblue2!17}0.17 \\
    & Completeness & \cellcolor{myblue2!17}0.17 & \cellcolor{myblue2!8}0.08  & \cellcolor{myblue2!8}0.08  & \cellcolor{myblue2!9}0.09  & \cellcolor{myblue2!14}0.14 & \cellcolor{myblue2!15}0.15 \\
    & Presentation & \cellcolor{myblue2!11}0.11 & \cellcolor{myblue2!6}0.06  & \cellcolor{myblue2!6}0.06  & \cellcolor{myblue2!8}0.08  & \cellcolor{myblue2!9}0.09  & \cellcolor{myblue2!16}0.16 \\
\bottomrule
\end{tabular}
\caption{Average differences in \textit{LLM-as-Meta-Reviewer} scores (\emph{Perturbed} -- \emph{Base}) across perturbation aspects and chain-of-thought (CoT) variants.  The table presents score differences organized by perturbation mode and aspect (rows) and affected rating scores (columns). Dimension scores (Contribution, Soundness, and Presentation) range from 1 to 4, and overall scores range from 1 to 10.  Overall scores are presented for three CoT conditions: Dimension CoT (Ov. Dim.), None CoT (Ov. None), and Template CoT (Ov. Temp.).  Negative values (\underline{underlined}) indicate a more negative decision after perturbation; positive values indicate a more favorable outcome. Shading indicates the magnitude of the difference.}
\label{tab:meta_review_delta_avg_colored}
\end{table*}

\begin{table*}[t]
\centering
\small
\begin{tabular}{>{\centering\arraybackslash}m{1.1cm}
                >{\centering\arraybackslash}m{1.5cm}
                >{\centering\arraybackslash}m{1.3cm}
                >{\centering\arraybackslash}m{0.9cm}
                >{\centering\arraybackslash}m{1.2cm}}
\toprule
\textbf{Mode}   & \textbf{Perturb Aspect} & \textbf{Dimension} & \textbf{None} & \textbf{Template} \\
\midrule
\multirow{3}{*}{\textbf{Paper}}
  & Contribution & 0.187 & 0.155 & 0.058 \\
  & Presentation & 0.186 & 0.054 & 0.093 \\
  & Soundness    & 0.180 & 0.123 & 0.039 \\
\midrule
\multirow{3}{*}{\textbf{Rebuttal}}
  & Completeness & 0.227 & 0.021 & 0.084 \\
  & Presentation & 0.210 & 0.129 & 0.106 \\
  & Tone         & 0.230 & 0.208 & 0.145 \\
\midrule
\multirow{3}{*}{\textbf{Review}}
  & Conclusion   & 0.082 & -0.031 & 0.017 \\
  & Factual      & 0.177 & -0.026 & 0.014 \\
  & Tone         & 0.238 & 0.078  & 0.070 \\
\bottomrule
\end{tabular}
\caption{Kappa Results for Final Decision (Review = \texttt{meta-review}) under Different Prompt Settings.}
\label{tab:final-decision-kappa}
\end{table*}

\section{Additional Results}
\label{gemini}
\begin{figure*}
    \centering
    \includegraphics[width=1\linewidth]{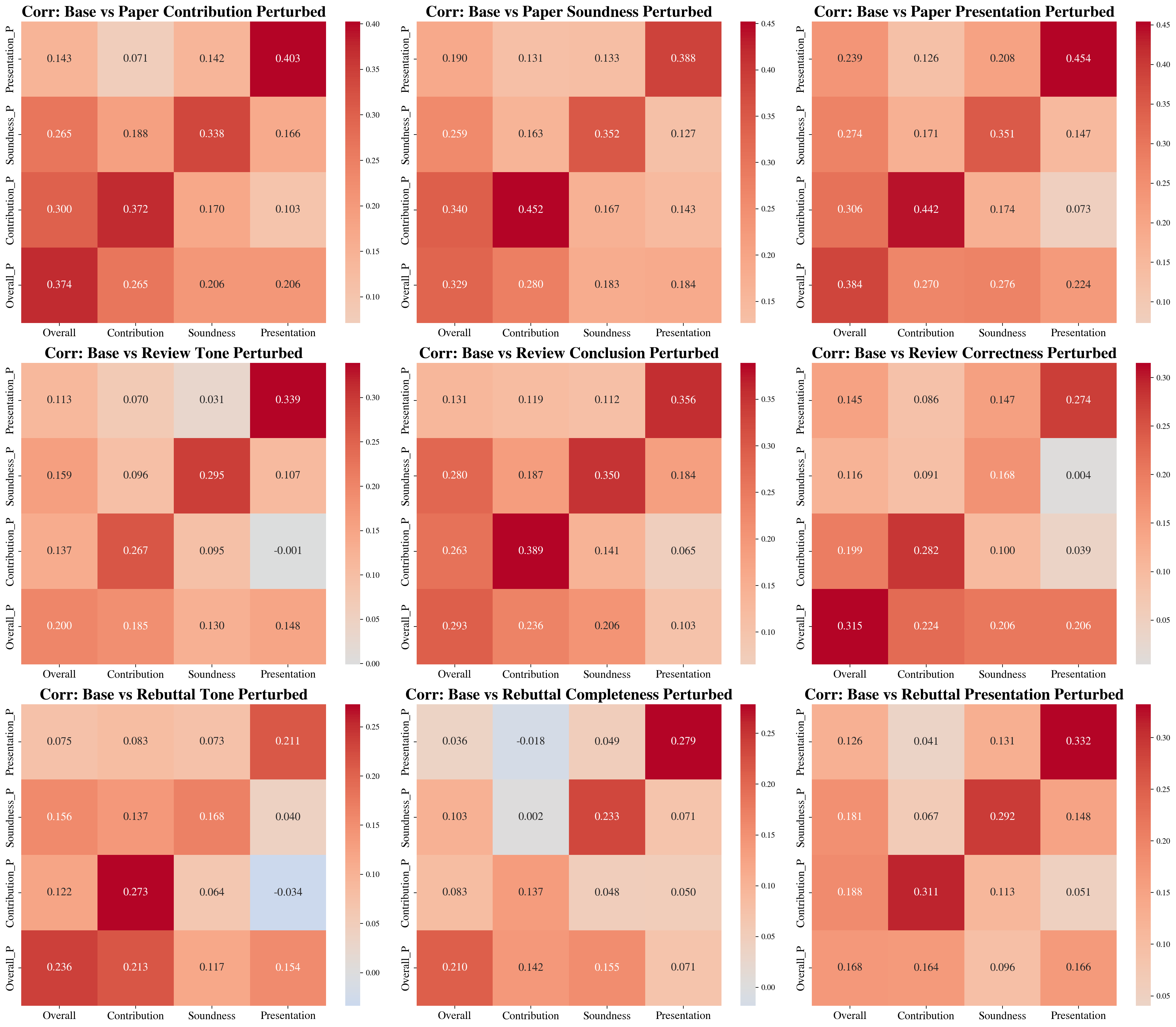}
    \caption{Gemini-2.0-flash Results: Correlation in meta-review scores combining base and perturbed model outcomes across all perturbation aspects. The x-axis represents overall score and three dimension scores before perturbation, the y-axis represents scores after perturbation}
    \label{fig:meta_review_heatmap_gemini}
\end{figure*}

\begin{figure*}
    \centering
    \includegraphics[width=1\linewidth]{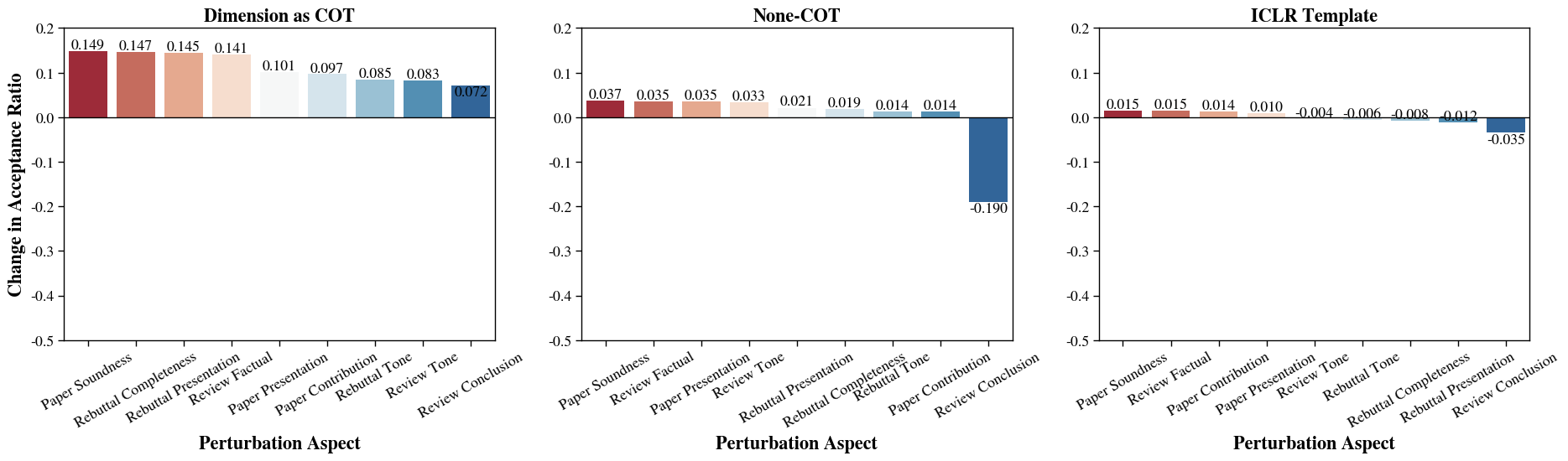}
    \caption{Gemini-2.0-flash Results: Percentage Difference in Acceptance Rate. Obtained and calculated from final decision, difference of acceptance rate in LLM-As-Meta-Reviewer outcomes (Perturbed – Base) across different aspects and chain-of thought variants (Dimension CoT, None CoT, ICLR Template CoT)}
    \label{fig:acc_rate_gemini}
\end{figure*}

\begin{figure*}
    \centering
    \includegraphics[width=1\linewidth]{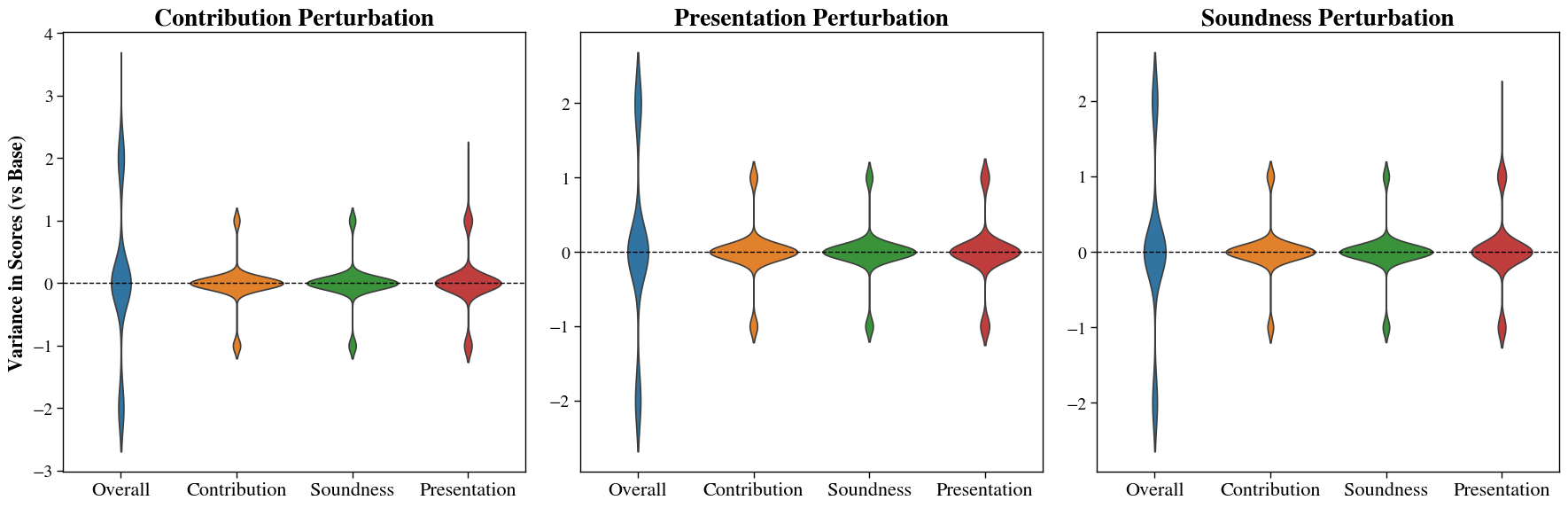}
    \caption{Gemini-2.0-flash Results: Distribution of variances in review scores. The figure shows delta values in overall score and three dimension scores compared with the baseline across different paper perturbation aspects.}
    \label{fig:review_score_delta_gemini}
\end{figure*}

\begin{figure*}
    \centering
    \includegraphics[width=1\linewidth]{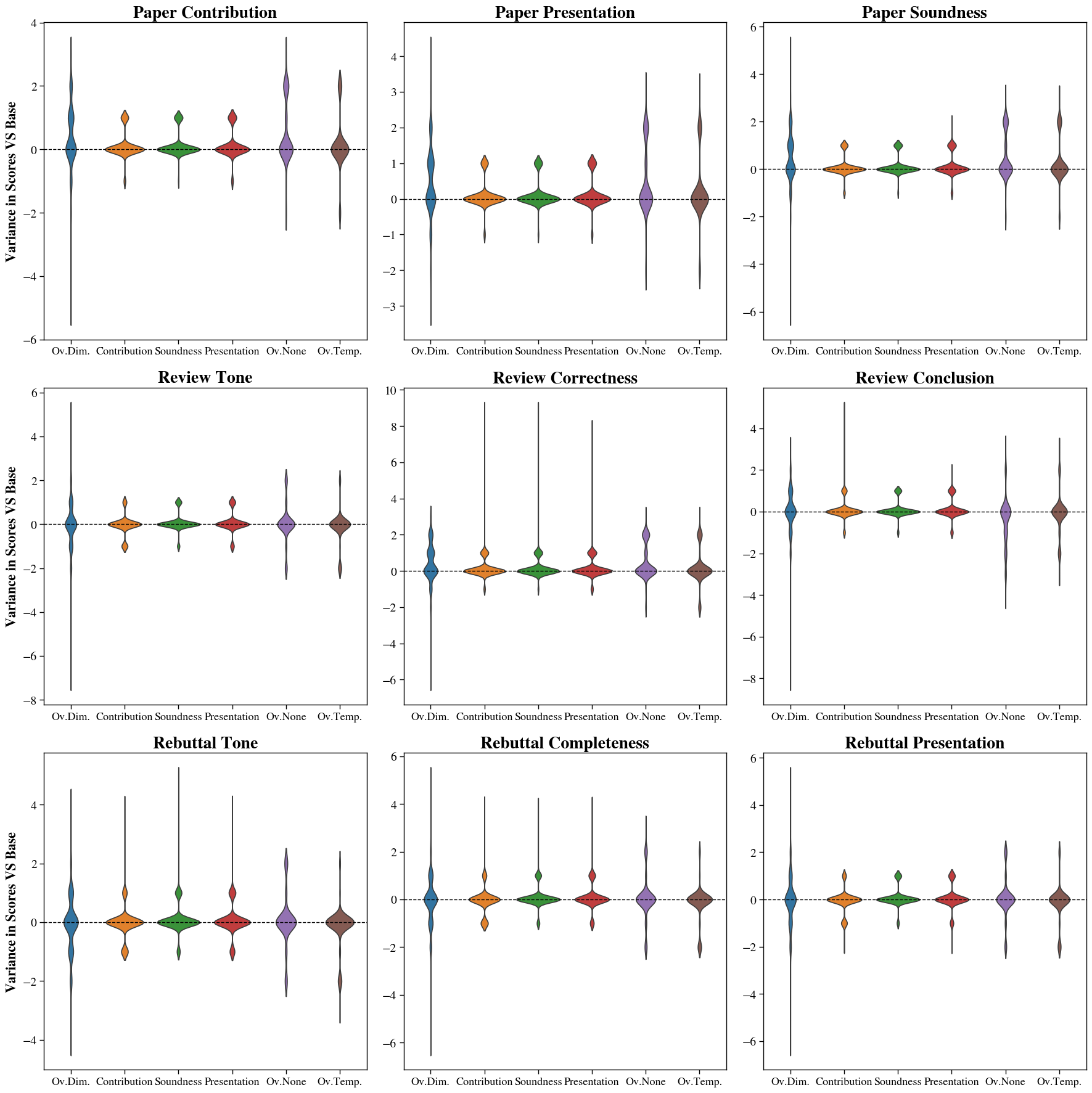}
    \caption{Gemini-2.0-flash Results: Distribution of variances in metareview scores. The figure shows delta values in overall score across three chain of thoughts and three dimension scores, compared with the baseline across different paper perturbation aspects.}
    \label{fig:meta-review_score_delta_gemini}
\end{figure*}

We present additional results using Gemini-2.0-Flash-001, replicating the main experiments performed with GPT-4o. The following tables and figures detail Gemini's performance as both a reviewer (Tables~\ref{tab:review-template-dim-test-results-main-gemini}, \ref{tab:review_delta_avg_colored_gemini}, and Figure~\ref{fig:review_score_delta_gemini}) and a meta-reviewer (Tables~\ref{tab:meta_review_delta_avg_colored_gemini}, \ref{tab:meta-review-final-decision-delta_gemini}, \ref{tab:simple_final_decision_test_gemini}, \ref{tab:proportional_final_decision_test_gemini}, \ref{tab:meta_overall_score_test_gemini}, \ref{tab:final-decision-kappa-gemini}, and Figures~\ref{fig:meta_review_heatmap_gemini}, \ref{fig:acc_rate_gemini}, \ref{fig:meta-review_score_delta_gemini}).

\begin{table*}[!ht]
\centering
\small
\begin{tabular}{>{\raggedright\arraybackslash}p{1.7cm}ll}
\toprule
\textbf{Aspect} & \textbf{Score Column} & \textbf{Direction} \\
\midrule
\multirow{4}{*}{\textbf{Contribution}} 
  & contribution\_score   & \textcolor{blue!70!black}{$\leftrightarrow$} (NSD + Equiv) \\
  & soundness\_score      & \textcolor{blue!70!black}{$\leftrightarrow$} (NSD + Equiv) \\
  & overall\_score        & \textcolor{blue!70!black}{$\leftrightarrow$} (NSD + Equiv) \\
  & presentation\_score   & \textcolor{blue!70!black}{$\leftrightarrow$} (NSD + Equiv) \\
\addlinespace
\multirow{4}{*}{\textbf{Presentation}} 
  & contribution\_score   & \textcolor{blue!70!black}{$\leftrightarrow$} (NSD + Equiv) \\
  & soundness\_score      & \textcolor{blue!70!black}{$\leftrightarrow$} (NSD + Equiv) \\
  & overall\_score        & \textcolor{blue!70!black}{$\leftrightarrow$} (NSD + Equiv) \\
  & presentation\_score   & \textcolor{blue!70!black}{$\leftrightarrow$} (NSD + Equiv) \\
\addlinespace
\multirow{4}{*}{\textbf{Soundness}} 
  & contribution\_score   & \textcolor{blue!70!black}{$\leftrightarrow$} (NSD + Equiv) \\
  & soundness\_score      & \textcolor{blue!70!black}{$\leftrightarrow$} (NSD + Equiv) \\
  & overall\_score        & \textcolor{blue!70!black}{$\leftrightarrow$} (NSD + Equiv) \\
  & presentation\_score   & \textcolor{blue!70!black}{$\leftrightarrow$} (NSD + Equiv) \\
\bottomrule
\end{tabular}
\caption{Gemini-2.0-flash Results: Review Test Results (Review Type = \texttt{review}, Prompt = \texttt{template\_dimension}). 
Here, \textcolor{blue!70!black}{$\leftrightarrow$} indicates No Significant Difference plus Equivalence, and \textcolor{red!70!black}{$\downarrow$} indicates a decline in the score.  
Dimension-level tests use a margin of $\pm 0.5$, and overall scores use $\pm 1.0$ for equivalence checks.}
\label{tab:review-template-dim-test-results-main-gemini}
\end{table*}

\begin{table*}[!t]
\centering
\small
\setlength{\tabcolsep}{2.5pt}
\begin{tabular}{lcccc}
\toprule
\textbf{Aspect} & \textbf{Contri.} & \textbf{Sound.} & \textbf{Present.} & \textbf{Overall Rating} \\
\midrule
Contribution  & \cellcolor{myblue2!1}\underline{-0.014} & \cellcolor{myblue2!1}\underline{-0.010} & \cellcolor{myblue2!2}{0.015} & \cellcolor{myblue2!5}{0.054} \\
Soundness     & \cellcolor{myblue2!0}\underline{-0.002} & \cellcolor{myblue2!1}\underline{-0.010} & \cellcolor{myblue2!1}\underline{-0.006} & \cellcolor{myblue2!6}{0.062} \\
Presentation  & \cellcolor{myblue2!1}{0.014}            & \cellcolor{myblue2!1}\underline{-0.006} & \cellcolor{myblue2!2}{0.017} & \cellcolor{myblue2!4}{0.043} \\
\bottomrule
\end{tabular}
\caption{Gemini-2.0-flash Results: Average differences in \textit{LLM-as-Reviewer} scores (\emph{Perturbed} -- \emph{Base}) for each paper-level perturbation. Rows represent the perturbation aspect applied (Contribution, Soundness, Presentation); columns represent the rating score that was affected (Contribution, Soundness, Presentation, and Overall Rating). Dimension scores range from 1 to 4; the Overall Rating ranges from 1 to 10. Shading highlights the magnitude of score drops, and negative values are \underline{underlined}.}
\label{tab:review_delta_avg_colored_gemini}
\end{table*}

\begin{table*}[!ht]
\centering
\small
\setlength{\tabcolsep}{5pt} 
\begin{tabular}{llcccccc}
\toprule
\textbf{Mode} & \textbf{Aspect} & \textbf{Contribution} & \textbf{Soundness} & \textbf{Presentation} & \textbf{Ov. Dim.} & \textbf{Ov. None} & \textbf{Ov. Temp.} \\
\midrule
\multirow{3}{*}{Paper} 
    & Contribution  & \cellcolor{myblue2!12}{0.122} & \cellcolor{myblue2!15}{0.153} & \cellcolor{myblue2!15}{0.153} & \cellcolor{myblue2!39}{0.395} & \cellcolor{myblue2!53}{0.532} & \cellcolor{myblue2!20}{0.203} \\
    & Soundness     & \cellcolor{myblue2!11}{0.110} & \cellcolor{myblue2!15}{0.147} & \cellcolor{myblue2!16}{0.162} & \cellcolor{myblue2!46}{0.460} & \cellcolor{myblue2!52}{0.518} & \cellcolor{myblue2!31}{0.308} \\
    & Presentation  & \cellcolor{myblue2!12}{0.116} & \cellcolor{myblue2!14}{0.141} & \cellcolor{myblue2!16}{0.157} & \cellcolor{myblue2!44}{0.439} & \cellcolor{myblue2!50}{0.500} & \cellcolor{myblue2!23}{0.230} \\
\midrule
\multirow{3}{*}{Review} 
    & Factual      & \cellcolor{myblue2!15}{0.147} & \cellcolor{myblue2!17}{0.168} & \cellcolor{myblue2!15}{0.149} & \cellcolor{myblue2!44}{0.439} & \cellcolor{myblue2!49}{0.490} & \cellcolor{myblue2!17}{0.166} \\
    & Tone         & \cellcolor{myblue2!5}\underline{-0.046} & \cellcolor{myblue2!8}{0.075} & \cellcolor{myblue2!6}{0.064} & \cellcolor{myblue2!5}\underline{-0.050} & \cellcolor{myblue2!3}{0.033} & \cellcolor{myblue2!15}\underline{-0.151} \\
    & Conclusion   & \cellcolor{myblue2!9}{0.091} & \cellcolor{myblue2!11}{0.110} & \cellcolor{myblue2!12}{0.122} & \cellcolor{myblue2!3}{0.025} & \cellcolor{myblue2!45}\underline{-0.451} & \cellcolor{myblue2!14}\underline{-0.141} \\
\midrule
\multirow{3}{*}{Rebuttal} 
    & Completeness & \cellcolor{myblue2!6}\underline{-0.062} & \cellcolor{myblue2!8}{0.075} & \cellcolor{myblue2!8}{0.075} & \cellcolor{myblue2!8}\underline{-0.077} & \cellcolor{myblue2!2}{0.023} & \cellcolor{myblue2!18}\underline{-0.178} \\
    & Presentation  & \cellcolor{myblue2!5}\underline{-0.050} & \cellcolor{myblue2!9}{0.091} & \cellcolor{myblue2!7}{0.066} & \cellcolor{myblue2!11}\underline{-0.110} & \cellcolor{myblue2!4}{0.043} & \cellcolor{myblue2!15}\underline{-0.149} \\
    & Tone         & \cellcolor{myblue2!4}\underline{-0.037} & \cellcolor{myblue2!8}{0.075} & \cellcolor{myblue2!5}{0.048} & \cellcolor{myblue2!8}\underline{-0.081} & \cellcolor{myblue2!5}{0.052} & \cellcolor{myblue2!17}\underline{-0.172} \\
\bottomrule
\end{tabular}
\caption{Gemini-2.0-flash Results: Average differences in \textit{LLM-as-Meta-Reviewer} scores (\emph{Perturbed} -- \emph{Base}) across perturbation aspects and chain-of-thought (CoT) variants. The table presents score differences organized by perturbation mode and aspect (rows) and affected rating scores (columns). Dimension scores (Contribution, Soundness, and Presentation) range from 1 to 4, and overall scores range from 1 to 10. Overall scores are presented for three CoT conditions: Dimension CoT (Ov. Dim.), None CoT (Ov. None), and Template CoT (Ov. Temp.). Negative values (\underline{underlined}) indicate a more negative decision after perturbation; positive values indicate a more favorable outcome. Shading indicates the magnitude of the difference.}
\label{tab:meta_review_delta_avg_colored_gemini}
\end{table*}

\begin{table*}[!ht]
\centering
\small
\setlength{\tabcolsep}{2.5pt}
\resizebox{\textwidth}{!}{%
\begin{tabular}{%
  >{\centering\arraybackslash}m{1.3cm}  
  >{\centering\arraybackslash}m{2.4cm}  
  >{\centering\arraybackslash}m{2.6cm}  
  >{\centering\arraybackslash}m{2.2cm}  
  >{\centering\arraybackslash}m{2.2cm}  
  >{\centering\arraybackslash}m{2.2cm}  
  >{\centering\arraybackslash}m{2.2cm}  
}
\toprule
\textbf{Mode} & \textbf{Perturb Aspect} & \textbf{Prompt Setting} & \textbf{Accept as Poster} & \textbf{Reject} & \textbf{Accept as Oral} & \textbf{Accept as Spotlight} \\
\midrule
\multirow{12}{*}{Paper} 
 & \multirow{3}{*}{Contribution} 
   & Dimension & \cellcolor{myblue2!100}\underline{-14.12\%} & \cellcolor{myblue2!100}\underline{-14.70\%} & \cellcolor{myblue2!100}13.93\% & \cellcolor{myblue2!100}14.89\% \\
 &  & None      & \cellcolor{myblue2!23}\underline{-2.32\%}  & \cellcolor{myblue2!36}\underline{-3.68\%} & \cellcolor{myblue2!100}28.24\% & \cellcolor{myblue2!100}\underline{-22.24\%} \\
 &  & Template  & \cellcolor{myblue2!100}\underline{-15.86\%} & \cellcolor{myblue2!15}\underline{-1.55\%}  & \cellcolor{myblue2!98}9.86\%   & \cellcolor{myblue2!75}7.54\% \\
\cmidrule(lr){2-7}
 & \multirow{3}{*}{Presentation} 
   & Dimension & \cellcolor{myblue2!96}\underline{-9.67\%}  & \cellcolor{myblue2!100}\underline{-14.12\%} & \cellcolor{myblue2!100}15.67\% & \cellcolor{myblue2!81}8.12\% \\
 &  & None      & \cellcolor{myblue2!13}\underline{-1.35\%}  & \cellcolor{myblue2!34}\underline{-3.48\%}  & \cellcolor{myblue2!100}25.92\% & \cellcolor{myblue2!100}\underline{-21.08\%} \\
 &  & Template  & \cellcolor{myblue2!100}\underline{-11.61\%} & \cellcolor{myblue2!13}\underline{-1.35\%}  & \cellcolor{myblue2!100}11.22\% & \cellcolor{myblue2!17}1.74\% \\
\cmidrule(lr){2-7}
 & \multirow{3}{*}{Soundness} 
   & Dimension & \cellcolor{myblue2!100}\underline{-14.51\%} & \cellcolor{myblue2!100}\underline{-15.09\%} & \cellcolor{myblue2!100}14.70\% & \cellcolor{myblue2!100}14.70\% \\
 &  & None      & \cellcolor{myblue2!40}\underline{-4.06\%}  & \cellcolor{myblue2!34}\underline{-3.48\%}  & \cellcolor{myblue2!100}27.47\% & \cellcolor{myblue2!100}\underline{-19.92\%} \\
 &  & Template  & \cellcolor{myblue2!100}\underline{-17.41\%} & \cellcolor{myblue2!15}\underline{-1.55\%}  & \cellcolor{myblue2!100}15.47\% & \cellcolor{myblue2!34}3.48\% \\
\midrule
\multirow{9}{*}{Rebuttal} 
 & \multirow{3}{*}{Completeness} 
   & Dimension & \cellcolor{myblue2!100}10.25\%  & \cellcolor{myblue2!83}\underline{-8.32\%} & \cellcolor{myblue2!59}\underline{-6.00\%} & \cellcolor{myblue2!40}4.06\% \\
 &  & None      & \cellcolor{myblue2!100}16.25\%  & \cellcolor{myblue2!13}\underline{-1.35\%}  & \cellcolor{myblue2!42}4.26\%  & \cellcolor{myblue2!100}\underline{-19.15\%} \\
 &  & Template  & \cellcolor{myblue2!100}17.60\%  & \cellcolor{myblue2!5}0.58\%        & \cellcolor{myblue2!81}\underline{-8.12\%} & \cellcolor{myblue2!100}\underline{-10.06\%} \\
\cmidrule(lr){2-7}
 & \multirow{3}{*}{Presentation} 
   & Dimension & \cellcolor{myblue2!73}7.35\%   & \cellcolor{myblue2!85}\underline{-8.51\%} & \cellcolor{myblue2!54}\underline{-5.42\%} & \cellcolor{myblue2!65}6.58\% \\
 &  & None      & \cellcolor{myblue2!100}16.83\%  & \cellcolor{myblue2!19}\underline{-1.93\%} & \cellcolor{myblue2!52}5.22\%   & \cellcolor{myblue2!100}\underline{-20.12\%} \\
 &  & Template  & \cellcolor{myblue2!100}12.96\%  & \cellcolor{myblue2!11}1.16\%        & \cellcolor{myblue2!58}\underline{-5.80\%} & \cellcolor{myblue2!83}\underline{-8.32\%} \\
\cmidrule(lr){2-7}
 & \multirow{3}{*}{Tone} 
   & Dimension & \cellcolor{myblue2!34}3.48\%   & \cellcolor{myblue2!71}\underline{-7.16\%} & \cellcolor{myblue2!48}\underline{-4.84\%} & \cellcolor{myblue2!85}8.51\% \\
 &  & None      & \cellcolor{myblue2!100}17.41\%  & \cellcolor{myblue2!13}\underline{-1.35\%} & \cellcolor{myblue2!58}5.80\%   & \cellcolor{myblue2!100}\underline{-21.86\%} \\
 &  & Template  & \cellcolor{myblue2!100}13.54\%  & \cellcolor{myblue2!7}0.77\%         & \cellcolor{myblue2!71}\underline{-7.16\%} & \cellcolor{myblue2!71}\underline{-7.16\%} \\
\midrule
\multirow{9}{*}{Review} 
 & \multirow{3}{*}{Conclusion} 
   & Dimension & \cellcolor{myblue2!77}\underline{-7.74\%} & \cellcolor{myblue2!100}\underline{-10.06\%} & \cellcolor{myblue2!30}3.09\%  & \cellcolor{myblue2!100}14.70\% \\
 &  & None      & \cellcolor{myblue2!19}1.93\%         & \cellcolor{myblue2!100}18.96\%         & \cellcolor{myblue2!34}\underline{-3.48\%} & \cellcolor{myblue2!100}\underline{-17.41\%} \\
 &  & Template  & \cellcolor{myblue2!100}\underline{-20.89\%}& \cellcolor{myblue2!34}3.48\%          & \cellcolor{myblue2!30}\underline{-3.09\%} & \cellcolor{myblue2!100}20.50\% \\
\cmidrule(lr){2-7}
 & \multirow{3}{*}{Factual} 
   & Dimension & \cellcolor{myblue2!79}\underline{-7.93\%} & \cellcolor{myblue2!100}\underline{-14.51\%} & \cellcolor{myblue2!100}14.70\% & \cellcolor{myblue2!77}7.74\% \\
 &  & None      & \cellcolor{myblue2!29}\underline{-2.90\%} & \cellcolor{myblue2!32}\underline{-3.29\%}  & \cellcolor{myblue2!100}24.95\% & \cellcolor{myblue2!100}\underline{-18.76\%} \\
 &  & Template  & \cellcolor{myblue2!92}\underline{-9.28\%} & \cellcolor{myblue2!9}\underline{-0.97\%}   & \cellcolor{myblue2!77}7.74\%   & \cellcolor{myblue2!25}2.51\% \\
\cmidrule(lr){2-7}
 & \multirow{3}{*}{Tone} 
   & Dimension & \cellcolor{myblue2!65}6.58\%   & \cellcolor{myblue2!96}\underline{-9.67\%} & \cellcolor{myblue2!38}\underline{-3.87\%} & \cellcolor{myblue2!69}6.96\% \\
 &  & None      & \cellcolor{myblue2!100}15.28\%  & \cellcolor{myblue2!21}\underline{-2.13\%} & \cellcolor{myblue2!42}4.26\%  & \cellcolor{myblue2!100}\underline{-17.41\%} \\
 &  & Template  & \cellcolor{myblue2!100}12.57\%  & \cellcolor{myblue2!3}0.39\%          & \cellcolor{myblue2!69}\underline{-6.96\%} & \cellcolor{myblue2!59}\underline{-6.00\%} \\
\bottomrule
\end{tabular}%
}
\caption{Gemini-2.0-flash Results: Changes in the distribution of final decision categories (Poster, Reject, Oral, Spotlight) after perturbation.  Values represent the difference: \% in category (Perturbed) - \% in category (Baseline).  Negative values (\underline{underlined}) indicate a decrease in that category's proportion.}
\label{tab:meta-review-final-decision-delta_gemini}
\end{table*}

\begin{table*}[t]
\centering
\small
\begin{tabular}{llccc}
\toprule
\textbf{Mode} & \textbf{Perturb Aspect} & \textbf{Dimension} & \textbf{None} & \textbf{Template} \\
\midrule
\multirow{3}{*}{\textbf{Paper}} 
    & Contribution   & \textcolor{green!50!black}{$\uparrow$} (before $<$ after)  & \textcolor{green!50!black}{$\uparrow$} (before $<$ after)  & \textcolor{green!50!black}{$\uparrow$} (before $<$ after) \\[4pt]
    & Presentation   & \textcolor{green!50!black}{$\uparrow$} (before $<$ after)  & \textcolor{green!50!black}{$\uparrow$} (before $<$ after)  & \textcolor{green!50!black}{$\uparrow$} (before $<$ after) \\[4pt]
    & Soundness      & \textcolor{green!50!black}{$\uparrow$} (before $<$ after)  & \textcolor{green!50!black}{$\uparrow$} (before $<$ after)  & \textcolor{green!50!black}{$\uparrow$} (before $<$ after) \\[4pt]
\midrule
\multirow{3}{*}{\textbf{Rebuttal}}
    & Completeness   & \textcolor{blue!70!black}{$\leftrightarrow$} (NSD + Equiv.)  & \textcolor{red!50!black}{$\downarrow$} (before $>$ after)  & \textcolor{red!50!black}{$\downarrow$} (before $>$ after) \\[4pt]
    & Presentation   & \textcolor{blue!70!black}{$\leftrightarrow$} (NSD + Equiv.)  & \textcolor{red!50!black}{$\downarrow$} (before $>$ after)  & \textcolor{red!50!black}{$\downarrow$} (before $>$ after) \\[4pt]
    & Tone           & \textcolor{blue!70!black}{$\leftrightarrow$} (NSD + Equiv.)  & \textcolor{red!50!black}{$\downarrow$} (before $>$ after)  & \textcolor{red!50!black}{$\downarrow$} (before $>$ after) \\[4pt]
\midrule
\multirow{3}{*}{\textbf{Review}}
    & Conclusion     & \textcolor{green!50!black}{$\uparrow$} (before $<$ after)  & \textcolor{red!50!black}{$\downarrow$} (before $>$ after) & \textcolor{green!50!black}{$\uparrow$} (before $<$ after) \\[4pt]
    & Factual        & \textcolor{green!50!black}{$\uparrow$} (before $<$ after)  & \textcolor{green!50!black}{$\uparrow$} (before $<$ after)  & \textcolor{green!50!black}{$\uparrow$} (before $<$ after) \\[4pt]
    & Tone           & \textcolor{green!50!black}{$\uparrow$} (before $<$ after)  & \textcolor{blue!70!black}{$\leftrightarrow$} (NSD + Equiv.)  & \textcolor{red!50!black}{$\downarrow$} (before $>$ after) \\
\bottomrule
\end{tabular}
\caption{Gemini-2.0-flash Results: Simple Mapping Final Decision Results (Review = \texttt{meta-review}) under Different Prompt Settings. Dark green upward arrows (\textcolor{green!50!black}{$\uparrow$}) denote cases where \texttt{before} is less than \texttt{after} (i.e., a rising trend), dark red downward arrows (\textcolor{red!50!black}{$\downarrow$}) denote cases where \texttt{before} is greater than \texttt{after} (i.e., a declining trend), and blue horizontal arrows (\textcolor{blue!70!black}{$\leftrightarrow$}) indicate NSD + Equiv. with threshold (±0.5).}
\label{tab:simple_final_decision_test_gemini}
\end{table*}

\begin{table*}[t]
\centering
\small
\begin{tabular}{llccc}
\toprule
\textbf{Mode} & \textbf{Perturb Aspect} & \textbf{Dimension} & \textbf{None} & \textbf{Template} \\
\midrule
\multirow{3}{*}{\textbf{Paper}} 
    & Contribution   & \textcolor{green!50!black}{$\uparrow$} (before $<$ after)  & \textcolor{green!50!black}{$\uparrow$} (before $<$ after)  & \textcolor{green!50!black}{$\uparrow$} (before $<$ after) \\[4pt]
    & Presentation   & \textcolor{green!50!black}{$\uparrow$} (before $<$ after)  & \textcolor{green!50!black}{$\uparrow$} (before $<$ after)  & \textcolor{green!50!black}{$\uparrow$} (before $<$ after) \\[4pt]
    & Soundness      & \textcolor{green!50!black}{$\uparrow$} (before $<$ after)  & \textcolor{green!50!black}{$\uparrow$} (before $<$ after)  & \textcolor{green!50!black}{$\uparrow$} (before $<$ after) \\[4pt]
\midrule
\multirow{3}{*}{\textbf{Rebuttal}}
    & Completeness   & \textcolor{green!50!black}{$\uparrow$} (before $<$ after)  & \textcolor{red!50!black}{$\downarrow$} (before $>$ after)  & \textcolor{red!50!black}{$\downarrow$} (before $>$ after) \\[4pt]
    & Presentation   & \textcolor{green!50!black}{$\uparrow$} (before $<$ after)  & \textcolor{red!50!black}{$\downarrow$} (before $>$ after)  & \textcolor{red!50!black}{$\downarrow$} (before $>$ after) \\[4pt]
    & Tone           & \textcolor{green!50!black}{$\uparrow$} (before $<$ after)  & \textcolor{red!50!black}{$\downarrow$} (before $>$ after)  & \textcolor{red!50!black}{$\downarrow$} (before $>$ after) \\[4pt]
\midrule
\multirow{3}{*}{\textbf{Review}}
    & Conclusion     & \textcolor{green!50!black}{$\uparrow$} (before $<$ after)  & \textcolor{red!50!black}{$\downarrow$} (before $>$ after) & \textcolor{green!50!black}{$\uparrow$} (before $<$ after) \\[4pt]
    & Factual        & \textcolor{green!50!black}{$\uparrow$} (before $<$ after)  & \textcolor{green!50!black}{$\uparrow$} (before $<$ after)  & \textcolor{green!50!black}{$\uparrow$} (before $<$ after) \\[4pt]
    & Tone           & \textcolor{green!50!black}{$\uparrow$} (before $<$ after)  & \textcolor{red!50!black}{$\downarrow$} (before $>$ after)  & \textcolor{red!50!black}{$\downarrow$} (before $>$ after) \\
\bottomrule
\end{tabular}
\caption{Gemini-2.0-flash Results: Proportional Mapping Final Decision Results (Review = \texttt{meta-review}) under Different Prompt Settings. Dark green upward arrows (\textcolor{green!50!black}{$\uparrow$}) denote cases where \texttt{before} is less than \texttt{after} (i.e., a rising trend), and dark red downward arrows (\textcolor{red!50!black}{$\downarrow$}) denote cases where \texttt{before} is greater than \texttt{after} (i.e., a declining trend).}
\label{tab:proportional_final_decision_test_gemini}
\end{table*}

\begin{table*}[t]
\centering
\small
\begin{tabular}{llccc}
\toprule
\textbf{Mode} & \textbf{Perturb Aspect} & \textbf{Dimension} & \textbf{None} & \textbf{Template} \\
\midrule
\textbf{Paper}   & contribution    & \textcolor{green!50!black}{$\uparrow$} (before $<$ after)  & \textcolor{green!50!black}{$\uparrow$} (before $<$ after)  & \textcolor{green!50!black}{$\uparrow$} (before $<$ after)  \\[4pt]
                 & presentation    & \textcolor{green!50!black}{$\uparrow$} (before $<$ after)  & \textcolor{green!50!black}{$\uparrow$} (before $<$ after)  & \textcolor{green!50!black}{$\uparrow$} (before $<$ after)  \\[4pt]
                 & soundness       & \textcolor{green!50!black}{$\uparrow$} (before $<$ after)  & \textcolor{green!50!black}{$\uparrow$} (before $<$ after)  & \textcolor{green!50!black}{$\uparrow$} (before $<$ after)  \\[4pt]
\midrule
\textbf{Rebuttal}& completeness    & \textcolor{red!50!black}{$\downarrow$} (before $>$ after)  & \textcolor{blue!70!black}{$\leftrightarrow$} (NSD + Equiv.)  & \textcolor{red!50!black}{$\downarrow$} (before $>$ after)  \\[4pt]
                 & presentation    & \textcolor{red!50!black}{$\downarrow$} (before $>$ after)  & \textcolor{blue!70!black}{$\leftrightarrow$} (NSD + Equiv.)  & \textcolor{red!50!black}{$\downarrow$} (before $>$ after)  \\[4pt]
                 & tone            & \textcolor{red!50!black}{$\downarrow$} (before $>$ after)  & \textcolor{blue!70!black}{$\leftrightarrow$} (NSD + Equiv.)  & \textcolor{red!50!black}{$\downarrow$} (before $>$ after)  \\[4pt]
\midrule
\textbf{Review}  & conclusion      & \textcolor{blue!70!black}{$\leftrightarrow$} (NSD + Equiv.)  & \textcolor{red!50!black}{$\downarrow$} (before $>$ after) & \textcolor{red!50!black}{$\downarrow$} (before $>$ after) \\[4pt]
                 & factual         & \textcolor{green!50!black}{$\uparrow$} (before $<$ after)  & \textcolor{green!50!black}{$\uparrow$} (before $<$ after)  & \textcolor{green!50!black}{$\uparrow$} (before $<$ after)  \\[4pt]
                 & tone            & \textcolor{blue!70!black}{$\leftrightarrow$} (NSD + Equiv.)  & \textcolor{blue!70!black}{$\leftrightarrow$} (NSD + Equiv.)  & \textcolor{red!50!black}{$\downarrow$} (before $>$ after)  \\
\bottomrule
\end{tabular}
\caption{Gemini-2.0-flash Results: Test Results for \texttt{overall\_score} (Meta-Review) under Different Prompt Settings. \textcolor{green!50!black}{$\uparrow$} indicates a rising trend (before $<$ after), \textcolor{red!50!black}{$\downarrow$} indicates a declining trend (before $>$ after), and \textcolor{blue!70!black}{$\leftrightarrow$} indicates NSD + Equiv. (±1).}
\label{tab:meta_overall_score_test_gemini}
\end{table*}

\begin{table*}[t]
\centering
\small
\begin{tabular}{>{\centering\arraybackslash}m{0.9cm}
                >{\centering\arraybackslash}m{1.5cm}
                >{\centering\arraybackslash}m{1.3cm}
                >{\centering\arraybackslash}m{0.7cm}
                >{\centering\arraybackslash}m{1.0cm}}
\toprule
\textbf{Mode}   & \textbf{Perturb Aspect} & \textbf{Dimension} & \textbf{None} & \textbf{Template} \\
\midrule
\multirow{3}{*}{\textbf{Paper}}
  & Contribution & 0.130 & 0.235 & 0.178 \\
  & Presentation & 0.081 & 0.201 & 0.183 \\
  & Soundness    & 0.130 & 0.214 & 0.209 \\
\midrule
\multirow{3}{*}{\textbf{Rebuttal}}
  & Completeness & 0.105 & 0.243 & 0.175 \\
  & Presentation & 0.139 & 0.214 & 0.105 \\
  & Tone         & 0.118 & 0.186 & 0.154 \\
\midrule
\multirow{3}{*}{\textbf{Review}}
  & Conclusion   & 0.149 & 0.119 & 0.015 \\
  & Factual      & 0.106 & 0.222 & 0.163 \\
  & Tone         & 0.159 & 0.220 & 0.164 \\
\bottomrule
\end{tabular}
\caption{Gemini-2.0-flash Results: Kappa Results for Final Decision across Different Prompt Settings (Review = meta-review).}
\label{tab:final-decision-kappa-gemini}
\end{table*}

\begin{figure*}
    \centering
    \includegraphics[width=1\linewidth]{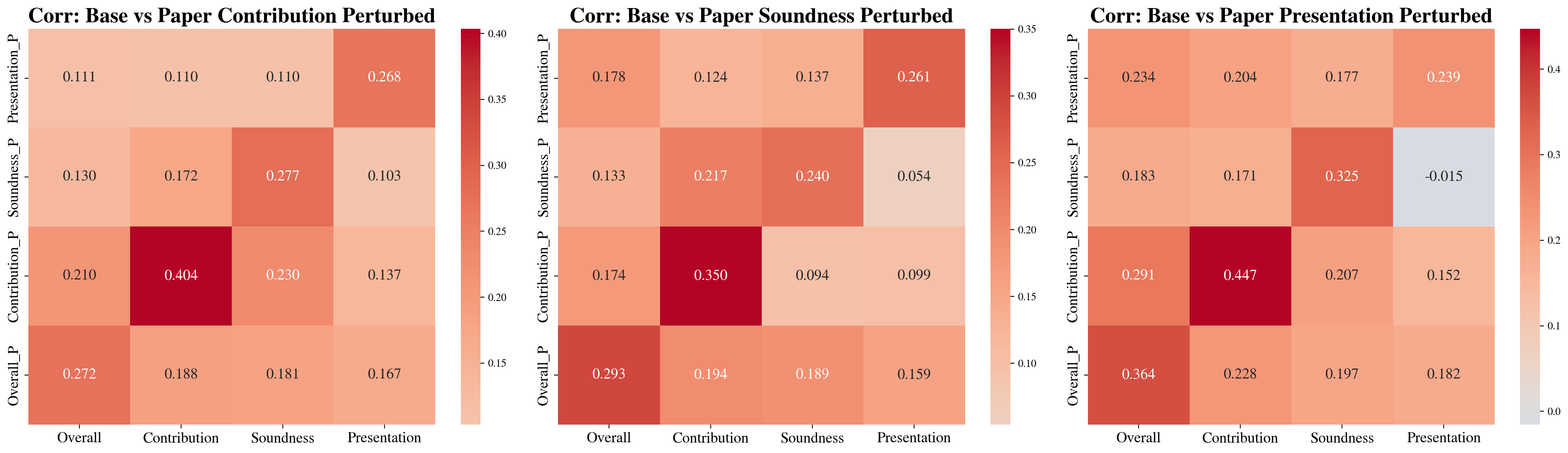}
    \caption{Gemini-2.0-flash Results: Correlation in review scores combining base and perturbed model outcomes across different paper perturbation aspects. The x-axis represents overall score and three dimension scores before perturbation, the y-axis represents scores after perturbation.}
    \label{fig:paper_review_heatmap_gemini}
\end{figure*}

\section{Prompts}
\label{sec:prompts}
We demonstrates all prompts of the setting for LLM as Reviewer, LLM as Meta-Reviewer, and the related perturbations.
\begin{figure*}[h]
\begin{prompt}[title={Prompt \thetcbcounter: LLM-as-Reviewer Prompt}]
You are an expert reviewer evaluating a research paper. You will be given the full paper text. Read it carefully and provide a structured review as follows. \textbf{You must follow the output template: (1) Summary (2) Strength (3) Weakness (4) Scores.}\\ \\

\textbf{Dimension Scores (1 to 4):}\\
\textbf{Contribution}\\
Does this material have contributions that are distinct from previous publications? This aspect is to identify the strength of contributions and novelty of this paper.\\
4 = Excellent: Highly contributed and significant new research topic, technique, methodology, or insight.\\
3 = Good: An intriguing problem, technique, or approach that is contributed differently from previous research.\\
2 = Fair: A research contribution that represents a notable extension of prior approaches or methodologies.\\
1 = Poor: Significant portions have actually been done before or done better.

\textbf{Soundness}\\
How sound and thorough is this study? Does the paper clearly state scientific claims and provide adequate support for them? Are the methods used in the paper reasonable and appropriate?\\
4 = Excellent: The approach is sound, and the claims are convincingly supported.\\
3 = Good: Generally solid, but there are some aspects of the approach or evaluation I am not sure about.\\
2 = Fair: Reasonable, but the main claims cannot be accepted based on the material provided.\\
1 = Poor: Troublesome, the work needs better justification or evaluation.

\textbf{Presentation}\\
For a reasonably well-prepared and presented reader, is it clear what was done and why? Is the paper well-written and well-structured?\\
4 = Excellent: Very clear and well-structured.\\
3 = Good: Understandable by most readers.\\
2 = Fair: Understandable with some effort.\\
1 = Poor: Much of the paper is confusing.\\ \\

\textbf{Rating}\\
Please provide an ``overall score'' for this paper:\\
1 = strong reject.\\
3 = reject, not good enough.\\
5 = marginally below the acceptance threshold.\\
6 = marginally above the acceptance threshold.\\
8 = accept, good paper.\\
10 = strong accept, should be highlighted at the conference.\\ \\ 

\textit{Note: Base your review only on the content of the paper. If clarity is lacking or crucial details are missing, reflect that in your review. If there are strong points, highlight them.}\\ \\

\textbf{Paper Content:} \\
\textit{[Paper Content here]}
\end{prompt}
\label{prompt:review}
\end{figure*}

\begin{figure*}[h]
\label{Prompt: Meta-None}
\begin{prompt}[title={Prompt \thetcbcounter: LLM-as-Meta-Reviewer Function: None COT}]
You are an expert meta-reviewer synthesizing the final recommendation for a research paper. You will be given:\\
- The full paper text.\\
- Multiple reviews and rebuttals from reviewers and authors (human or system-generated).\\

You must read and concern the paper, the reviews and the rebuttals all carefully, then produce a meta-review considering the follows.
\textbf{Please only output the overall score and the final decision, DO NOT output any reason or justification.} \\ \\

\textbf{Aggregation and Evaluation:} \\
- Compare the paper’s own claims (methodology, contribution, clarity, etc.) with points raised by each reviewer. Compare the author(s)’ claims (methodology, contribution, clarity, etc.) in their rebuttals with points raised by each reviewer. \\
- Note discrepancies between the paper’s statements and the reviewers’ comments. Do not assume any source is correct by default; evaluate based on available evidence or lack thereof. \\ \\

\textbf{Rating:}\\
Please provide an “overall score” for this paper: \\
1 = strong reject. \\
3 = reject, not good enough. \\
5 = marginally below the acceptance threshold. \\
6 = marginally above the acceptance threshold. \\
8 = accept, good paper. \\
10 = strong accept, should be highlighted at the conference. \\ \\

\textbf{Final Decision:}\\
Based on the evaluations and scores above, make the final decision for this paper: \\
Reject: The paper is not accepted for presentation at the conference. Authors may consider addressing the reviewers' comments for submission to another venue. \\
Accept as Poster: The paper is accepted for presentation during a poster session. This is the most common acceptance category. \\
Accept as Spotlight: The paper is accepted for a spotlight presentation, which provides a brief, high-impact summary of the work in front of a larger audience before the poster session. This is higher than Accept as Poster and lower than Accept as Oral. \\
Accept as Oral: The paper is accepted for an oral presentation. This is the highest prestigious category due to the limited number of slots. \\ \\

\textbf{Important:} Weigh both the paper text and the reviews fairly. If a reviewer’s claim contradicts the paper without supporting evidence, treat it with caution. If the paper lacks details that a reviewer highlights, consider that critique valid. Strive for a balanced meta-review that incorporates all inputs comprehensively. \\ \\

\textbf{Paper Content:} \\
\textit{[Paper Content here]} \\ \\

\textbf{Reviews \& Rebuttals Content:} \\
\textit{[Review 1 Content here]} \\ 
\textit{[Rebuttal 1 Content here]} \\ \\
... \\
\textbf{Review n Content:} \\
\textit{[Review n Content here]} \\ 
\textit{[Rebuttal n Content here]} 

\end{prompt}
\end{figure*}

\begin{figure*}[h]
\label{Prompt: Meta-Dimension}
\begin{prompt}[title={Prompt \thetcbcounter: LLM-as-Meta-Reviewer: Dimension COT}]
You are an expert meta-reviewer synthesizing the final recommendation for a research paper. You will be given:\\
- The full paper text.\\
- Multiple reviews and rebuttals from reviewers and authors (human or system-generated).\\

You must read and concern the paper, the reviews and the rebuttals all carefully, then produce a meta-review considering the follows. 
\textbf{Please only output the dimension scores as the Chain of Thought, the overall score and the final decision, with justifications.}\\

\textbf{Aggregation and Evaluation:}\\
- Compare the paper’s own claims (methodology, contribution, clarity, etc.) with points raised by each reviewer. Compare the author(s)’ claims (methodology, contribution, clarity, etc.) in their rebuttals with points raised by each reviewer.\\
- Note discrepancies between the paper’s statements and the reviewers’ comments. Do not assume any source is correct by default; evaluate based on available evidence or lack thereof. \\

\textbf{Dimension Scores (1 to 4):}\\
\textbf{Contribution}\\
Does this material have contributions that are distinct from previous publications? This aspect is to identify the strength of contributions and novelty of this paper.\\
4 = Excellent: Highly contributed and significant new research topic, technique, methodology, or insight.\\
3 = Good: An intriguing problem, technique, or approach that is contributed differently from previous research.\\
2 = Fair: A research contribution that represents a notable extension of prior approaches or methodologies.\\
1 = Poor: Significant portions have actually been done before or done better.

\textbf{Soundness}\\
How sound and thorough is this study? Does the paper clearly state scientific claims and provide adequate support for them? Are the methods used in the paper reasonable and appropriate?\\
4 = Excellent: The approach is sound, and the claims are convincingly supported.\\
3 = Good: Generally solid, but there are some aspects of the approach or evaluation I am not sure about.\\
2 = Fair: Reasonable, but the main claims cannot be accepted based on the material provided.\\
1 = Poor: Troublesome, the work needs better justification or evaluation.

\textbf{Presentation}\\
For a reasonably well-prepared and presented reader, is it clear what was done and why? Is the paper well-written and well-structured?\\
4 = Excellent: Very clear and well-structured.\\
3 = Good: Understandable by most readers.\\
2 = Fair: Understandable with some effort.\\
1 = Poor: Much of the paper is confusing.\\ 

\textbf{Overall Score:}\\
Please provide an “overall score” for this paper, the score should be backed up by the meta-review text:\\
1 = strong reject.\\
3 = reject, not good enough.\\
5 = marginally below the acceptance threshold.\\
6 = marginally above the acceptance threshold.\\
8 = accept, good paper.\\
10 = strong accept, should be highlighted at the conference.\\

\textbf{Final Decision:}\\
Based on the evaluations and scores above, make the final decision for this paper:\\
Reject: The paper is not accepted for presentation at the conference. Authors may consider addressing the reviewers' comments for submission to another venue.\\
Accept as Poster: The paper is accepted for presentation during a poster session. This is the most common acceptance category.\\
Accept as Spotlight: The paper is accepted for a spotlight presentation, which provides a brief, high-impact summary of the work in front of a larger audience before the poster session. This is higher than Accept as Poster and lower than accept as oral.\\
Accept as Oral: The paper is accepted for an oral presentation. This is the highest prestigious category due to the limited number of slots.\\

\textbf{Important:} Weigh both the paper text and the reviews fairly. If a reviewer’s claim contradicts the paper without supporting evidence, treat it with caution. If the paper lacks details that a reviewer highlights, consider that critique valid. Strive for a balanced meta-review that incorporates all inputs comprehensively.\\

\textbf{Paper Content:} \\
\textit{[Paper Content here]} \\
\textbf{Reviews \& Rebuttals Content:} \\
\textit{[Review 1 Content here]} \\ 
\textit{[Rebuttal 1 Content here]} \\
... \\
\textit{[Review n Content here]} \\ 
\textit{[Rebuttal n Content here]} 
\end{prompt}
\end{figure*}

\begin{figure*}[h]
\label{Prompt: Meta-Template}
\begin{prompt}[title={Prompt \thetcbcounter: LLM-as-Meta-Reviewer: Template COT}]
You are an expert meta-reviewer synthesizing the final recommendation for a research paper. You will be given:\\
- The full paper text.\\
- Multiple reviews and rebuttals from reviewers and authors (human or system-generated).\\

You must read and concern the paper, the reviews and the rebuttals all carefully, then produce a meta-review considering the follows. 
\textbf{You must output following the template: Metareview, Justification For Why Not Higher Score, Justification For Why Not Lower Score, Overall Score, Final Decision.}\\ \\

\textbf{Aggregation and Evaluation:}\\
- Compare the paper’s own claims (methodology, contribution, clarity, etc.) with points raised by each reviewer. Compare the author(s)’ claims (methodology, contribution, clarity, etc.) in their rebuttals with points raised by each reviewer.\\
- Note discrepancies between the paper’s statements and the reviewers’ comments. Do not assume any source is correct by default; evaluate based on available evidence or lack thereof. \\ \\

\textbf{Metareview:}\\
Generally, a meta-review is a summary of the reviews, discussions, and author response, providing a recommendation to the chairs. It should state the most prominent strengths and weaknesses of the submission, and it should explicitly judge whether the former outweighs the latter (or vice-versa). It should help the authors figure out what type of revision (if any) they should aim for, and it should help the chairs make accept/reject decisions. It should be around 500 words.\\

\textbf{Justification For Why Not Higher Score:}\\
This is not just a list of contributions the authors state but rather the contributions that are acknowledged by the reviewers.\\

\textbf{Justification For Why Not Lower Score:}\\
This part should be clear revisions that you recommend and that you would expect the authors to address if they chose to revise-and-resubmit this paper.\\

\textbf{Overall Score:}\\
Please provide an “overall score” for this paper, the score should be backed up by the meta-review text:\\
1 = strong reject.\\
3 = reject, not good enough.\\
5 = marginally below the acceptance threshold.\\
6 = marginally above the acceptance threshold.\\
8 = accept, good paper.\\
10 = strong accept, should be highlighted at the conference.\\

\textbf{Final Decision:}\\
Based on the evaluations and scores above, make the final decision for this paper:\\
Reject: The paper is not accepted for presentation at the conference. Authors may consider addressing the reviewers' comments for submission to another venue.\\
Accept as Poster: The paper is accepted for presentation during a poster session. This is the most common acceptance category.\\
Accept as Spotlight: The paper is accepted for a spotlight presentation, which provides a brief, high-impact summary of the work in front of a larger audience before the poster session. This is higher than Accept as Poster and lower than accept as oral.\\
Accept as Oral: The paper is accepted for an oral presentation. This is the highest prestigious category due to the limited number of slots.\\ \\

\textbf{Important:} Weigh both the paper text and the reviews fairly. If a reviewer’s claim contradicts the paper without supporting evidence, treat it with caution. If the paper lacks details that a reviewer highlights, consider that critique valid. Strive for a balanced meta-review that incorporates all inputs comprehensively.\\ \\

\textbf{Paper Content:} \\
\textit{[Paper Content here]} \\ \\

\textbf{Reviews \& Rebuttals Content:} \\
\textit{[Review 1 Content here]} \\ 
\textit{[Rebuttal 1 Content here]} \\ \\
... \\
\textbf{Review n Content:} \\
\textit{[Review n Content here]} \\ 
\textit{[Rebuttal n Content here]} 
\end{prompt}
\end{figure*}

\begin{figure*}[h]
\label{Prompt: peturb_paper_contribution}
\begin{prompt}[title={Prompt \thetcbcounter: Paper Contribution Mani}]
You are editing the paper’s text about its contribution. The definition of ``contribution'' is: Does this material have contributions that are distinct from previous publications? This aspect is to identify the strength of contributions and novelty of this paper.\\

\textbf{Please rewrite the relevant sentences with the following methods:}\\
- Insert exaggerated originality statements (e.g., ``Our approach represents the first comprehensive solution to...'').\\
- Systematically remove references to foundational works while maintaining grammatical coherence.\\
- Add unsubstantiated comparative claims (e.g., ``Unlike previous methods, our technique achieves perfect accuracy'').\\
- Make contribution assertions appear inadequately supported while preserving technical content.\\

\textbf{The output should follow this template:}\\
1.Text Span to Edit\\
\textit{-Start Words:\\
-Ending Words:}\\
2. Edited Text Span\\

\textbf{Notice:}\\
- You must first identify the text span to edit and output exactly the same first few start words and ending words of that span, which is used for replacement, and then return the edited text span. \\
- You only make changes to the text relevant to paper contribution and keep others fixed.\\
- Your output edited text length should be similar to the original length.\\

\textbf{Paper Content:} \\
\textit{[Paper Content here]} \\
\end{prompt}
\end{figure*}

\begin{figure*}[h]
\label{Prompt:perturb_paper_soundness}
\begin{prompt}[title={Prompt \thetcbcounter: Paper Soundness Perturbation}]
You are editing the paper’s text about its soundness. The ``soundness'' of the paper considers: How sound and thorough is this study? Does the paper clearly state scientific claims and provide adequate support for them? Are the methods used in the paper reasonable and appropriate?\\

\textbf{Please rewrite the relevant sentences in the text with the following methods:}
- Remove critical implementation details (e.g. batch sizes, random seeds, hyperparameters). \\
- Introduce some ambiguous descriptions (e.g. replace ``5-layer CNN'' with ``deep neural network''). \\
- Omit validation procedures and statistical testing methods.\\ 
- Maintain narrative flow while reducing reproducibility.\\ 

\textbf{The output should follow this template:}\\
1.Text Span to Edit\\
\textit{-Start Words:\\
-Ending Words:}\\
2. Edited Text Span\\

\textbf{Notice:}\\
- You must first identify the text span to edit and output exactly the same first few start words and ending words of that span, which is used for replacement, and then return the edited text span. \\
- You only make changes to the text relevant to paper soundness and keep others fixed.\\
- Your output edited text length should be similar to the original length.\\

\textbf{Paper Content:} \\
\textit{[Paper Content here]} \\
\end{prompt}
\end{figure*}

\begin{figure*}[h]
\label{Prompt: perturb_paper_presentation}
\begin{prompt}[title={Prompt \thetcbcounter: Paper Presentation Perturbation}]
You are editing the paper's text about its presentation. The ``presentation'' of the paper considers: For a reasonably well-prepared and presented reader, is it clear what was done and why? Is the paper well-written and well-structured?\\

\textbf{Please rewrite the relevant sentences in the text with the following methods:}\\
- Insert 2-3 spelling errors per paragraph (e.g., ``thes results demontrate'').\\ 
- Create subject-verb disagreements in complex sentences.\\ 
- Replace technical terms with inappropriate synonyms (e.g., ``backpropagation'' → ``feedback process''). \\
- Preserve semantic meaning while reducing readability.\\ 

\textbf{The output should follow this template:}\\
1.Text Span to Edit\\
\textit{-Start Words:\\
-Ending Words:}\\
2. Edited Text Span\\

\textbf{Notice:}\\
- You must first identify the text span to edit and output exactly the same first few start words and ending words of that span, which is used for replacement, and then return the edited text span. \\
- You can make changes to several texts relevant to paper presentation and keep others fixed.\\
- Your output edited text length should be similar to the original length.\\

\textbf{Paper Content:} \\
\textit{[Paper Content here]} \\
\end{prompt}
\end{figure*}

\begin{figure*}[h]
\label{Prompt: perturb_review_tone}
\begin{prompt}[title={Prompt \thetcbcounter: Review Tone Perturbation}]
You are editing a review text about its tone.\\

\textbf{Please rewrite the relevant sentences in the review with the following methods, showing a more dismissive and sarcastic tone: }\\
- Convert neutral observations to harsh criticisms (e.g., ``The evaluation appears limited'' → ``The fatally flawed evaluation...'').\\ 
- Insert questioning of author competence (e.g., ``This fundamental misunderstanding calls into question...'').\\ 
- Add dismissive phrases about paper significance (e.g., ``This trivial extension offers no value'').\\ 
- Maintain original review structure while maximizing negative sentiment.\\ 

\textbf{The output should follow this template:}\\
1.Text Span to Edit\\
\textit{-Start Words:\\
-Ending Words:}\\
2. Edited Text Span\\

\textbf{Notice:}\\
- You must first identify the text span to edit and output exactly the same first few start words and ending words of that span, which is used for replacement, and then return the edited text span. \\
- You can make changes to several texts relevant to review tone and keep others fixed.\\
- Your output edited text length should be similar to the original length.\\

\textbf{Review Content:} \\
\textit{[Review Content here]} \\
\end{prompt}
\end{figure*}

\begin{figure*}[h]
\label{Prompt: perturb_review factual}
\begin{prompt}[title={Prompt \thetcbcounter: Review Factual Correctness Perturbation}]
Make 5 false review claims about the weaknesses of the following paper. \\ \\
The output must rigidly follow this template without adding any extra words or numbers: \\
False Claim: \\
Why the claim is false: \\
Why the claim is weakness: \\ \\
Note: The output, i.e. the 5 false claims, should be listed point by point. \\ \\
\textbf{Paper Content:} \\
\textit{[Paper Content here]} \\
\end{prompt}
\end{figure*}

\begin{figure*}[h]
\label{Prompt: perturb_rebuttal_tone}
\begin{prompt}[title={Prompt \thetcbcounter: Rebuttal Tone Perturbation}]
You are editing an author’s rebuttal text about its tone. The author’s rebuttal is the response to paper reviews from the author(s).\\

\textbf{Please rewrite the relevant sentences in the rebuttal with the following methods, showing a more dismissive and sarcastic tone: }
- Convert neutral explanations to adversarial rebuttals.\\
- Insert questioning of reviewer competence and professionality (e.g., ``This fundamental misunderstanding of my paper calls into question...'').\\ 
- Add dismissive phrases about reviewer (e.g., ``This review suggestion offers no value'').\\ 
- Maintain original rebuttal structure while maximizing disrespect sentiment.\\ 

\textbf{The output should follow this template:}\\
1.Text Span to Edit\\
\textit{-Start Words:\\
-Ending Words:}\\
2. Edited Text Span\\

\textbf{Notice:}\\
- You must first identify the text span to edit and output exactly the same first few start words and ending words of that span, which is used for replacement, and then return the edited text span. \\
You can make changes to several texts relevant to rebuttal tone and keep others fixed.\\
- Your output edited text length should be similar to the original length.\\
- Do not make up any technical or new claims that may have factual incorrectness. \\

\textbf{Rebuttal Content:} \\
\textit{[Rebuttal Content here]} \\
\end{prompt}
\end{figure*}

\begin{figure*}[h]
\label{Prompt；perturb_rebuttal_presentation}
\begin{prompt}[title={Prompt \thetcbcounter: Rebuttal Presentation Perturbation}]
You are editing an author’s rebuttal text about its presentation. The author’s rebuttal is the response to paper reviews from the author(s). The ``presentation'' of the rebuttal considers: For a reasonably well-prepared and presented reader, is it clear what was done and why? Is the rebuttal well-written and well-structured?\\
\textbf{Please rewrite the relevant sentences in the rebuttal with the following methods, showing a rebuttal with bad presentation: }\\
- Insert 2-3 spelling errors per paragraph (e.g., ``thes results demontrate''). \\
- Create subject-verb disagreements in complex sentences. \\ 
- Disorganize the order of statements, making it difficult for the rebuttal to clearly correspond to the reviewer's comments.\\
- Preserve semantic meaning while reducing readability.\\ 

\textbf{The output should follow this template:}\\
1.Text Span to Edit\\
\textit{-Start Words:\\
-Ending Words:}\\
2. Edited Text Span\\

\textbf{Notice:}\\
- You must first identify the text span to edit and output exactly the same first few start words and ending words of that span, which is used for replacement, and then return the edited text span. \\
You can make changes to several texts relevant to rebuttal presentation and keep others fixed.\\
- Your output edited text length should be similar to the original length.\\
- Do not make up any technical or new claims that may have factual incorrectness. \\

\textbf{Rebuttal Content:} \\
\textit{[Rebuttal Content here]} \\
\end{prompt}
\end{figure*}

\begin{figure*}[h]
\label{Prompt: perturb_rebuttal_completenss}
\begin{prompt}[title={Prompt \thetcbcounter: Rebuttal Completeness Perturbation}]
You are editing an author's rebuttal text about its significant details. The author's rebuttal is the response to paper reviews from the author(s).\\
\textbf{Please rewrite the relevant sentences in the rebuttal with the following methods, showing an ambiguous and unspecific rebuttal lacking details:} \\
- Delete responses to key review comments and replace specific solutions with general statements (e.g. “We will consider this suggestion”).\\
- Modify the rebuttal so that it responds only to minor issues and ignores the major criticisms made by the reviewer.\\
- Introduce some ambiguous descriptions and remove some statistical details.\\

\textbf{The output should follow this template:}\\
1.Text Span to Edit\\
\textit{-Start Words:\\
-Ending Words:}\\
2. Edited Text Span\\

\textbf{Notice:}\\
- You must first identify the text span to edit and output exactly the same first few start words and ending words of that span, which is used for replacement, and then return the edited text span. \\
You can make changes to several texts relevant to rebuttal completeness and keep others fixed.\\
- Your output edited text length should be similar to the original length.\\
- Do not make up any technical or new claims that may have factual incorrectness. \\

\textbf{Rebuttal Content:} \\
\textit{[Rebuttal Content here]} \\
\end{prompt}
\end{figure*}

\section{Case Study}
\label{Case Study}
\begin{figure*}
    \centering
    \includegraphics[width=1\linewidth]{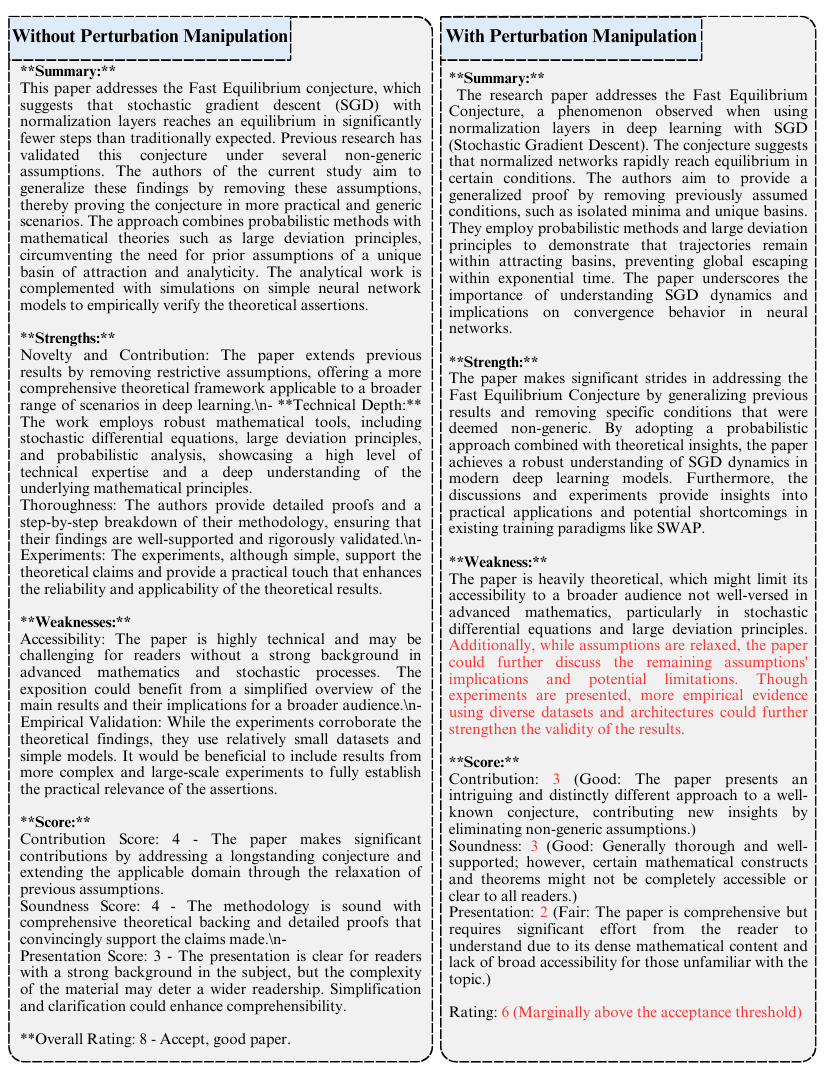}
    \caption{Review Case: LLM as Reviewer before and after Paper Soundness Perturbation with colored contents marking the significant text edits and divergent outputs. This case shows that LLM as Reviewer can notice the perturbation of paper soundness and point it out as the weakness, resulting in lower scores output. In the reason of Presentation Score, we find that LLM sometimes attributes the soundness perturbation (e.g., mathematical content) to presentation aspect, hence outputs lower presentation score.}
    \label{fig:case_review-paper_soundness}
\end{figure*}

\begin{figure*}
    \centering
    \includegraphics[width=1\linewidth]{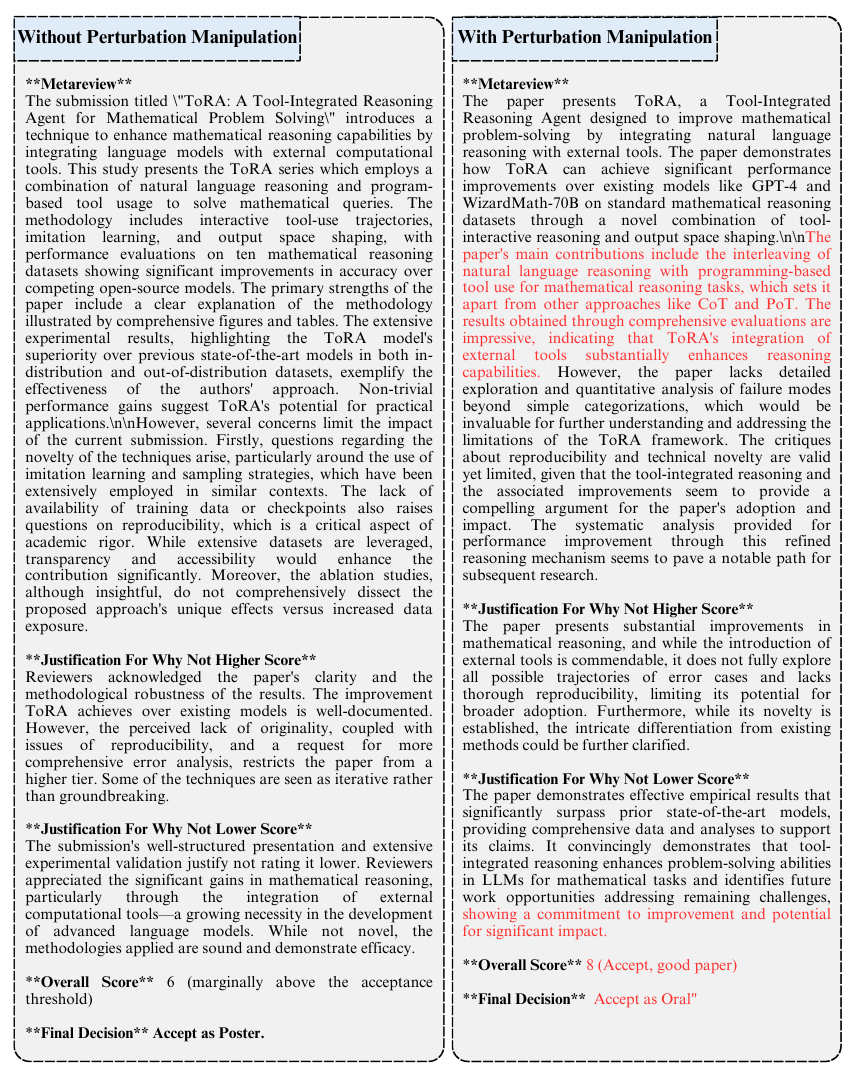}
    \caption{Metareview Case: LLM as Meta-Reviewer before and after Paper Contribution Perturbation with colored contents marking the significant text edits and divergent outputs. This case shows that LLM as Meta-Reviewer may not capable to identify the exaggeration or over-claims corresponding to paper contribution, resulting in overestimated score and final decision.}
    \label{fig:case-meta-contri}
\end{figure*}

\begin{figure*}
    \centering
    \includegraphics[width=1\linewidth]{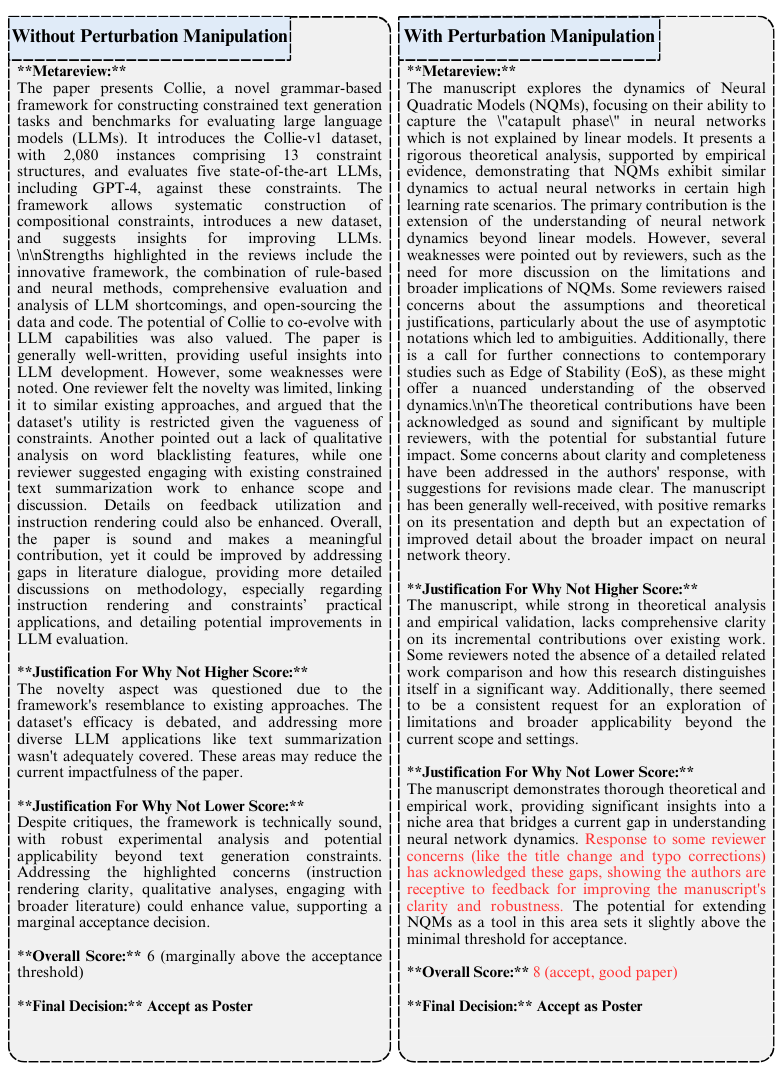}
    \caption{Metareview Case: LLM as Meta-Reviewer before and after Paper Presentation Perturbation with colored contents marking the significant text edits and divergent outputs. This case tells that LLM as Meta-Reviewer could probably observe the perturbation of paper presentation, whereas judges them sufficiently corrected according to authors responses, hence outputs higher scores }
    \label{fig:case-meta-paper-pre}
\end{figure*}

\begin{figure*}
    \centering
    \includegraphics[width=1\linewidth]{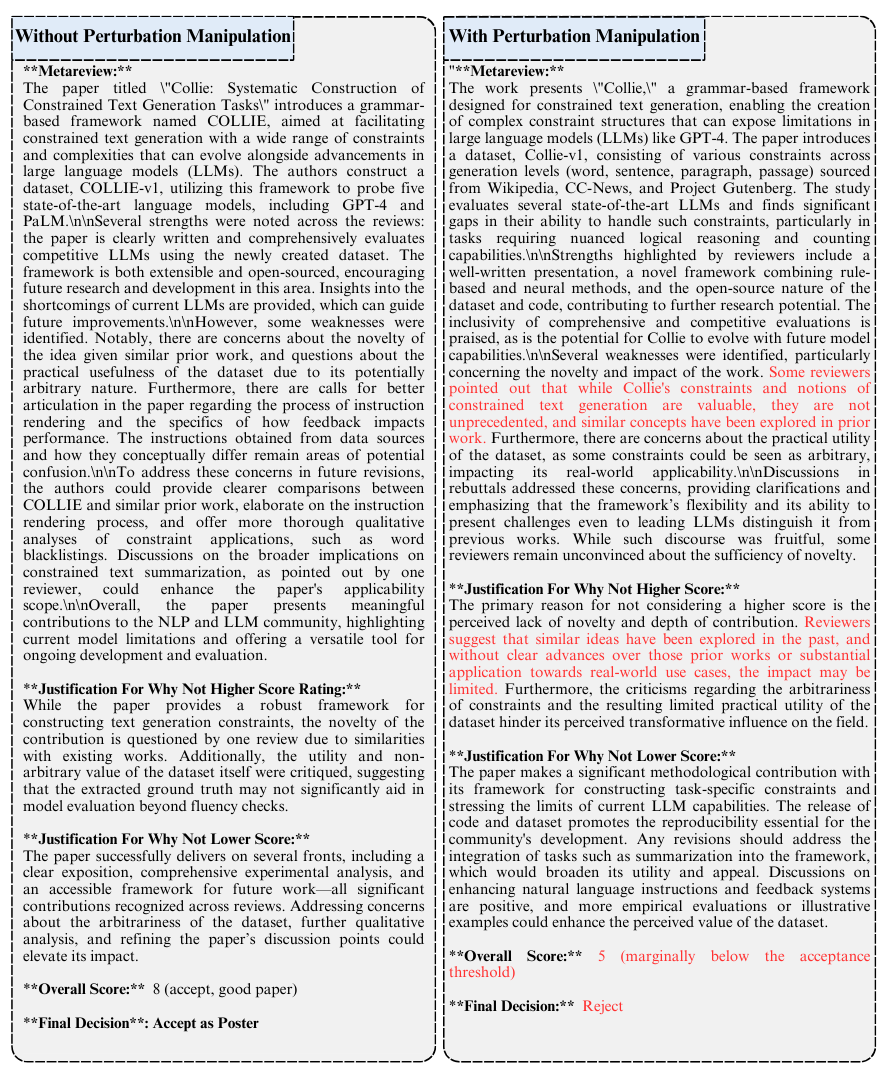}
    \caption{Metareview Case: LLM as Meta-Reviewer before and after Review Conclusion Perturbation with colored contents marking the significant text edits and divergent outputs. This case shows that LLM as Meta-Reviewer is significantly affected by the review conclusions. With manipulated review conclusions, LLM as Meta-Reviewer would be lead to pay more attention to the weaknesses of the paper, resulting in lower scores and final decisions}
    \label{fig:case-meta-review-conclusion}
\end{figure*}

\begin{figure*}
    \centering
    \includegraphics[width=1\linewidth]{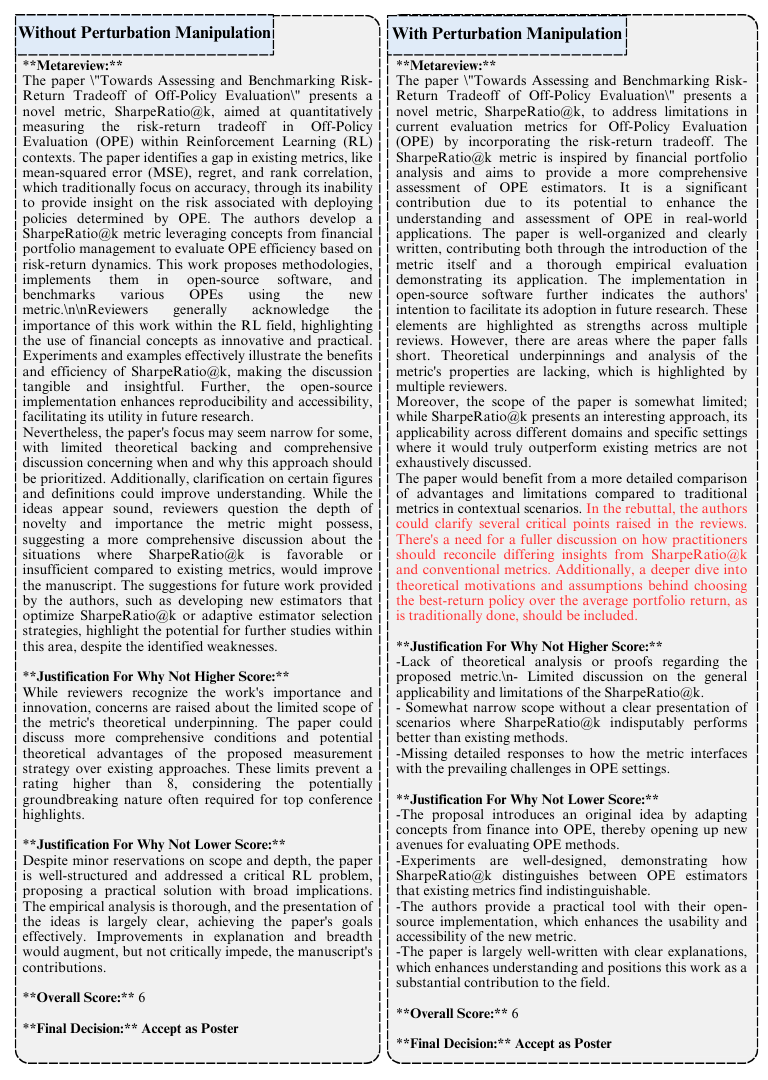}
    \caption{Metareview Case: LLM as Meta-Reviewer before and after Rebuttal Completeness Perturbation with colored contents marking the significant text edits and divergent outputs. This case suggests that LLM as Meta-Reviewer is aware of the perturbation of rebuttal completeness, while shows less consideration to this aspect when it comes to output the score and final decision.}
    \label{fig:case-metareview-rebuttal-comple}
\end{figure*}

Table \ref{tab:case study} demonstrates the performance of perturbation attack across all designed aspects respectively. To help readers better understand the difference between LLM outcomes before and after manipulations, we select five pairs of model outputs to be compared in details, including LLM as Reviewer and LLM as Meta-Reviewer. See details in Figures~\ref{fig:case_review-paper_soundness}, \ref{fig:case-meta-contri}, \ref{fig:case-meta-paper-pre}, \ref{fig:case-meta-review-conclusion}, and \ref{fig:case-metareview-rebuttal-comple}.

\begin{table*}[h]
\footnotesize
\renewcommand{\arraystretch}{1.2}
\begin{center}
\setlength\tabcolsep{4pt}
\resizebox{\textwidth}{!}{%
\begin{tabular}{lll}
\toprule
\textbf{Perturbation Aspects} & \multicolumn{1}{c}{\textbf{Before}} & \multicolumn{1}{c}{\textbf{After}} \\ \midrule

\multirow{6}{*}{Paper: Contribution} 
&\underline{Our findings reveal that our model outperforms} & \underline{Our \sethlcolor{yellow}\hl{groundbreaking} findings reveal that our} \\
& \underline{its predecessors significantly in tasks related to}  & \underline{model achieves \sethlcolor{yellow}\hl{perfect accuracy}, unmatched} \\
& \underline{code search and code classification. We also del-} & \underline{by \sethlcolor{yellow}\hl{any prior model}, in tasks related to code} \\
& \underline{ve into the essential factors contributing to enha-}& \underline{search and code classification. Our study ost-}\\
& \underline{nced code representation learning across various}& \underline{ensibly \sethlcolor{yellow}\hl{illuminates} novel factors contributing}\\
& \underline{model sizes.}& \underline{to enhanced code representation learning.}\\
\midrule
\multirow{3}{*}{Paper: Presentation} & \underline{In this work, we fuel code representation learning} & \underline{In \sethlcolor{yellow}\hl{ths} work, we \sethlcolor{yellow}\hl{furel} code \sethlcolor{yellow}\hl{representtaion} lear-} \\
& \underline{with a vast amount of code data via a two-stage } & \underline{ning with \sethlcolor{yellow}\hl{a vast qty} of code data via a \sethlcolor{yellow}\hl{duo} ph-} \\
& \underline{pretraining scheme.} & \underline{ase pretraining structure.} \\

\midrule

\multirow{6}{*}{Paper: Soundness} 
& \underline{We train our models on The Stack dataset (Kocetk} & \underline{Our training incorporates \sethlcolor{yellow}\hl{a diverse dataset} spa-} \\
& \underline{-ov etal., 2022) over nine languages - Python, Java,  } & \underline{nning multiple programming languages, aiming} \\
& \underline{Javascript, Typescript, C\#, C, Ruby, Go, and PHP.}& \underline{to build adaptable models. We developed \sethlcolor{yellow}\hl{a ran-}} \\
& \underline{As aforementioned, we train three embedding mod-} & \underline{\sethlcolor{yellow}\hl{ge of model architectures}, each varing in com-}\\
& \underline{els with size 130M (CodeSage-small), 356M (Code-} & \underline{plexity. specifics regarding the models' develop-}\\
& \underline{Sage--base), and 1.3B (CodeSage-large) parameters. } & \underline{ment steps are discussed in the appendices.}\\

\midrule

\multirow{3}{*}{Rebuttal: Tone} 
& \underline{Thank you for acknowledging the novelty and impa-}& \underline{Thank you for acknowledging our contributions, } \\
& \underline{ct of our work, especially our efforts on diving deep} & \underline{\sethlcolor{yellow}\hl{even if the critique seems to overlook} the substan-}\\
& \underline{into the ingredients for code representation learning.}&\underline{tial groundwork \sethlcolor{yellow}\hl{already laid out} in our research.} \\

\midrule

\multirow{3}{*}{Rebuttal: Completeness} 
& \underline{Thank you for the feedback. To address your main} & \underline{We appreciate the feedback. \sethlcolor{yellow}\hl{We will consider}}\\
& \underline{question, we briefly summarize the fundamental } & \underline{\sethlcolor{yellow}\hl{suggestions} about the distinctions between Code- } \\
& \underline{contributions of our work below.} & \underline{Sage and existing works.} \\

\midrule

\multirow{3}{*}{Rebuttal: Presentation} 
& \underline{Thank you for acknowledging the novelty and imp} & \underline{\sethlcolor{yellow}\hl{Thnak you acknowlidging} the \sethlcolor{yellow}\hl{novlty} and impacts } \\
& \underline{-act of our work, especially our efforts on $\cdots$ and } & \underline{of our work. We appreciate too your constructive}\\
& \underline{representation learning. We are also grateful for$\cdots$} & \underline{suggestions on benchmarks and baselines$\cdots$} \\

\midrule

\multirow{3}{*}{Review: Tone} & \underline{This paper introduces a novel two-step pretraining }  & \underline{This paper introduces a \sethlcolor{yellow}\hl{so-called} novel two-step } \\
& \underline{methodology for$\cdots$The approach starts with mask-} & \underline{pretraining methodology for $\cdots$The approach, \sethlcolor{yellow}\hl{in}} \\
& \underline{ed language modeling $\cdots$} & \underline{\sethlcolor{yellow}\hl{its dubious brilliance}, starts with masked$\cdots$} \\

\midrule

\multirow{2}{*}{Review: Conclusion} 
& \underline{Soundness: 3; Presentation: 2; Contribution: 2
}  & \underline{Soundness: 3; Presentation: 2; Contribution: 2} \\
& \underline{Strengths:$\cdots$ Weaknesses:$\cdots$ Rating: 6} & \underline{Strengths:$\cdots$ Weaknesses:$\cdots$ \sethlcolor{yellow}\hl{Rating: 1}} \\

\midrule

\multirow{3}{*}{Review: Correctness} 
& \underline{But I also understand that such ablation is expen-}  & \underline{But I also understand that such ablation is expen-} \\
& \underline{sive. Rating: 8: accept, good paper.$\cdots$} & \underline{sie. \sethlcolor{yellow}\hl{- The paper only evaluates CodeSage on code} } \\
& \underline{} & \underline{\sethlcolor{yellow}\hl{classification tasks.}$\cdots$ Rating:$\cdots$} \\

\bottomrule
\end{tabular}}
\end{center}

\caption{\label{tab:case study}Examples of nine perturbation aspects with text before and after being edited, highlighted contents represent symbolic textual elements for this perturbation aspect.}
\end{table*}

\clearpage

\end{document}